%% file: main.tex
\documentclass[10pt,twocolumn,letterpaper]{article}

\usepackage[pagenumbers]{cvpr} 
\usepackage{multirow}
\usepackage{makecell}
\input{preamble}

\definecolor{cvprblue}{rgb}{0.21,0.49,0.74}
\usepackage[pagebackref,breaklinks,colorlinks,citecolor=cvprblue]{hyperref}

\def\paperID{8681} 
\def\confName{CVPR}
\def\confYear{2024}

\title{Anything in Any Scene: Photorealistic Video Object Insertion}

\author{Chen Bai, Zeman Shao, Guoxiang Zhang, Di Liang, Jie Yang, Zhuorui Zhang, Yujian Guo,\\ Chengzhang Zhong, Yiqiao Qiu, Zhendong Wang, Yichen Guan, Xiaoyin Zheng, Tao Wang, Cheng Lu\\
XPeng Motors\\
{\tt\small \{chenbai, zemans, guoxiangz, liangd2, yangj23, zhangzr, guoyj4, chengzhangz, yiqiaoq}\\
{\tt\small zhendongw, guanyc, xiaoyinz, taow, luc\}@xiaopeng.com}
}

\begin{document}
\maketitle
\input{sec/0_abstract}    
\input{sec/1_intro}

\input{sec/2_related}
\input{sec/3_method}

\input{sec/4_experiment}
\input{sec/5_conclusion}
{
    \small
    \bibliographystyle{ieeenat_fullname}
    \bibliography{main}
}

\input{sec/X_suppl}

\end{document}

%% file: preamble.tex
\usepackage[dvipsnames]{xcolor}


%% file: sec/0_abstract.tex
\begin{abstract}
Realistic video simulation has shown significant potential across diverse applications, from virtual reality to film production.
This is particularly true for scenarios where capturing videos in real-world settings is either impractical or expensive.
Existing approaches in video simulation often fail to accurately model the lighting environment, represent the object geometry, or achieve high levels of photorealism.
In this` paper, we propose Anything in Any Scene, a novel and generic framework for realistic video simulation that seamlessly inserts any object into an existing dynamic video with a strong emphasis on physical realism.
Our proposed general framework encompasses three key processes:
1) integrating a realistic object into a given scene video with proper placement to ensure geometric realism;
2) estimating the sky and environmental lighting distribution and simulating realistic shadows to enhance the light realism;
3) employing a style transfer network that refines the final video output to maximize photorealism.
We experimentally demonstrate that Anything in Any Scene framework produces simulated videos of great geometric realism, lighting realism, and photorealism.
By significantly mitigating the challenges associated with video data generation, our framework offers an efficient and cost-effective solution for acquiring high-quality videos.
Furthermore, its applications extend well beyond video data augmentation, showing promising potential in virtual reality, video editing, and various other video-centric applications.
Please check our project website~\url{https://anythinginanyscene.github.io} for access to our project code and more high-resolution video results.
\end{abstract}

%% file: sec/1_intro.tex
\section{Introduction}
\label{sec:intro}
The image and video simulation has exhibited success in various applications, ranging from virtual reality to film production.
The capability to generate diverse and high-quality visual content through realistic image and video simulation holds the potential to advance these fields, introducing new possibilities and applications.
Although the images and videos captured in real-world settings are invaluable for their authenticity, they often suffer from the limitation of long-tail distribution.
This results in common scenarios being over-represented, while rare yet crucial situations are under-represented, presenting a challenge known as the out-of-distribution problem.
Traditional methods of addressing these limitations through video collection and editing prove impractical or excessively costly due to the inherent difficulty in encompassing all possible situations.
The significance of video simulation, especially through the integration of existing videos with newly inserted objects, becomes paramount in overcoming these challenges. 
By generating large-scale, diverse, and realistic visual content, video simulation contributes to the enhancement of applications in virtual reality, video editing, and video data augmentation.

\begin{figure*}
     \centering
     \begin{subfigure}[t]{0.32\textwidth}
         \centering
         \includegraphics[width=\textwidth]{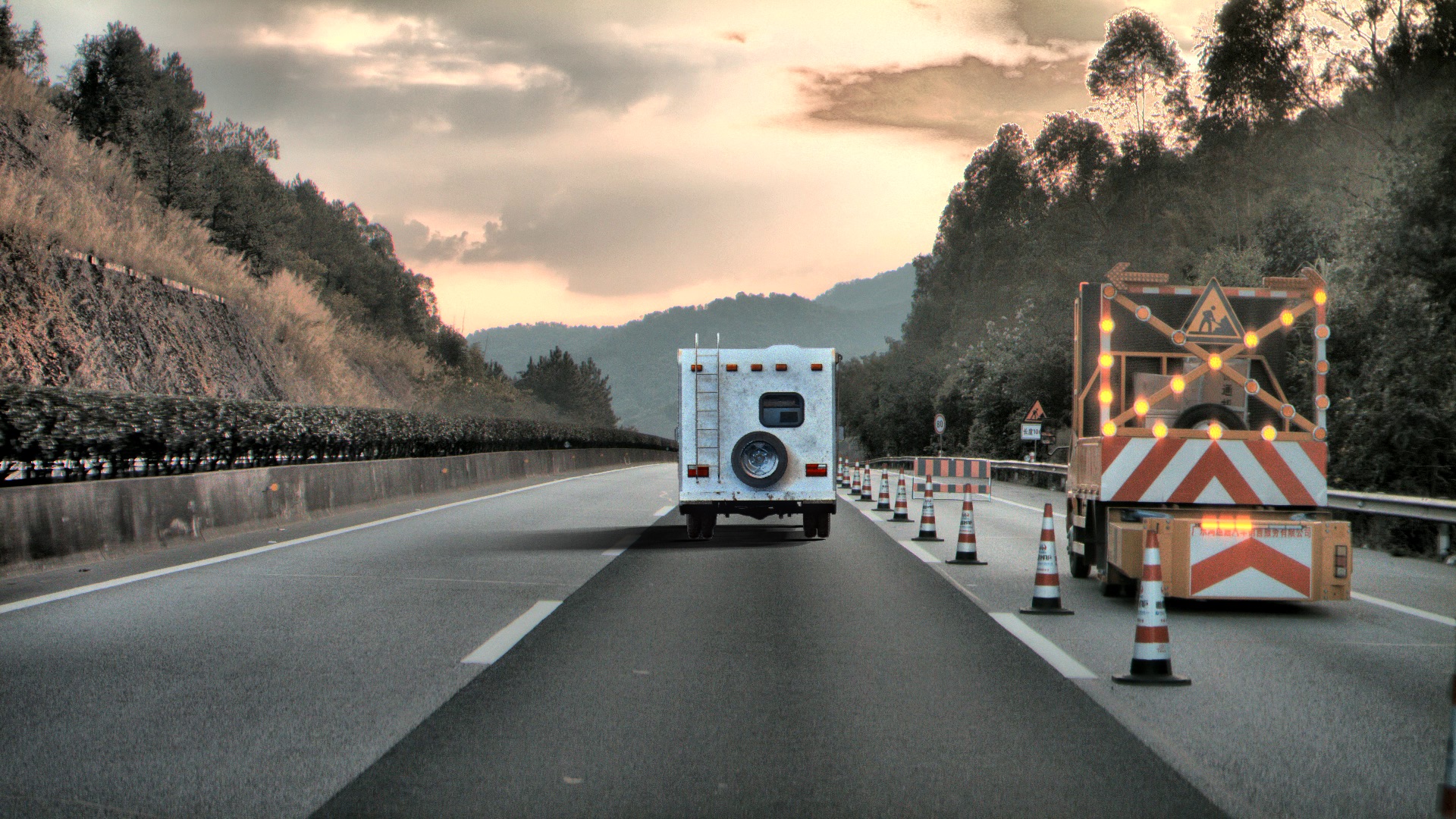}
         \caption{The inserted car has an inconsistent shadow to another car because of the wrong lighting environment estimated.}
         \label{fig:failure-light}
     \end{subfigure}
     \hfill
     \begin{subfigure}[t]{0.32\textwidth}
         \centering
         \includegraphics[width=\textwidth]{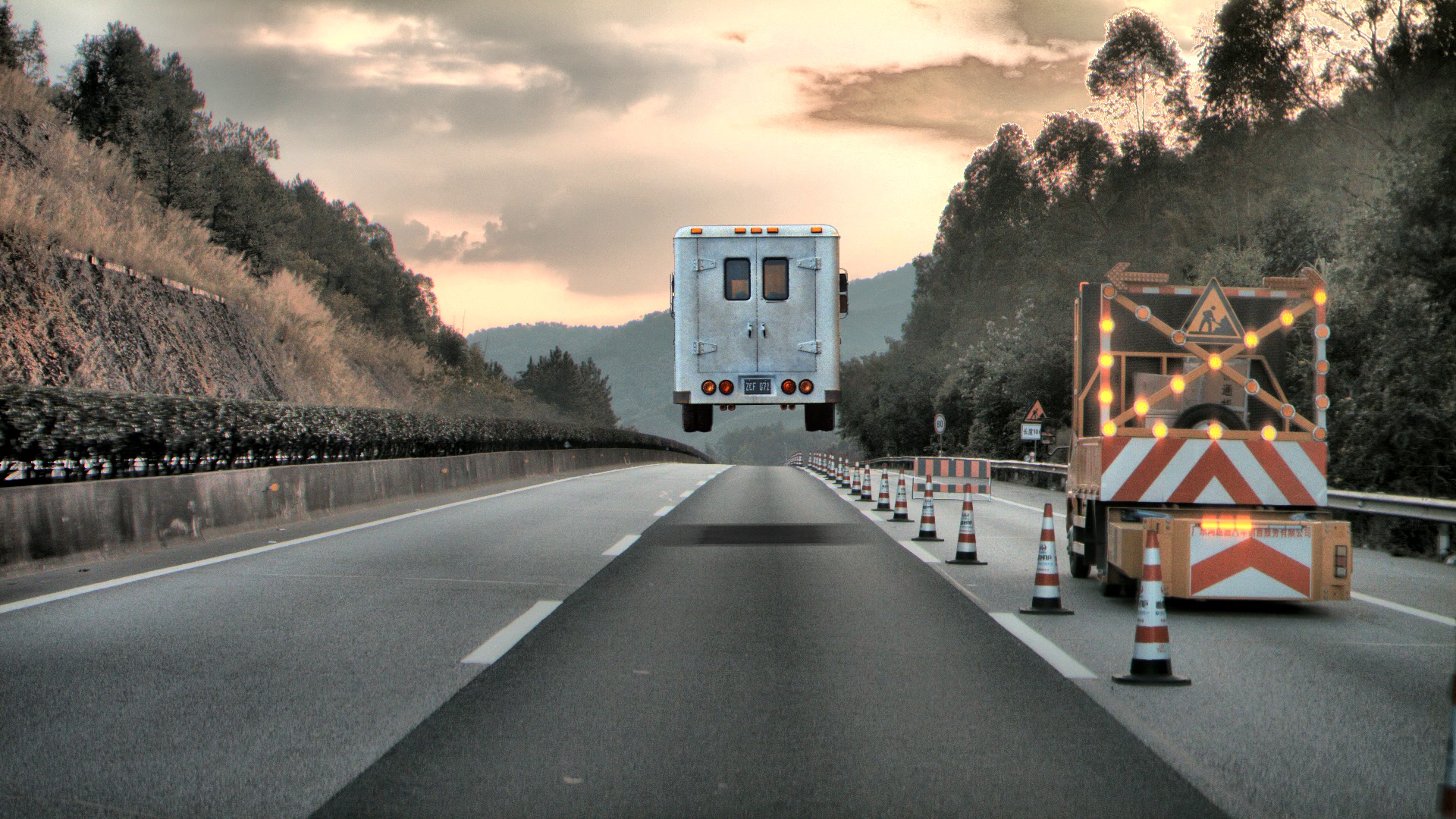}
         \caption{The car is in the air because of a wrong placement location determined.}
         \label{fig:failure-placement}
     \end{subfigure}
     \hfill
     \begin{subfigure}[t]{0.32\textwidth}
         \centering
         \includegraphics[width=\textwidth]{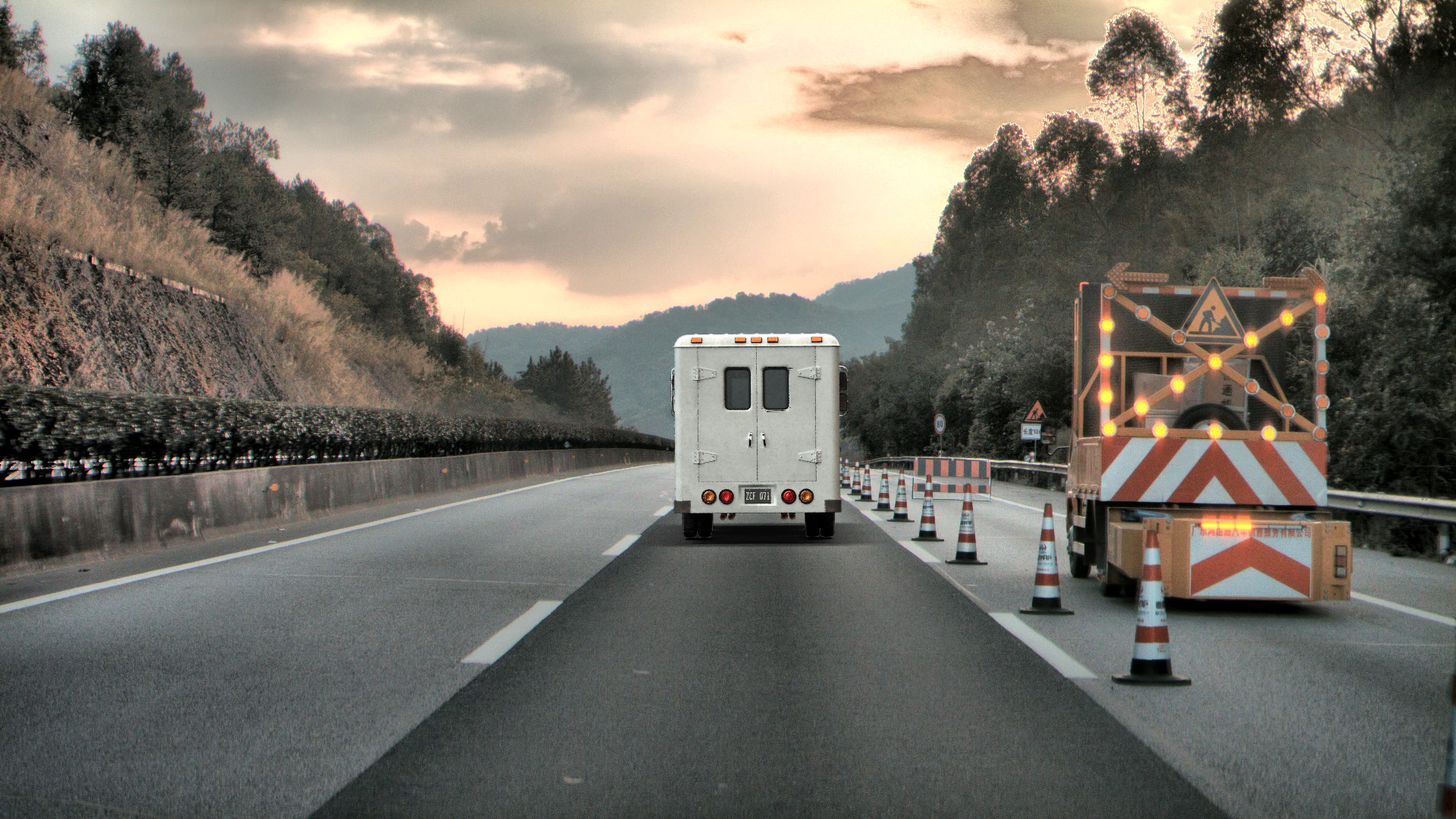}
         \caption{The inserted car in the scene has a significant difference in texture compared to another car, which makes the image lack photorealism.}
         \label{fig:failure-style}
     \end{subfigure}
        \caption{Examples of simulated video frame with wrong lighting environment estimation, false object placement position, and unrealistic texture style, which make the image lack physical realism}
        \label{fig:failure}
        \vspace{-2.5mm}
\end{figure*}

However, generating a realistic simulated video with consideration of physical realism is still a challenging open problem.
Existing methods often exhibit limitations by concentrating on specific settings, particularly indoor environments~\cite{savva2017minos, savva2019habitat, coumans2016pybullet, xia2018gibson, kolve2017ai2}.
These methods may not adequately address the complexities of outdoor scenes, including diverse lighting conditions and fast-moving objects.
Methods relying on 3D model registration are constrained in integrating only limited classes of objects~\cite{dosovitskiy2017carla, richter2016playing, ros2016synthia, li2019aads}.
Many approaches neglect essential factors such as modeling the lighting environment, proper object placement, and achieving photorealism~\cite{martinez2017beyond, dosovitskiy2017carla}.
Failed cases are illustrated in Figure~\ref{fig:failure}.
Consequently, these limitations significantly constrain their applications in fields that need highly scalable, geometrically consistent, and realistic scene video simulation, such as autonomous driving and robotics.

In this paper, we propose a comprehensive framework Anything in Any Scene for the photorealistic video object insertion that addresses these challenges.
The framework is designed to have universal applicability, and is adaptable to both indoor and outdoor scenes, ensuring physical accuracy in terms of geometric realism, lighting realism, and photorealism.
Our goal is to create video simulations that are not only beneficial for visual data augmentation in machine learning but also adaptable to various video applications, such as virtual reality and video editing. 

\begin{figure*}[t]
     \centering
     \includegraphics[width=\textwidth]{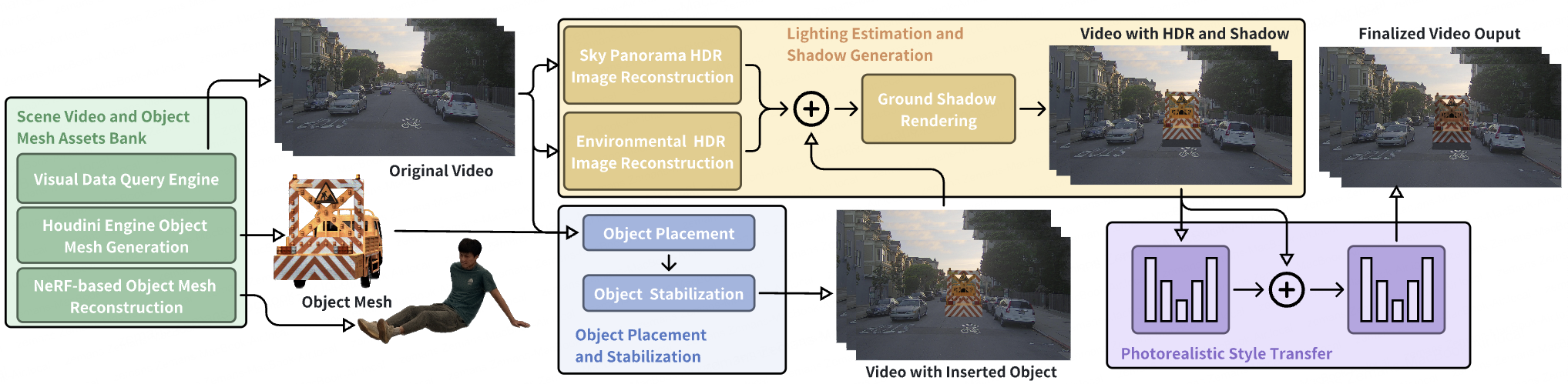}
    \caption{Overview of proposed Anything in Any Scene framework for photorealistic video object insertion}
    \label{fig:overview}
    \vspace{-2.5mm}
\end{figure*}

The overview of our Anything in Any Scene framework is shown in Figure~\ref{fig:overview}.
We detail our novel and scalable pipeline for building a diverse asset bank of scene video and object mesh in Section~\ref{sec:method-data}.
We introduce a visual data query engine designed to efficiently retrieve relevant video clips from visual queries using descriptive keywords.
Following this, we present two methods for generating 3D meshes, leveraging existing 3D assets as well as multi-view image reconstructions. 
This allows the insertion of any desired object without limitation, even if it is highly irregular or semantically weak.
In Section~\ref{sec:method-render}, we detail our approach for integrating objects into dynamic scene video with a focus on maintaining physical realism.
We design an object placement and stabilization method described in Section~\ref{sec:method-render-placement}, ensuring the inserted object is stably anchored across continuous video frames.
Addressing the challenge of creating realistic lighting and shadow effects, we estimate sky and environmental lighting and generate realistic shadows during the rendering process, as described in Section~\ref{sec:method-render-lighting}. 
The resulting simulated video frames inevitably contain unrealistic artifacts that differ from real-world captured videos, such as imaging quality discrepancies in noise level, color fidelity, and sharpness.
We adopt a style transfer network to enhance the photorealism in Section~\ref{sec:method-render-style}.

The simulated videos produced from our proposed framework reach a high degree of lighting realism, geometrical realism, and photorealism, outperforming the others both qualitatively and quantitatively as shown in Section~\ref{sec:experiment-result}.
We further showcase in Section~\ref{sec:experiment-downstream} the application of our simulated videos in the training perception algorithm to verify its practical value.
The Anything in Any Scene framework is able to create a large-scale, low-cost video dataset for data augmentation with time efficiency and realistic visual quality, which alleviates the burden of video data generation and potentially ameliorates the long-tail distribution and out-of-distribution challenges.
With its generic framework design, the Anything in Any Scene framework can easily incorporate improved models and new modules, such as an improved 3D mesh reconstruction method, further enhancing video simulation performance.

Our main contributions can be summarized as follows:
\begin{enumerate}
    \item We introduce a novel and scalable Anything in Any Scene framework for video simulation, capable of integrating any object into any dynamic scene video.
    \item Our framework uniquely focuses on preserving geometric realism, lighting realism, and photorealism in video simulations, ensuring high-quality and realistic outputs.
    \item  We conducted extensive validations, demonstrating the ability of the framework to produce realistic video simulations, significantly expanding the scope and potential application in this field.
\end{enumerate}

%% file: sec/2_related.tex
\section{Related Work}
\label{sec:related}

\textbf{Image Synthesis and Editing}: Encompassing tasks from image inpainting to style transfer has attracted significant attention in both academic and industry communities.
The traditional methods are mostly based on pixels, patches, and low-level image features, often lacking high-level semantic information.
Specifically, the image inpainting methods replicate pixels or patches for image recovery~\cite{criminisi2003object,komodakis2007image, hays2007scene, barnes2009patchmatch, barnes2015patchtable}.
The non-parametric-based texture synthesis methods re-sample the pixels of a given source texture to generate photorealistic textures~\cite{efros1999texture,kwatra2003graphcut}.
The style transfer methods, such as image analogies~\cite{hertzmann2023image}, perform example-based stylization using patches. 

Deep learning networks, particularly Generative Adversarial Networks (GAN)~\cite{gan}, have demonstrated significant capabilities in computer vision and image processing tasks, achieving impressive success in image generation.
Various GANs, such as MGANs~\cite{mgan}, SGAN~\cite{sgan}, and PSGAN~\cite{psgan}, have shown remarkable proficiency in the task of texture synthesis.
Additionally, GANs have been successfully applied to contextual image inpainting~\cite{pathak2016context} and multi-scale image completion~\cite{yang2017high}.
The pix2pix~\cite{pix2pix} and cycleGAN~\cite{cycleGAN} leverage GAN architecture to train generative models for style transfer.
The images generated by GANs tend to be less blurred and exhibit higher realism, aligning closely with distributions of training image data.

\textbf{Video Synthesis and Editing}: Transitioning from image to video synthesis requires addressing additional challenges, particularly maintaining temporal consistency.

Unconditional video synthesis methods, such as~\cite{saito2017temporal},~\cite{tulyakov2018mocogan}, and~\cite{vondrick2016generating}, take a random noise as input and model both spatial and temporal correlation to generate video.
However, they often result in constrained motion patterns in output video sequences.
In contrast, conditional video synthesis methods employ conditional GAN~\cite{cgan} to train a generative model for video generation based on input content.
In~\cite{wang2018video} and its following work~\cite{wang2019few}, the generative network is conditioned on the previous frame of the source video for each subsequent frame generation.
~\cite{mallya2020world} take this approach further by considering all previously generated frames, achieving improved long-term temporal consistency in their video synthesis.

Additionally, the automatic video synthesis methods proposed in ~\cite{lee2019inserting} and ~\cite{huang2019temporally} insert the object's video into another video using spatial and temporal information.
Recently, the GeoSim framework proposed in~\cite{geosim} has achieved impressive results in car insertion into a given real-world driving scene video, though its application to less common objects and diverse types of scene video remains limited.
Our work seeks to bridge this gap, expanding the potential for any object insertion in any scene video.

%% file: sec/3_method.tex
\section{Scene Video and Object Mesh Assets Bank}
\label{sec:method-data}
Our goal with the Anything in Any Scene framework is to generate large-scale and high-quality simulation videos by composition of dynamic scene videos and objects of interest.
To achieve this, an assets bank of both scene videos and object meshes is required for simulated video composition.

In order to efficiently locate target videos for composition from a large-scale video assets bank, we proposed a visual data query engine that is used to retrieve the relevant scene video clips for simulated video composition based on the given visual clue descriptors.
The mesh model of the target object is required before its insertion into an existing video clip. 
We introduced the 3D mesh generation of the target object by using the Houdini Engine from existing 3D assets and a NeRF-based 3D reconstruction from multi-view images, which enables theatrically unlimited classes of objects to be inserted into the existing scene video.

Detailed descriptions of our mesh assets bank can be found in supplementary materials.

\section{Realistic Video Simulation}
\label{sec:method-render}
To achieve video simulation with geometric realism, lighting realism, and photorealism, our proposed framework consists of the following three main components:
\begin{enumerate}
  \item Object Placement and Stabilization (Section~\ref{sec:method-render-placement})
  \item Lighting and Shadow Generation (Section~\ref{sec:method-render-lighting})
  \item Photorealistic Style Transfer (Section~\ref{sec:method-render-style})
\end{enumerate}
\input{sec/3_2_1_method_placement}
\input{sec/3_2_2_method_lighting}

\input{sec/3_2_3_method_style}

%% file: sec/3_2_1_method_placement.tex
\begin{figure*}[t]
     \centering
     \begin{subfigure}[t]{0.327\textwidth}
         \centering
         \includegraphics[width=\textwidth]{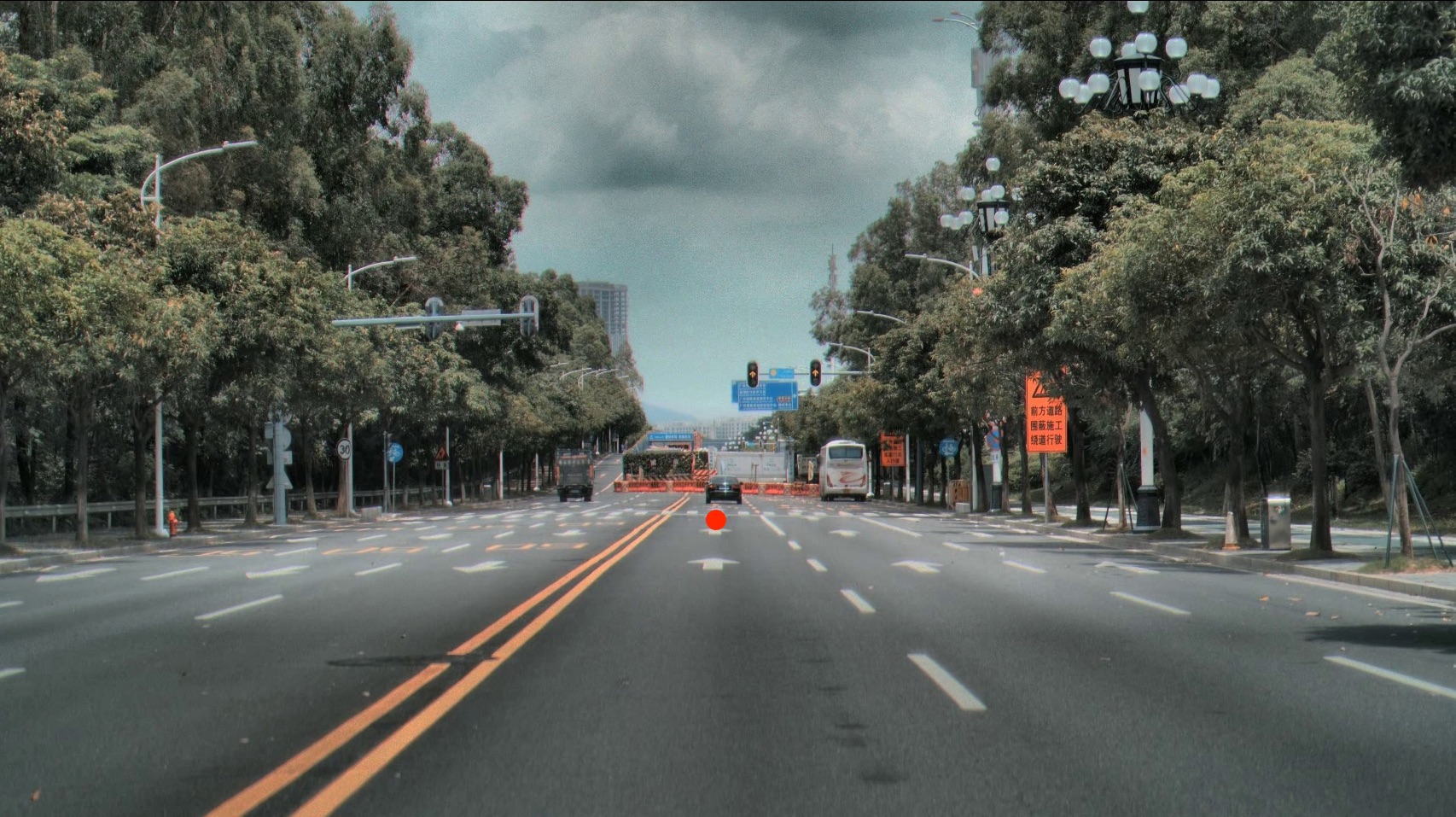}
         \caption{The first frame of video clip $I_1$}
         \label{fig:place-2d}
     \end{subfigure}
     \hfill
     \begin{subfigure}[t]{0.325\textwidth}
         \centering
         \includegraphics[width=\textwidth]{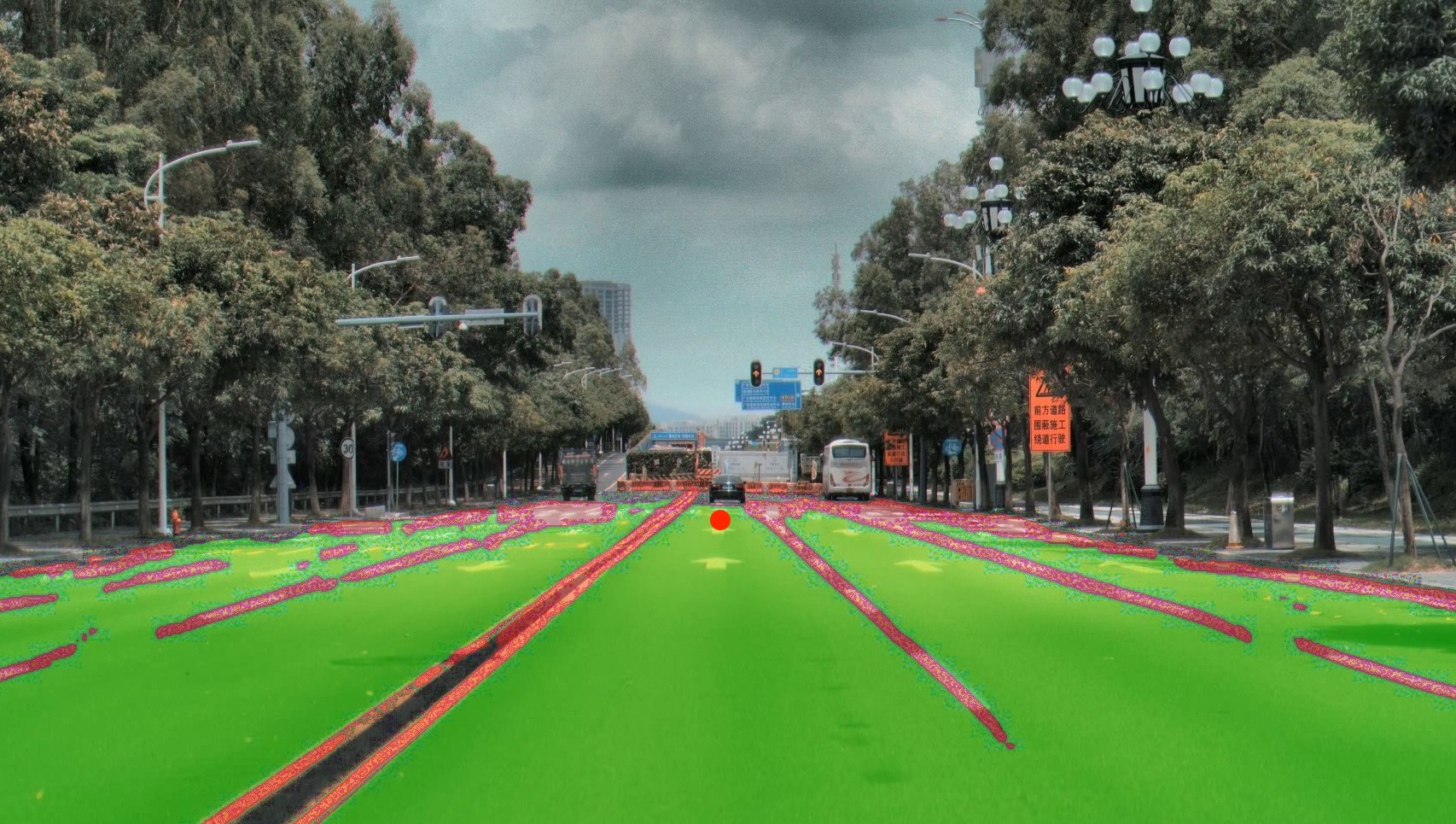}
         \caption{The estimated segmentation mask $\hat{M}_1$}
         \label{fig:place-seg}
     \end{subfigure}
     \hfill
     \begin{subfigure}[t]{0.338\textwidth}
         \centering
         \includegraphics[width=\textwidth]{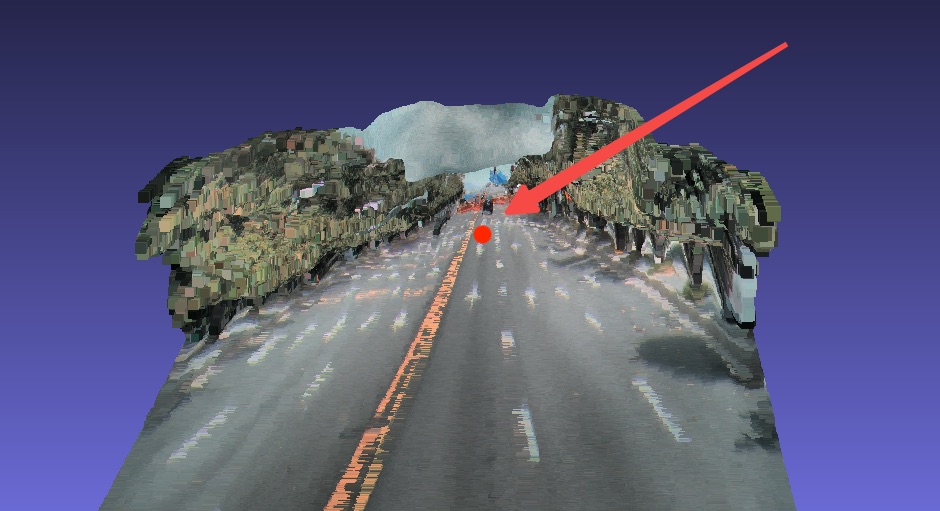}
         \caption{The object placement location in 3D scene}
         \label{fig:place-3d}
     \end{subfigure}
     \hfill
    \caption{Example of driving scene video for object placement. The red point in each image is the location for object insertion.}
     \label{fig:place}
     \vspace{-2.5mm}
\end{figure*}

\subsection{Object Placement and Stabilization}
\label{sec:method-render-placement}
Inserting an object into a background video for video composition requires the object placement location determined for each frame in the video sequence.
We designed and proposed a novel object placement method with the consideration of occlusion with other existing objects in the scene, which is described in Section~\ref{sec:method-render-placement-projection}.

However, placement locations that are independently estimated from each single frame could yield unrealistic movement tracks since the video temporal information has not been considered.
To address this issue, we propose an object placement stabilization method in Section~\ref{sec:method-render-placement-stable} to correct the placement location in each frame. 
We employ optical flow tracking between consecutive frames to ensure the inserted object behaves realistically across the continuous video frames.

\subsubsection{Object Placement}
\label{sec:method-render-placement-projection}
Suppose there are $N+T$ continuous frames, the first $N$ frames are the target frames that we aim to integrate the inserted object into, and the last $T$ frames are used as reference for object placement.
We assume the world coordinate of camera location in the frame $I_{N+T}$ is the origin $O_w = [0,0,0,1]$, and the camera coordinate system is aligned with the world coordinate system at this frame $I_{N+T}$.
We place the inserted object at the location of origin in the world coordinate which is the same location as the camera itself in the frame $I_{N+T}$.
To determine the pixel coordinates for object placement in the first $N$ consecutive frames, we project the origin from the world coordinate to the pixel coordinate based on the camera intrinsic matrix $\mathbf{K}$ and the camera pose including rotation matrix $\mathbf{R}_n$ and translation vector $\mathbf{t}_n$ at each frame $I_n$.
The placement pixel coordinate $\Tilde{o}_n$ at the frame $I_n$ is determined by:
\begin{equation}
    \Tilde{o}_n = \mathbf{K}[\mathbf{R}_n|\mathbf{t}_n] O_w
    \label{equ:project}
\end{equation}

The placement of the inserted object within a video clip should avoid occlusion with other existing objects in the scene.
We estimated the semantic segmentation mask $\hat{M}_{n}$ for each frame $I_{n}$ by using off-the-shelf models.
The pixel $\hat{M}_{n}(\Tilde{o}_n)$ denotes the category at the pixel location $\Tilde{o}_n$, representing the origin in the world coordinate projected into the pixel coordinate in the frame $I_n$.
This predicted category serves as a reference to determine whether the projected point location for object insertion is occluded by other objects in the scene.

We show an example of a driving scene in Figure~\ref{fig:place}.
The first frame of the video clip and its associated estimated segmentation mask are shown in Figure~\ref{fig:place-2d} and Figure~\ref{fig:place-seg}.
The red point in Figure~\ref{fig:place-3d} is the origin of the world coordinate and also the camera location in the frame $I_{N+T}$, we placed the object at this location. 
As the estimated segmentation mask shown in Figure~\ref{fig:place-seg}, the green region indicates the road area and the red region indicates the road lane.
After the object placement location is projected back from the world coordinate to the pixel coordinate, the placement is located in the road area as indicated in the semantic segmentation, which is a plausible place to insert a road vehicle in a driving scene video.

\subsubsection{Object Placement Stabilization}
\label{sec:method-render-placement-stable}

Firstly, we select a 3D point with world coordinate $P_w = [X, Y, Z, 1]$, and follow the Equation~\ref{equ:project} to project it from the world coordinate into the pixel coordinate $\Tilde{p}_n$ in each frame $I_n$ of the first $N+1$ frames.
We then estimate the optical flow between each two consecutive frames and obtain the selected 3D point $P_w$ pixel coordinate $\hat{p}_n$ in the frame $I_{n}$ through the image warping of $\Tilde{p}_{n+1}$ and the estimated optical flow.
The object placement stabilization can be interpreted as the optimization of camera pose for each frame $I_n$.
Specifically, we optimize the camera pose rotation matrix $\mathbf{R}_n$ and translation vector $\mathbf{t}_n$ at each frame $I_n$ by minimizing the 3D-to-2D projection error of $\hat{p}_n$ with the comparison to $\Tilde{p}_n$.
To achieve a better performance in placement stabilization, we select M points and optimize the rotation matrix $\mathbf{R}'_n$ and translation vector $\mathbf{t}'_n$, which can be expressed as:
\begin{equation}
    \begin{split}
    (\mathbf{R}'_n,\mathbf{t}'_n) 
    & = \operatorname*{arg\,min}  \sum^M_{i=1}(\hat{p}_n-\Tilde{p}_n)^2 \\
    & = \operatorname*{arg\,min}_{(\mathbf{R}_n,\mathbf{t}_n)}  \sum^M_{i=1}(\hat{p}_n-\mathbf{K}[\mathbf{R}_n|\mathbf{t}_n] P_w)^2
    \end{split}
\end{equation}
Lastly, we update the rotation matrix and translation vector in Equation~\ref{equ:project} by $\mathbf{R}'_n$ and $\mathbf{t}'_n$, and calculate the updated object placement pixel coordinate $\Tilde{o}_n$ for each frame $I_n$.

We also adjust $X$ and $Y$ values of the selected 3D point $P_w$ to ensure that the projected 2D point can be tracked in consecutive frames based on the estimated optical flow.
For example in the driving scene view, we shifted the selected 3D points by adjusting the $Y$ value so that the projected 2D points are the corner points of the white road lane.

%% file: sec/3_2_2_method_lighting.tex
\subsection{Lighting Estimation and Shadow Generation}
\label{sec:method-render-lighting}
One important key to creating a realistic simulated video with an integrated object is to generate accurate lighting and shading effects for the inserted object.
The position and luminance of the lighting in the scene, such as the sun for the outdoor scene and the environment for the indoor scene, affect the inserted object's visual appearance during the rendering process.

To simulate an accurate lighting and shading effect during the rendering process, we first introduced a High Dynamic Range (HDR) panoramic image reconstruction method in Section~\ref{sec:method-render-lighting-sky}.
Lastly, we rendered the shadow of the inserted object based on the estimated position of the main lighting source in Section~\ref{sec:method-render-lighting-shadow}.

\begin{figure}[t]
     \centering
     \begin{subfigure}[t]{0.156\textwidth}
         \includegraphics[width=\textwidth]{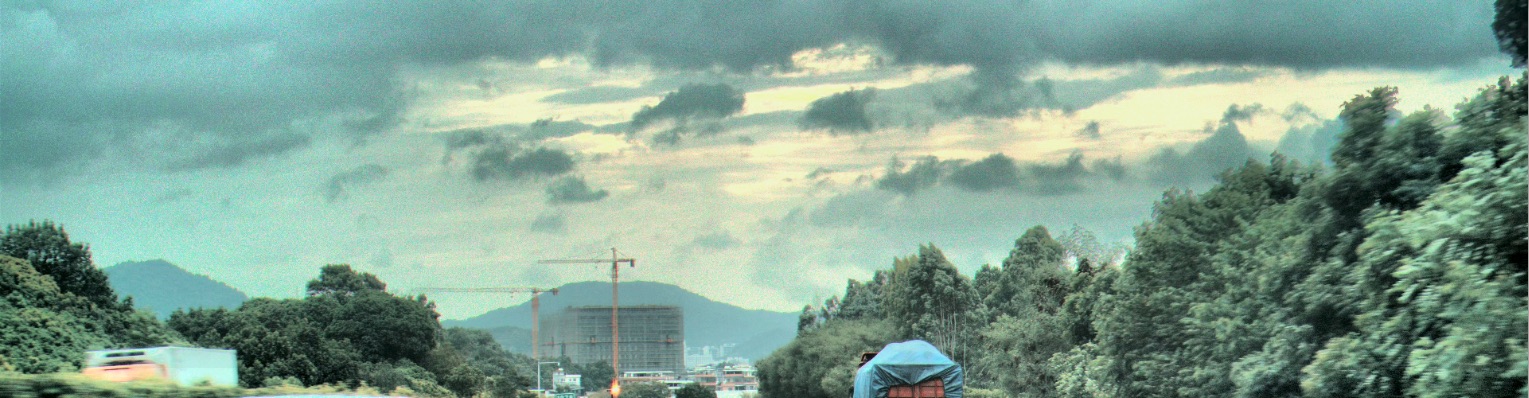}
     \end{subfigure}
     \hfill
     \begin{subfigure}[t]{0.156\textwidth}
         \includegraphics[width=\textwidth]{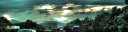}
     \end{subfigure}
     \hfill
     \begin{subfigure}[t]{0.156\textwidth}
         \includegraphics[width=\textwidth]{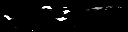}
     \end{subfigure}
     
     \begin{subfigure}[t]{0.156\textwidth}
         \includegraphics[width=\textwidth]{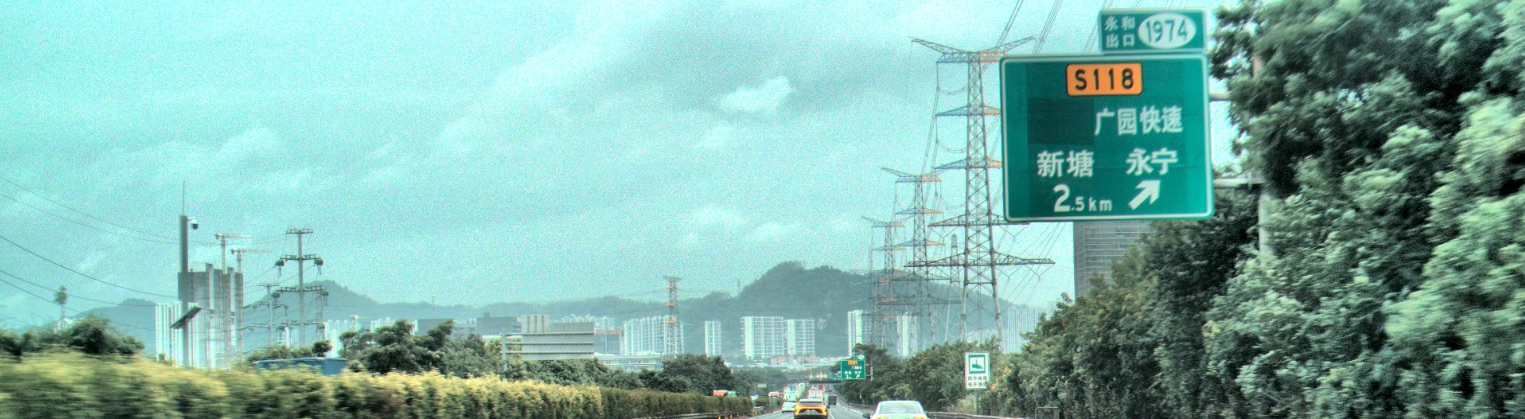}
     \end{subfigure}
     \hfill
     \begin{subfigure}[t]{0.156\textwidth}
         \includegraphics[width=\textwidth]{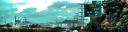}
     \end{subfigure}
     \hfill
     \begin{subfigure}[t]{0.156\textwidth}
         \includegraphics[width=\textwidth]{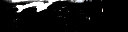}
     \end{subfigure}
     
     \begin{subfigure}[t]{0.156\textwidth}
         \includegraphics[width=\textwidth]{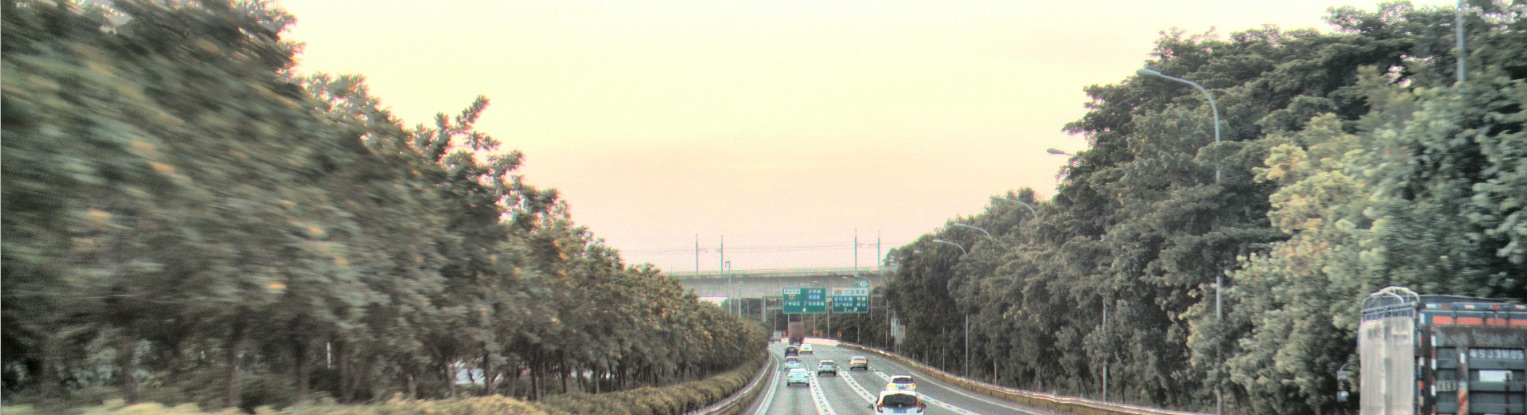}
         \caption*{Original Image}
     \end{subfigure}
     \hfill
     \begin{subfigure}[t]{0.156\textwidth}
         \includegraphics[width=\textwidth]{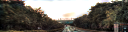}
         \caption*{HDR Image}
     \end{subfigure}
     \hfill
     \begin{subfigure}[t]{0.156\textwidth}
         \includegraphics[width=\textwidth]{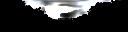}
         \caption*{Lighting Distribution}
     \end{subfigure}

    \caption{Examples of original sky image, reconstructed HDR image, and its associated sun lighting distribution map}
    \label{fig:light-sun}
    \vspace{-2.5mm}
\end{figure}

\subsubsection{HDR Panoramic Image Reconstruction}
\label{sec:method-render-lighting-sky}
The Low Dynamic Range (LDR) images captured by the consumer camera are usually over-saturated due to the extremely high brightness of the main lighting compared to surrounding environmental lighting, which makes it much more difficult to estimate the position and luminance distribution of the main lighting. 
To address this issue, we first use an image inpainting network to infer the surround view of lighting distribution for rendering.
We then adapt a sky HDR reconstruction network to identify the lighting source position and generate the HDR panoramic image.

\textbf{Panorama Image Inpainting}:
The image captured by the consumer camera has a limited Field of View (FOV), which leads to missing lighting in the rendering process.
We address this task by translating it into an inpainting task which infers a panorama image from a limited FOV image.
Furthermore, we aim to infer the surround view image by using a diffusion model~\cite{ho2020denoising,sohl2015deep}.
We proposed to use an image-to-image diffusion model which is a conditional diffusion model that converts samples from a standard Gaussian distribution into samples from a data distribution through an iterative denoising process conditional on an input.
In our task, we adapt an existing model~\cite{saharia2022palette} and make it conditional on the input image to generate a panoramic image.

\textbf{Luminance Distribution Estimation}:
The HDR image reconstruction method proposed in~\cite{shin2023hdr} utilizes a Generative Adversarial Network (GAN) to train encoder-decoder networks that model the sun and sky luminance distribution.
The input is a single outdoor LDR panoramic image and a U-Net~\cite{unet} architecture network with ResNet~\cite{resnet} as its backbone is used to estimate the sky region luminance distribution $L_{sky}$.

Another modified VGG16 network~\cite{vgg} is employed to estimate the sun position probability map $x_{i,j}$ which represents the probability at pixel $(i, j)$ in the input LDR panoramic image containing the sun.
The output feature maps from the CNN blocks the VGG are concatenated together as input fed into the convolutional layers for encoding the sun radiance map which is the Dirac delta function expressed by:
\begin{equation}
    \delta(x_{i,j}, \tau, \beta) = \frac{\tau}{\beta\sqrt{\pi}} exp(-\frac{(1-x_{i,j})^2}{\beta})
    \label{equ:rad}
\end{equation}
where $\tau$ and $\beta$ are the transmittance and sharpness values of the sky.
The sun radiance map is then merged with sun regions to generate the sun region luminance distribution $L_{sun}$.
The $L_{sun}$ and $L_{sky}$ are applied to an inverse tone mapping operation and blended to generate the final output HDR map $L$.

We adapt this method in our lighting estimation module and follow the same process as described in~\cite{shin2023hdr} that uses GAN to re-train the network for generating HDR map $L$.
We then applied $L$ to the inserted object in the video frame.

\begin{figure}[t]
     \centering
     \begin{subfigure}[t]{0.5\textwidth}
         \centering
         \includegraphics[width=\textwidth]{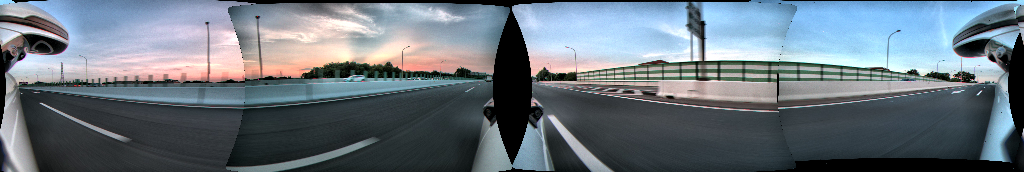}
         \caption{Original Environmental Panoramic Image}
         \label{fig:light-origin}
     \end{subfigure}
     \vfill
     \begin{subfigure}[t]{0.5\textwidth}
         \centering
         \includegraphics[width=\textwidth]{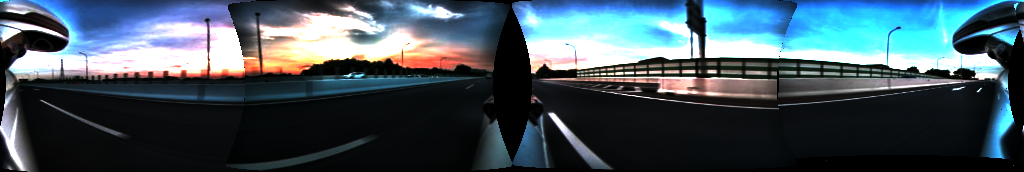}
         \caption{Reconstructed HDR Environmental Panoramic Image}
         \label{fig:light-hdr}
     \end{subfigure}
     \hfill
    \caption{Examples of Original and Reconstructed HDR Environmental Panoramic Image}
    \label{fig:light}
    \vspace{-2.5mm}
\end{figure}

\textbf{Environmental HDR Image Reconstruction:}
As for the outdoor scenario, the sun as the main lighting is not the only one that can affect the visual appearance of the inserted object, we also need to consider the environmental lighting due to the diffuse reflection in order to achieve more realistic rendering outcomes.
To reconstruct the environmental HDR image, we collect multiple side-view LDR images of the scene and recover them into HDR images by using an existing model to learn the continuous exposure value representations~\cite{chen2023learning}.
We followed the same process to estimate the camera extrinsic parameters for each side-view image and stitch them into one HDR panoramic image (Example of the environmental HDR image as shown in Figure~\ref{fig:light}).
Thus we obtained the estimated environmental light distribution from the multiple side-view images, then we can apply it to the inserted object rendering process.

\subsubsection{Object Shadow Generation}
\label{sec:method-render-lighting-shadow}
\begin{figure}[t]
     \centering
     \begin{subfigure}[t]{0.23\textwidth}
         \centering
         \includegraphics[width=\textwidth]{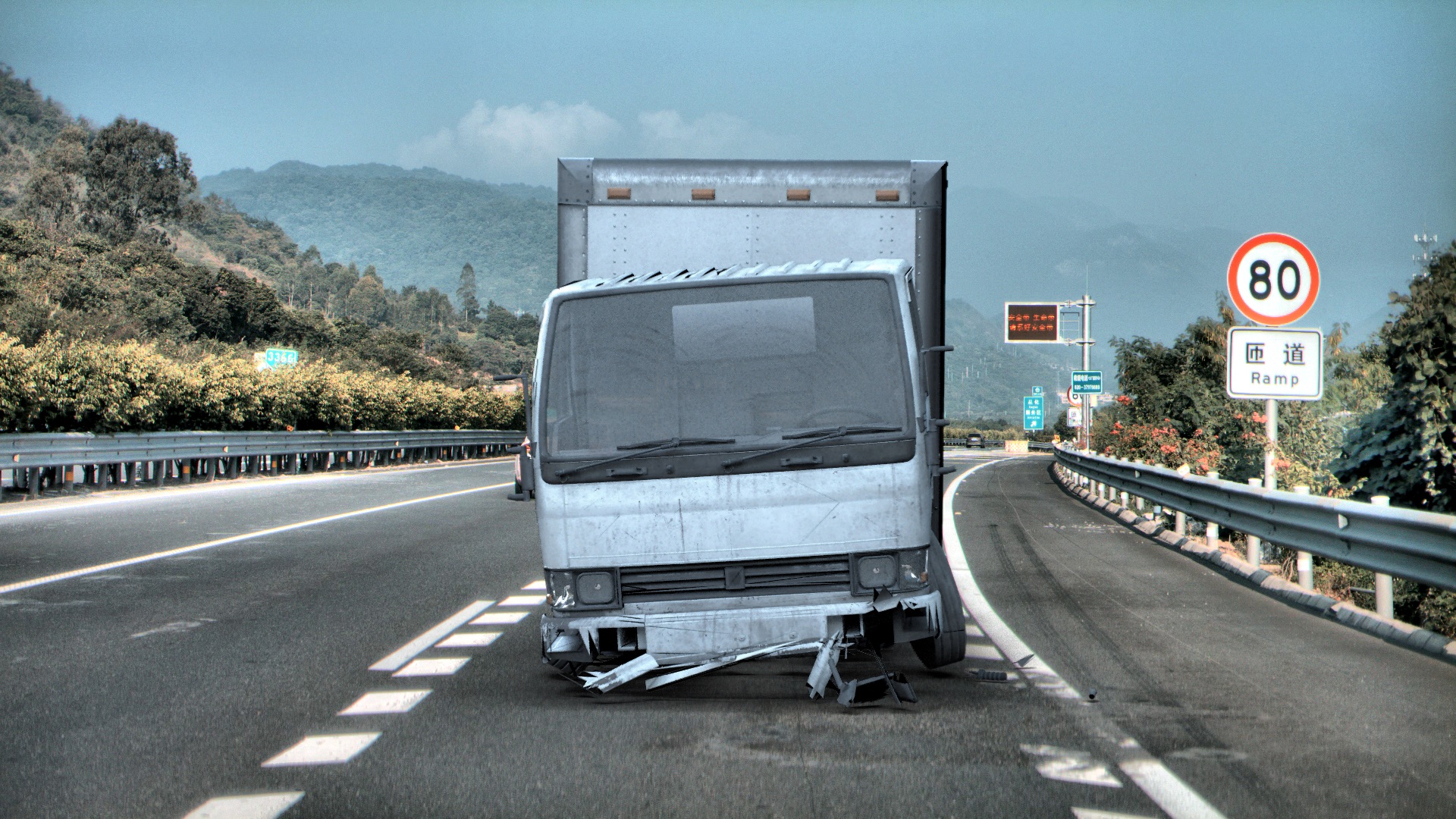}
         \caption{Image with generated shadow}
         \label{fig:shadow-rgb}
     \end{subfigure}
     \hfill
     \begin{subfigure}[t]{0.23\textwidth}
         \centering
         \includegraphics[width=\textwidth]{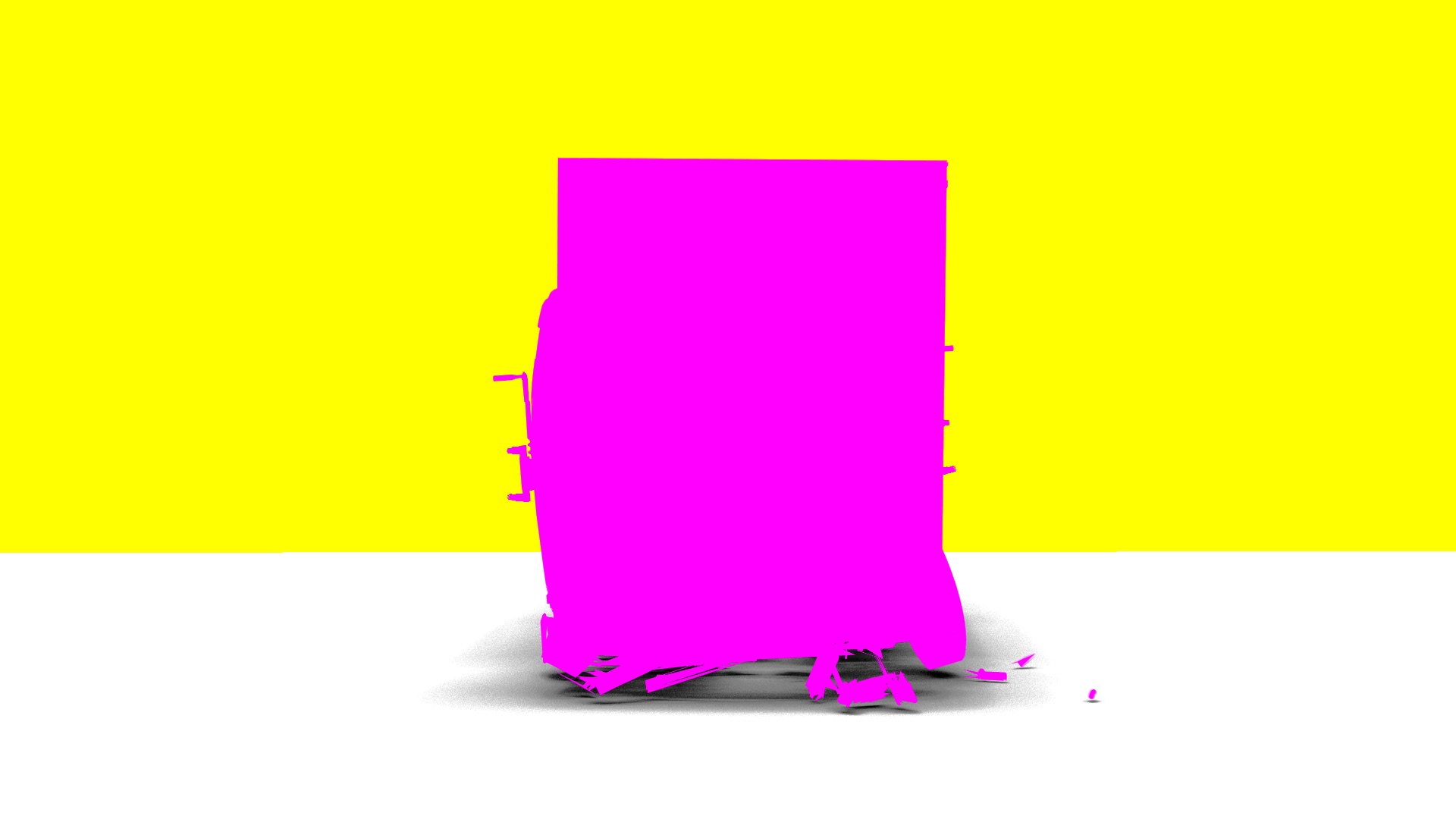}
         \caption{Associated segmentation mask}
         \label{fig:shadow-seg}
     \end{subfigure}
     \hfill
    \caption{Example of generated shadow for the inserted object}
    \label{fig:shadow}
    \vspace{-2.5mm}
\end{figure}

Since we've estimated the location and distribution of the main lighting source, \textit{i.e.} sun for outdoor scene and light for indoor scene, 
we rendered the shadow of the inserted object by the 3D graphics application Vulkan~\cite{vulkan} which offers higher performance and more efficient computing resource usage.
Furthermore, we integrated the ray tracing into the Vulkan application for a better performance of realistic rendering~\cite{raytracing}.
Examples of the generated shadow for the inserted objects are shown in Figure~\ref{fig:shadow}.

%% file: sec/3_2_3_method_style.tex
\subsection{Photorealistic Style Transfer}
\label{sec:method-render-style}
The simulated videos inevitably contain unrealistic artifacts, such as inconsistent illumination and color balancing, which are not included in videos captured in the real-world scenario.
To address this issue, we proposed to use an image inpainting network that faithfully transfers the style to enhance the photorealism of simulated video sequences.

Specifically, we adapt the coarse-to-fine mechanism proposed in~\cite{yu2018generative}, which is originally designated to inpaint missing regions in an image.
We utilized the coarse network and refinement network in~\cite{yu2018generative}, both of them consist of dilated convolution layers to generate the refined image based on the input image.
We modified the input configuration for the two networks.
The coarse network takes an image with black pixels filled in the foreground region, a binary mask indicating the foreground region, and a foreground image of the inserted object with black pixels filled in the background region.
The refinement network takes the same input as the coarse network along with output from the coarse network, and it generates final refined image results.

To train the generative model, we adopt the same training strategy proposed in~\cite{yu2018generative} which uses the WGAN~\cite{arjovsky2017wasserstein} loss, and its objective function can be expressed as:
\begin{equation}
    \min_G\max_{D\in\mathcal{D}} \mathbb{E}_{x \sim\ \mathbb{P}_r}[D(x)] - \mathbb{E}_{\Tilde{x} \sim\ \mathbb{P}_g}[D(\Tilde{x})]
    \label{equ:gan}
\end{equation}
where $\mathcal{D}$ is the set of 1-Lipschitz functions, $\mathbb{P}_r$ is the data distribution and $\mathbb{P}_g$ is the model distribution implicitly defined by $\Tilde{x} = G(z)$, and $z$ is the input to the generator.

We added the gradient penalty term proposed in~\cite{gulrajani2017improved} to improve the WGAN and applied it to pixels in the foreground region.
Thus the penalty function can be expressed as:
\begin{equation}
    \lambda \mathbb{E}_{\hat{x} \sim\ \mathbb{P}_{\hat{x}}} 
    (||\triangledown_{\hat{x}} D(\hat{x}) \odot (1-m)||_2 - 1)^2
    \label{equ:penalty}
\end{equation}
where $\hat{x}$ sampled from the straight line between points
sampled from the distribution $\mathbb{P}_r$ and $\mathbb{P}_g$, and m is the input binary mas of the foreground region.

%% file: sec/4_experiment.tex
\begin{figure*}[t]
    \centering
        \begin{subfigure}[t]{0.24\textwidth}
            \centering
            \includegraphics[width=\textwidth]{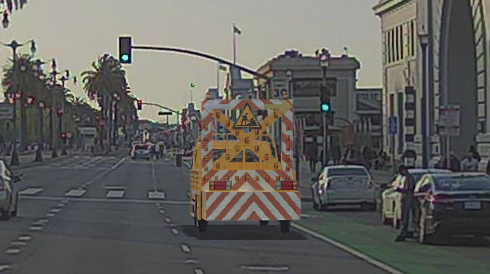} 
        \end{subfigure} 
        \begin{subfigure}[t]{0.24\textwidth}
            \centering
            \includegraphics[width=\textwidth]{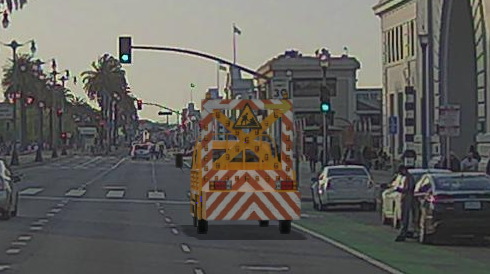} 
        \end{subfigure} 
        \begin{subfigure}[t]{0.24\textwidth}
            \centering
            \includegraphics[width=\textwidth]{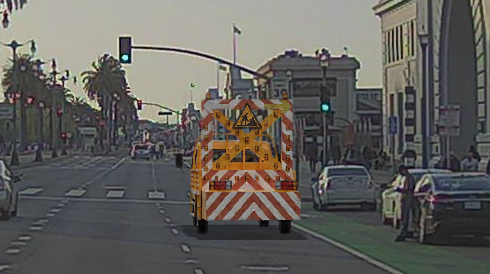} 
        \end{subfigure} 
        \begin{subfigure}[t]{0.24\textwidth}
            \centering
            \includegraphics[width=\textwidth]{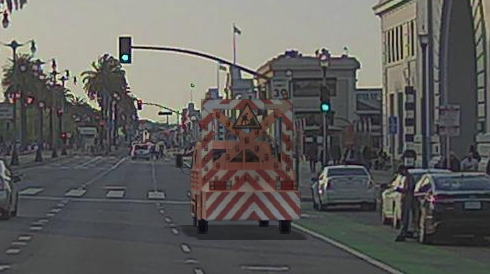} 
        \end{subfigure}
        \begin{subfigure}[t]{0.24\textwidth}
            \centering
            \includegraphics[width=\textwidth]{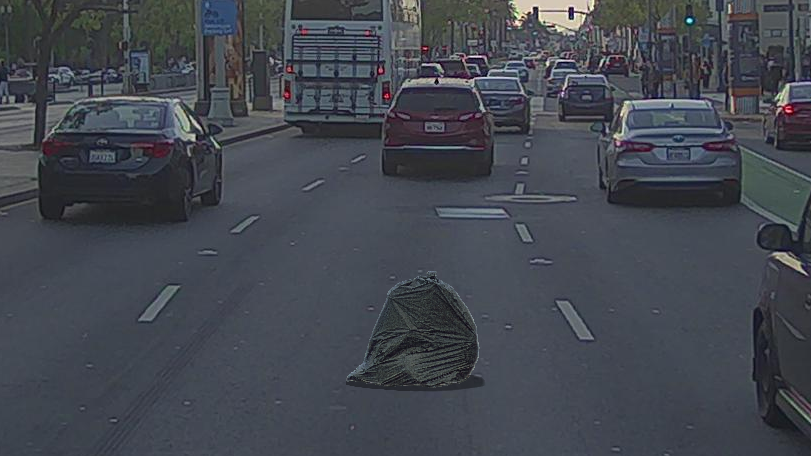} 
        \end{subfigure} 
        \begin{subfigure}[t]{0.24\textwidth}
            \centering
            \includegraphics[width=\textwidth]{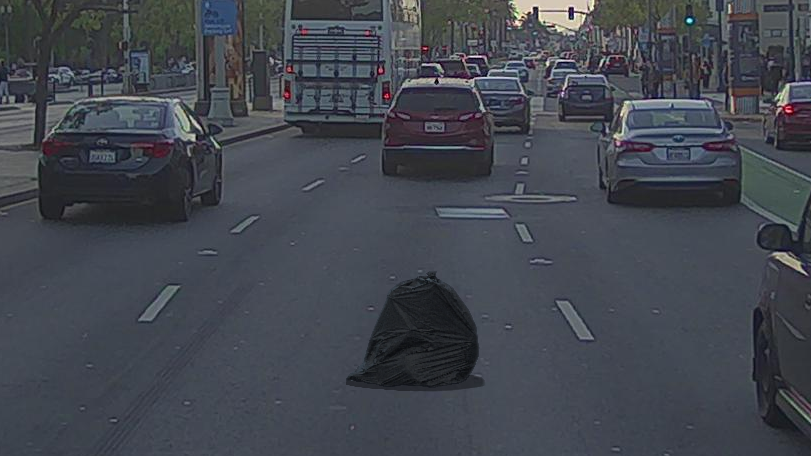} 
        \end{subfigure} 
        \begin{subfigure}[t]{0.24\textwidth}
            \centering
            \includegraphics[width=\textwidth]{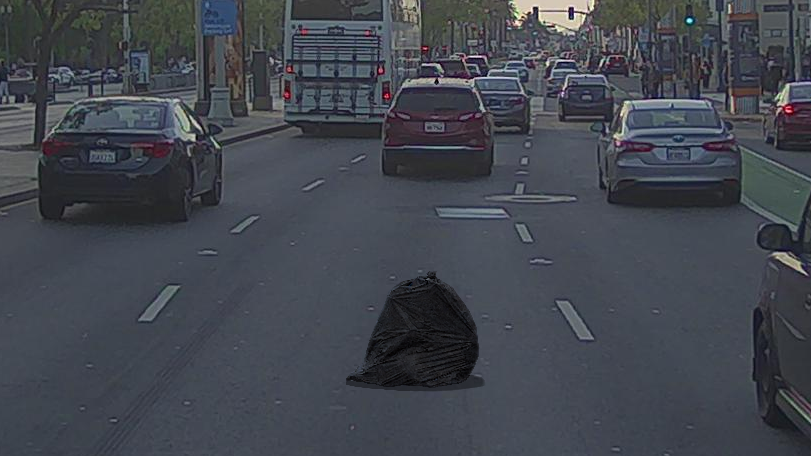} 
        \end{subfigure} 
        \begin{subfigure}[t]{0.24\textwidth}
            \centering
            \includegraphics[width=\textwidth]{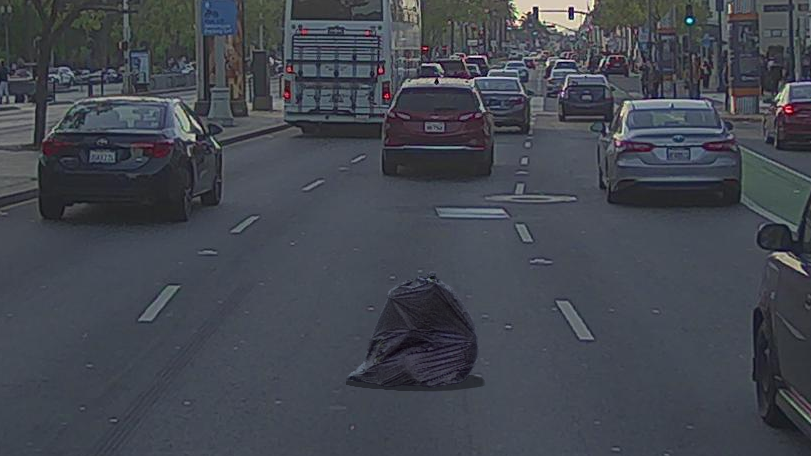} 
        \end{subfigure}
        \begin{subfigure}[t]{0.24\textwidth}
            \centering
            \vspace{-4.5mm}
             \caption{DoveNet}\label{fig:compare-transfer-dovenet}
        \end{subfigure} 
        \begin{subfigure}[t]{0.24\textwidth}
            \centering
            \vspace{-4.5mm}
             \caption{StyTR2}\label{fig:compare-transfer-stytr}
        \end{subfigure} 
        \begin{subfigure}[t]{0.24\textwidth}
            \centering
            \vspace{-4.5mm}
             \caption{PHDiffusion}\label{fig:compare-transfer-diffusion}
        \end{subfigure} 
        \begin{subfigure}[t]{0.24\textwidth}
            \centering
            \vspace{-4.5mm}
             \caption{Ours}\label{fig:compare-transfer-ours}
        \end{subfigure} 
    \caption{Qualitative comparison of the simulated video frame from PandaSet dataset using different style transfer networks.}
    \label{fig:compare-transfer-sample}
    \vspace{-1.0mm}
\end{figure*}
\section{Experimental Evaluation}
\label{sec:experiment}
In this section, we describe the evaluation details of our proposed method for video simulation. 
We introduce evaluation metrics in Section~\ref{sec:experiment-metrics} to quantify performance. 
The video datasets covering both indoor and outdoor scenes used for validation are listed in Section~\ref{sec:experiment-data}. 
We perform an ablation analysis to evaluate the effectiveness of each module of our framework in Section~\ref{sec:experiment-result}. 
Lastly, we showcase the application of the framework in downstream perception tasks in Section~\ref{sec:experiment-downstream}.

\subsection{Evaluation Metrics}
\label{sec:experiment-metrics}
We adopt the following two evaluation metrics used in~\cite{geosim} to assess the quality of simulated videos generated by our proposed framework.
We report the average values for each metric across all video frames in a dataset.

\textbf{Human Score:} This metric measures the percentage of participants who prefer the results from one method over those from the baseline method in a human A/B test.
Detailed descriptions of human study can be found in supplementary materials.
Additionally, the complete set of video pairs and GUI application used in this study is available on our website at~\url{https://anythinginanyscene.github.io}.
We encourage peer researchers to download and review these video comparisons, or to conduct their own human studies for verification of our results.

\textbf{Frechet Inception Distance (FID)}: This metric quantifies the realism and diversity of the generated images by comparing the distribution of generated images with that of groundtruth images.
Lower scores indicate greater similarity, with a zero score implying identical image sets.

\subsection{Evaluation Data}
\label{sec:experiment-data}
To demonstrate the performance of our method for realistic video composition of various scene videos and objects, we validate our method using both outdoor and indoor scene video datasets and diverse inserted object items.

\textbf{Outdoor Scene Video}: PandaSet~\cite{pandaset} is a multi-modal dataset capturing self-driving scenes in various conditions, including different times of day and weather.
We utilized 95 out of all 103 video clips from this dataset, each containing 8 seconds of frames sampled at 10 Hz.

\textbf{Indoor Scene Video}: ScanNet++~\cite{scannet} is a large-scale dataset of indoor scenes created by 3D scanning real environments
The dataset includes DSLR images, RGB-D sequences, and semantic and instance annotations, providing a comprehensive resource for evaluating our methods.
We provide the experimental results of the indoor scene video dataset in the supplementary materials.

\textbf{Object Mesh Assets}: We used the methods introduced in Section~\ref{sec:method-data} to generate 3D object meshes, focusing on various objects, including different types of vehicles and pedestrian models.

\subsection{Experimental Results}
\label{sec:experiment-result}
To assess the performance of various style transfer networks, we compared different methods: a CNN-based method DoveNet~\cite{DoveNet2020}, transformer-based method StyTR2~\cite{deng2022stytr2}, diffusion model-based method PHDiffusion~\cite{lu2023painterly}, and our method introduced in Section~\ref{sec:method-render-style}. 
For the human study, we use our framework without the style transfer module as the baseline for comparison.
We summarize the result of the comparison in Table~\ref{tab:style}.
Our transfer network achieved the lowest FID at 3.730 and the highest human score at 61.11\%, outperforming the alternative methods.

\textbf{Ablation Studies}:
To investigate the effectiveness of each key module, we conducted ablation studies and evaluated the performance.
We removed one module from our framework at a time: placement (w/o placement), HDR image reconstruction (w/o HDR), shadow generation (w/o shadow), and style transfer (w/o style transfer).
In this human study, the w/o style transfer method served as the baseline, and was compared to all other ablation methods. The results are summarized in Table~\ref{tab:ablation}.
The absence of placement, HDR, and style transfer modules resulted in higher FIDs.
Notably, adding shadows significantly enhanced the perceived realism for human observers, though this improvement was not proportionately reflected in the FID score. 
This discrepancy suggests a potential gap between computational assessments of perceptual quality and human judgment, as also noted in previous research~\cite{geosim}.
Our proposed method achieved a human score above 50\%, and the others scored below 50\%, highlighting the contribution of each module in our proposed framework.

\begin{table}[t]
\centering
\begin{tabular}{c  c  c } 
\hline
Method  & \makecell{Human Score(\%)}  &  FID \\ 
\hline
\hline
\textbf{Proposed method} & \textbf{61.11} & \textbf{3.730}  \\ 
\hline
StyTR2 style transfer  & 58.89 & 4.091   \\ 
\hline
PHDiffusion style transfer  & 47.22 & 4.554   \\ 
\hline
DoveNet style transfer  & 47.78 & 3.999   \\ 
\hline
w/o style transfer  & N/A & 4.499   \\ 
\hline

\end{tabular}
\caption{Experimental results for different style transfer networks plugged into our Anything in Any Scene framework.}
\label{tab:style}
\end{table}
\begin{table}[t]
\centering
\begin{tabular}{c  c  c } 
\hline
Method  & \makecell{Human Score(\%)}  &  FID \\ 
\hline
\hline
\textbf{Proposed method} & \textbf{61.11} & \underline{3.730}  \\ 
\hline
w/o placement   & 25.56 &  4.327  \\ 
\hline
w/o HDR  & 43.05 &  3.793  \\ 
\hline
w/o shadow   & 37.78 &  \underline{3.485}  \\ 
\hline
w/o style transfer  & N/A & 4.499   \\ 
\hline

\end{tabular}
\caption{Experimental results for ablation analysis of modules in our Anything in Any scene framework. Note that the baseline w/o style transfer method theoretically has a human score of 50\%}
\label{tab:ablation}
\vspace{-2.5mm}
\end{table}
\begin{figure*}[t]
    \centering
    
        \begin{subfigure}[t]{0.196\textwidth}
            \centering
            \includegraphics[width=\textwidth]{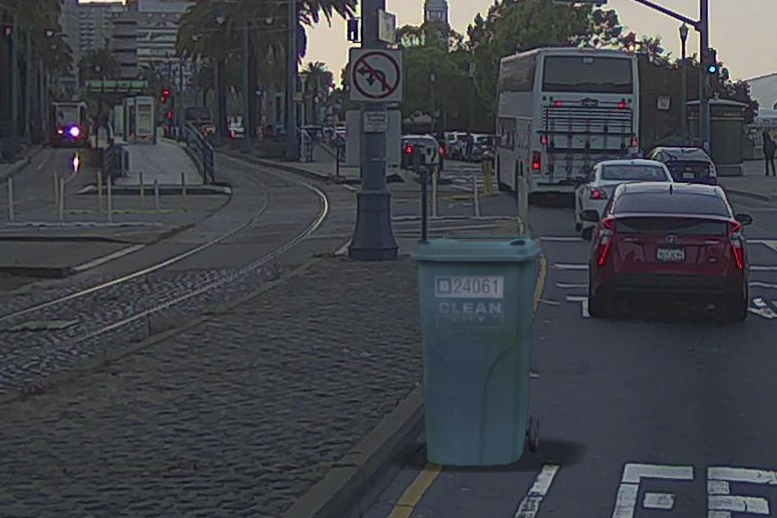} 
        \end{subfigure}    
        \begin{subfigure}[t]{0.196\textwidth}
            \centering
            \includegraphics[width=\textwidth]{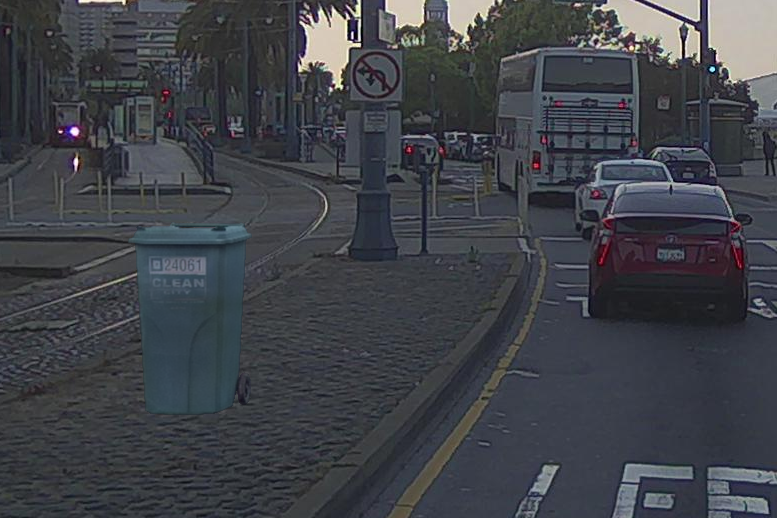} 
        \end{subfigure} 
        \begin{subfigure}[t]{0.196\textwidth}
            \centering
            \includegraphics[width=\textwidth]{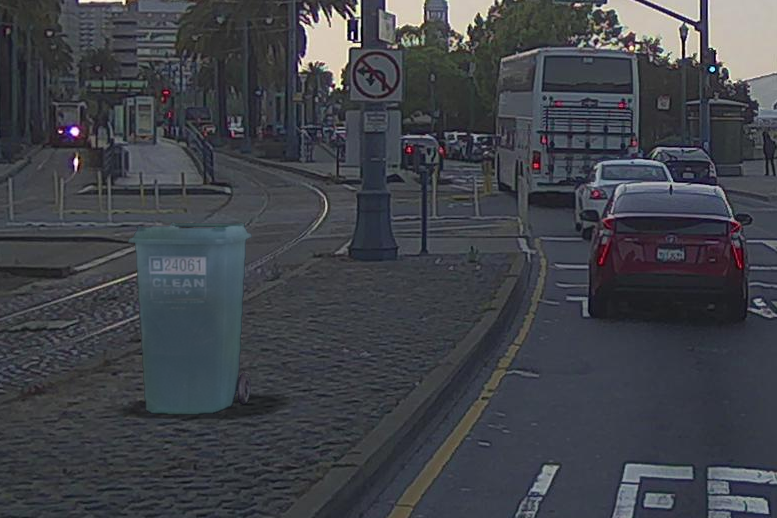} 
        \end{subfigure} 
        \begin{subfigure}[t]{0.196\textwidth}
            \centering
            \includegraphics[width=\textwidth]{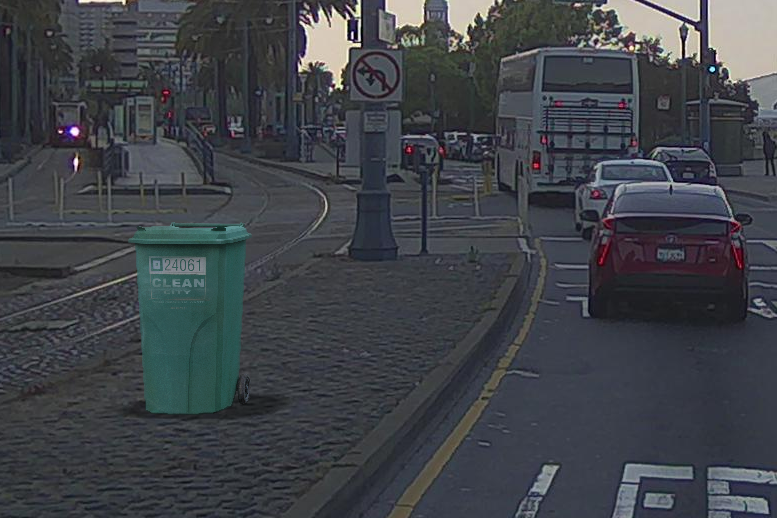} 
        \end{subfigure}
        \begin{subfigure}[t]{0.196\textwidth}
            \centering
            \includegraphics[width=\textwidth]{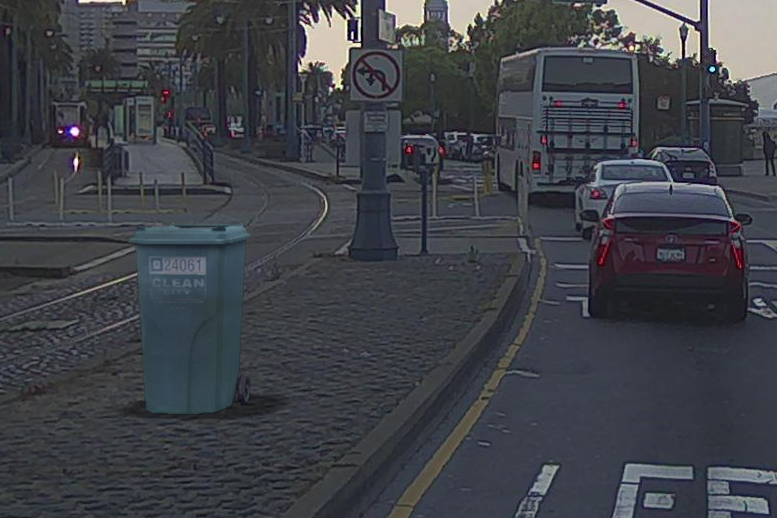} 
        \end{subfigure} 
        \begin{subfigure}[t]{0.196\textwidth}
            \centering
            \includegraphics[width=\textwidth]{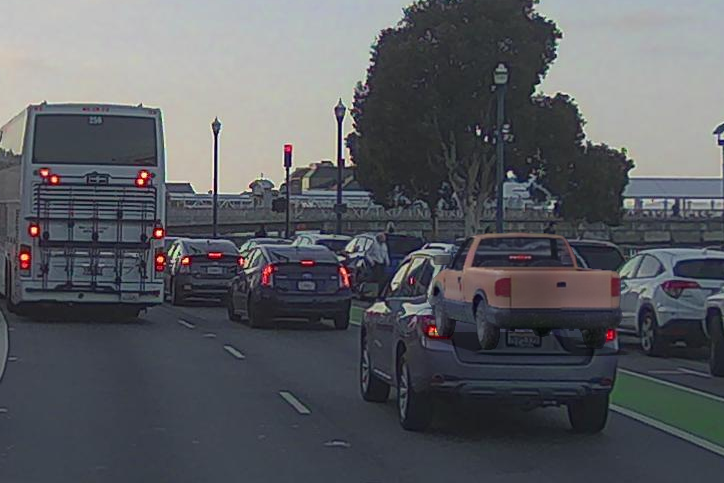} 
        \end{subfigure} 
        \begin{subfigure}[t]{0.196\textwidth}
            \centering
            \includegraphics[width=\textwidth]{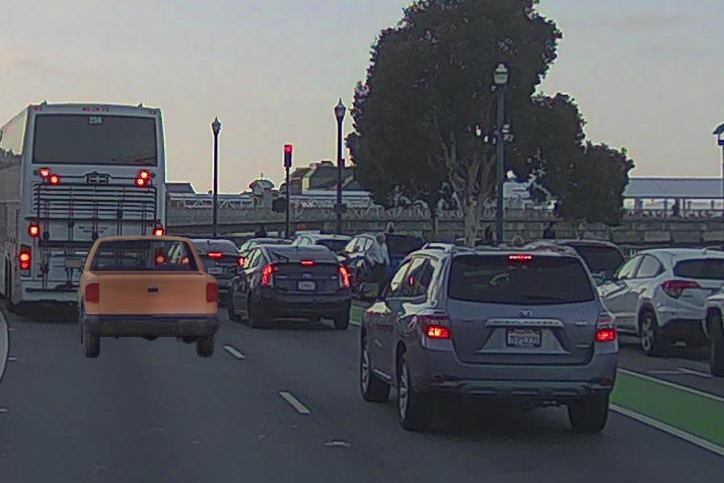} 
        \end{subfigure} 
        \begin{subfigure}[t]{0.196\textwidth}
            \centering
            \includegraphics[width=\textwidth]{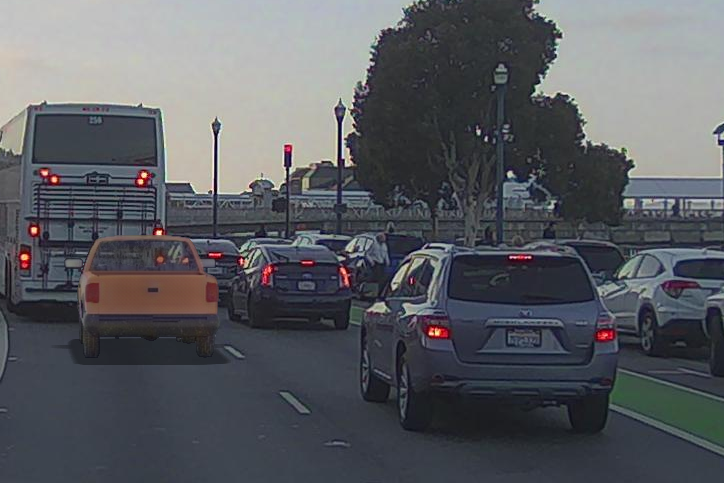} 
        \end{subfigure} 
        \begin{subfigure}[t]{0.196\textwidth}
            \centering
            \includegraphics[width=\textwidth]{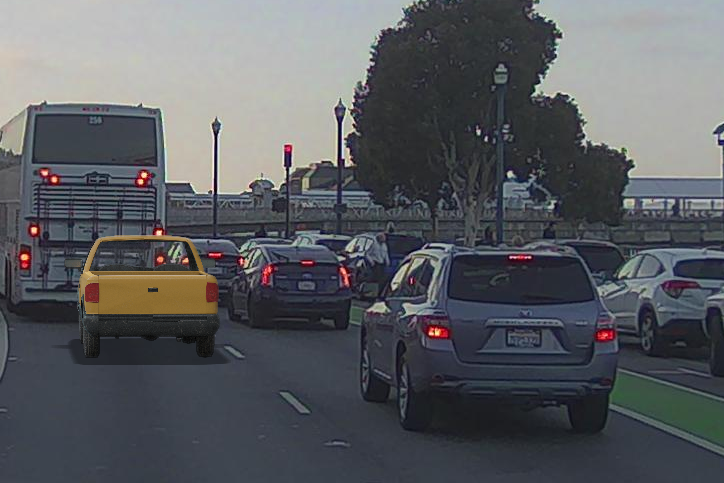} 
        \end{subfigure} 
        \begin{subfigure}[t]{0.196\textwidth}
            \centering
            \includegraphics[width=\textwidth]{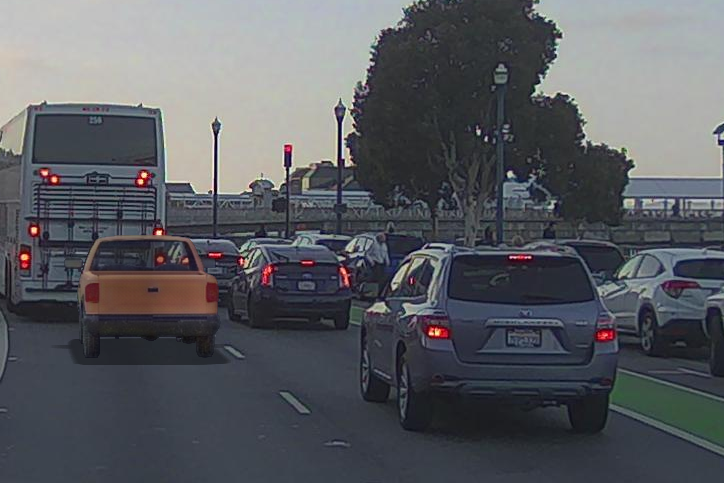} 
        \end{subfigure} 
        \vfill
        \begin{subfigure}[t]{0.19\textwidth}
            \centering
            \vspace{-4.5mm}
             \caption{w/o placement}\label{fig:compare-placement}
        \end{subfigure} 
        \begin{subfigure}[t]{0.19\textwidth}
            \centering
            \vspace{-4.5mm}
             \caption{w/o shadow}\label{fig:compare-shadow}
        \end{subfigure} 
        \begin{subfigure}[t]{0.19\textwidth}
            \centering
            \vspace{-4.5mm}
             \caption{w/o HDR}\label{fig:compare-hdr}
        \end{subfigure} 
        \begin{subfigure}[t]{0.19\textwidth}
            \centering
            \vspace{-4.5mm}
             \caption{w/o style transfer}\label{fig:compare-transfer}
        \end{subfigure} 
        \begin{subfigure}[t]{0.19\textwidth}
            \centering
            \vspace{-4.5mm}
             \caption{Ours}\label{fig:compare-ours}
        \end{subfigure}

    \caption{Qualitative comparison of the simulated video frame from PandaSet dataset under various rendering conditions.}
    \label{fig:compare}
    \vspace{-1.5mm}
\end{figure*}
\textbf{Qualitative comparison}:
In Figure~\ref{fig:compare-transfer-sample}, we provide a qualitative comparison of sample video frames using different style transfer networks applied to the outdoor scene dataset PandaSet.
Figure~\ref{fig:compare-transfer-dovenet}, ~\ref{fig:compare-transfer-stytr} and \ref{fig:compare-transfer-diffusion} show images refined by DoveNet, StyTR2, and PHDiffusion, respectively. 
The inserted object in these images exhibits a color tone that is not consistent with the scene's lighting and weather conditions.
Conversely, the image refined by our proposed method as shown in Figure~\ref{fig:compare-transfer-ours} demonstrates the best visual quality among the four, aligning with the results reported in Table~\ref{tab:style} that show our method outperforming others in both FID and human study scores.
This indicates that an improved style transfer network can significantly enhance photorealism within our Anything in Any Scene framework.

Furthermore, we evaluate the visual quality of videos generated by the Anything in Any Scene framework by removing one module at a time, using the outdoor scene dataset PandaSet as a reference. 
This evaluation is visually illustrated with two comparison samples in Figure~\ref{fig:compare}.
In Figure~\ref{fig:compare-hdr} and Figure~\ref{fig:compare-transfer}, we observe that the inserted object exhibits color textures that are inconsistent with the surrounding environment and other objects in the scene. 
Furthermore, Figure~\ref{fig:compare-shadow} highlights an instance where the inserted object lacks a generated shadow. 
This absence creates a visual effect where the object appears as if it is in the air, highlighting the importance of shadow rendering for realistic simulation.
In contrast, Figure~\ref{fig:compare-ours} shows the visual quality of videos generated by our framework, where the inserted object displays a high degree of consistency with the scene in terms of geometry, lighting, and overall photorealism. 
This demonstrates the capability of the Anything in Any Scene framework to achieve realistic integration of objects into diverse scene settings.

\subsection{Downstream Perception Evaluation}
\label{sec:experiment-downstream}
Real-world datasets often exhibit a long-tailed class distribution, where a few common classes are over-represented, while a majority of classes are under-represented. 
This imbalance poses significant challenges for deep learning models, leading to biases towards common classes during training and worse performance on rare classes during inference.

To address this problem, we investigate the usage of the Anything in Any Scene framework to generate synthetic images containing rare cases for data augmentation.
We perform the evaluation on the CODA dataset~\cite{li2022coda}, an amalgamation of image data from KITTI~\cite{kitti}, nuScenes~\cite{nuscene}, and ONCE~\cite{once} datasets, including 1,500 real-world driving scenes and over 30 object categories

The goal of this task is to insert 9 different rare object categories into images from the CODA2022 validation dataset, with each category comprising less than 0.4\% of total bounding boxes.
We trained three models: YOLOX-S, YOLOX-L, and YOLOX-X~\cite{yolox}, on a subset of 2930 images from the dataset, reserving another 977 images for testing.
We then employed our Anything in Any Scene framework to augment these training images by inserting various objects into them.
This process produced an augmented set of training images that replaced the original ones in the training dataset.
We applied the same training strategy and re-train the models on the augmented training dataset.

We evaluate the performance of the three models by training them on both the original and the augmented datasets, followed by testing them on the same test dataset.
The results, detailed in Table~\ref{tab:downstream}, indicate an improvement in mean Average Precision (mAP) for all three models. 
Specifically, there is an enhancement of 3.7\% in mAP for YOLOX-S, 1.1\% for YOLOX-L, and 2.6\% for YOLOX-X.

\begin{table}[t]
\centering
\begin{tabular}{llcc} 
\hline

Method & Data &  mAP \\
\hline
\hline
\multirow{2}{*}{YOLOX-S} & Original          &  0.186            & \multirow{2}{*}{$0.037\uparrow$}\\  
                        & Original + Ours   &  \textbf{0.223}   & \\ 
\hline
\multirow{2}{*}{YOLOX-L} & Original          &  0.260             & \multirow{2}{*}{$0.011\uparrow$}\\ 
                        & Original + Ours   &  \textbf{0.271}    &\\ 
\hline
\multirow{2}{*}{YOLOX-X} & Original          &  0.249             & \multirow{2}{*}{$0.026\uparrow$}\\ 
                        & Original + Ours   &  \textbf{0.275}\\ 
\hline
\end{tabular}
\caption{Performance of the YOLOX models trained on the original images from the CODA dataset compared to their performance when trained on a combination of original and augmented images using our Anything in Any Scene framework. We report the mAP that represents the mean for all 9 object categories.}
\label{tab:downstream}
\vspace{-5.0mm}
\end{table}

%% file: sec/5_conclusion.tex
\section{Conclusion}
In this work, we proposed an innovative and scalable framework, Anything in Any Scene, designed for realistic video simulation.
Our proposed framework seamlessly integrates a wide range of objects into diverse dynamic videos, ensuring the preservation of geometric realism, lighting realism, and photorealism.
Through extensive demonstrations, we have shown its efficacy in alleviating challenges associated with video data collection and generation, offering a cost-effective and time-efficient solution adaptable to a variety of scenarios.
The application of our framework has shown notable improvements in downstream perception tasks, particularly in addressing the long-tailed distribution issue in object detection.
The flexibility of our framework allows for straightforward integration of improved models for each of its modules, our framework stands as a robust foundation for future explorations and innovations in the field of realistic video simulation.

%% file: sec/X_suppl.tex
\clearpage
\setcounter{page}{1}
\maketitlesupplementary
\global\csname @topnum\endcsname 0
\global\csname @botnum\endcsname 0

In this supplementary material, we include additional technical details and a broader range of quantitative and qualitative results of our proposed method.
We first describe additional details on the assets bank in Section~\ref{sec:suppl-data}, 
the object placement in Section~\ref{sec:suppl-place}, 
the lighting estimation and shadow generation in Section~\ref{sec:suppl-light}, 
and the photorealistic style transfer in Section~\ref{sec:suppl-style}.
We then introduce the details of how we conducted the human study to compare different simulated videos in Section~\ref{sec:suppl-study}.

Furthermore, we also present the quantitative validation results of our method using the indoor dataset ScanNet++ in Section~\ref{sec:suppl-indoor}, and further details on the downstream tasks we conducted are available in Section~\ref{sec:suppl-downstream}.
We provide more details of the result of downstream task we performed in Section~\ref{sec:suppl-downstream}.
To visually underscore the effectiveness of our approach, we include an extensive gallery of simulated videos generated by our framework alongside others for comparative analysis in Section~\ref{sec:suppl-vis}.

Finally, we kindly suggest that reviewers view our supplementary video files (\textit{sample\_video\_outdoor.mp4} for an outdoor scene and \textit{sample\_video\_indoor.mp4} for an indoor scene) to better appreciate the capabilities of our simulation method through these representative examples.

\input{sec/X_suppl_data}
\input{sec/X_suppl_place}
\input{sec/X_suppl_light}
\input{sec/X_suppl_style}
\input{sec/X_suppl_study}
\input{sec/X_suppl_results}
\input{sec/X_suppl_vis}

%% file: sec/X_suppl_data.tex
\begin{figure}
     \centering
     \begin{subfigure}[t]{0.45\textwidth}
         \centering
         \includegraphics[width=\textwidth]{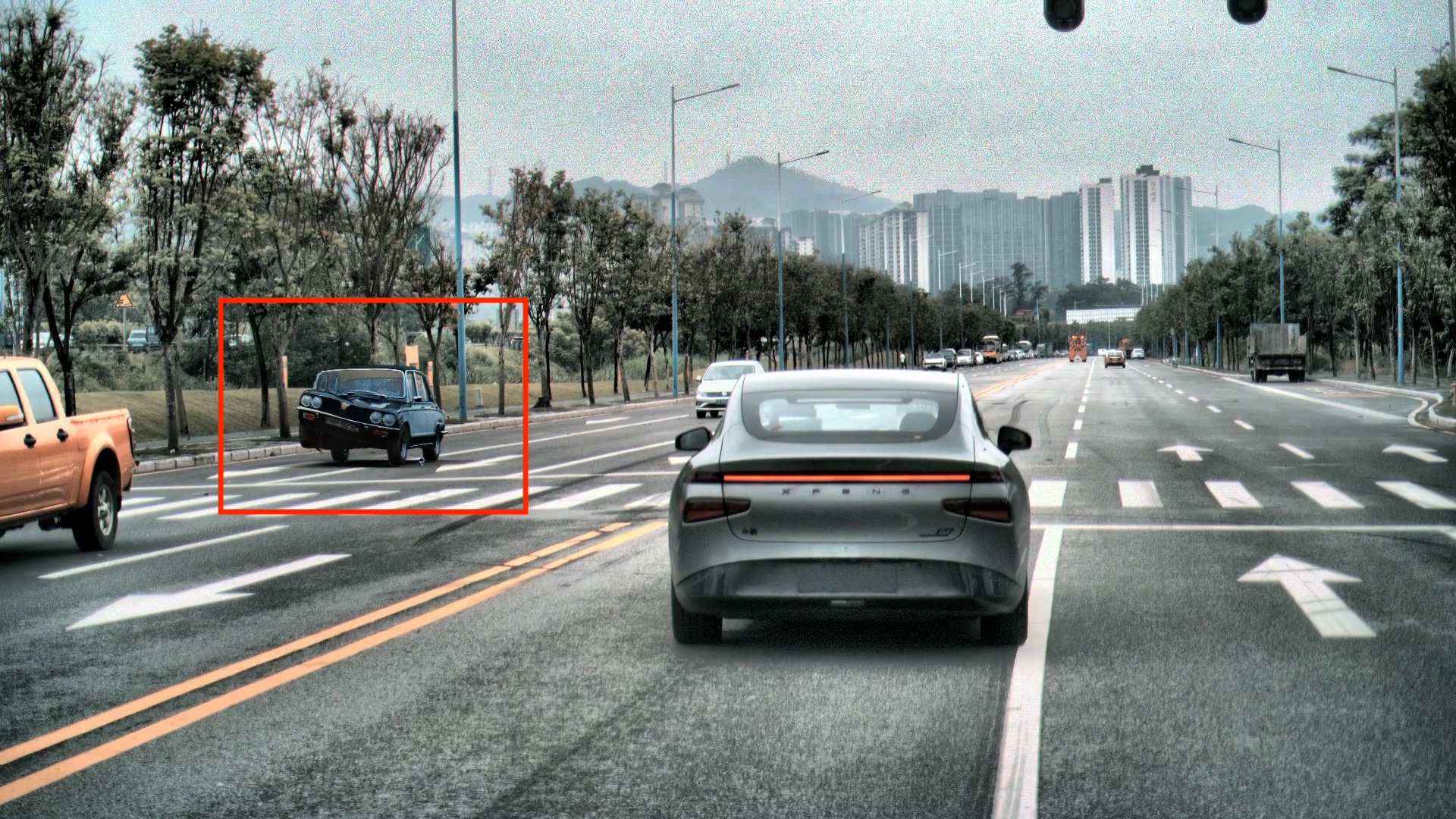}
         \caption{A crashed sedan object mesh generated by Houdini engine}
         \label{fig:data-houdini}
     \end{subfigure}
     \vfill
     \begin{subfigure}[t]{0.45\textwidth}
         \centering
         \includegraphics[width=\textwidth]{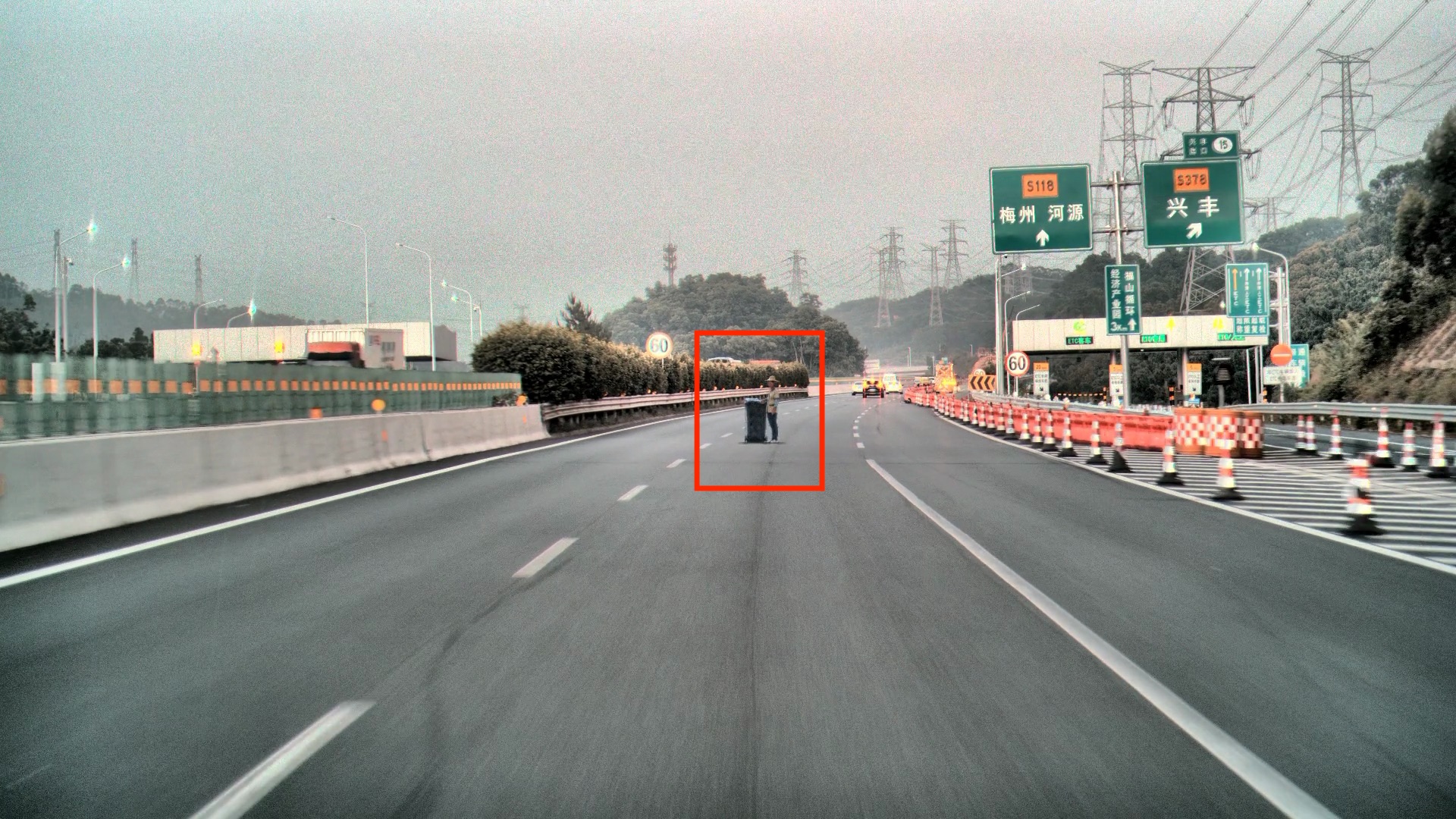}
         \caption{A person object mesh reconstructed by NeRF-based method.}
         \label{fig:data-nerf}
     \end{subfigure}
     \hfill
    \caption{Examples of generated object mesh for video simulation}
    \label{fig:data-mesh}
\end{figure}

\section{Assets Bank Details}
\label{sec:suppl-data}
Our Anything in Any Scene framework aims to create large-scale simulation videos by integrating dynamic scene videos with objects of interest. 
This requires an asset bank of scene videos and object meshes. 
To facilitate this, we developed a visual data query engine for efficiently selecting scene videos based on visual descriptors. 
Additionally, we employ the Houdini Engine and Neural Radiance Fields (NeRF)-based reconstruction for 3D mesh generation, enabling the integration of diverse objects into these videos.

\subsection{Visual Data Query Engine}
In order to efficiently locate target videos for composition from a large-scale video assets bank, our method leverages a visual data query engine.
This engine is designed to retrieve clusters of video clips that visually match the provided descriptive words.
To handle large-scale image and video data with detailed visual features, we employ the Bag of Visual Words (BoVW) approach.

We first estimate semantic segmentation masks for each frame in the scene video assets bank. 
This segmentation breaks down each video frame into labeled regions of interest. 
Following this, we utilize the Scale Invariant Feature Transform (SIFT) algorithm to extract visual features from these segmented regions.
We detect key points in each frame and compute descriptors represented by feature vectors for the regions containing these key points. 
These descriptors are then clustered, with the centroid of each cluster representing a 'visual word' in the BoVW. 
The frequency of these visual words across the video dataset is used to build a frequency histogram for each video.
Consequently, the BoVW representation allows us to effectively retrieve matching videos based on the occurrence and frequency of the given visual words, improving the process of selecting appropriate videos for our Anything in Any Scene framework.

\subsection{Object Mesh Generation}
The mesh model of a target object is required before its insertion into an existing video clip.
We employ the following two methods to generate the object mesh models.

\textbf{Houdini Engine for Object Mesh Generation}
To create visually appealing and physically accurate object meshes, we utilize the Houdini Engine~\cite{houdini} that leverages the physics-based rendering capabilities to enhance existing object mesh models with realistic physical effects
The Houdini Engine, known as a robust 3D animation procedural tool, can produce a wide range of physical effects such as deformation, animations, reflections, and particle visual effects.
As an example is shown in Figure~\ref{fig:data-houdini}, the Houdini engine can transform a truck model into a crashed one by applying deformation effects.
Furthermore, it can simulate diverse realistic physical effects, such as smoke from a crashed car, using its particle visual system. 
This approach is particularly critical for creating object meshes that are challenging or expensive to capture in real-world scenarios. 

\textbf{NeRF-based Object Mesh Reconstruction} 
In order to also cover the objects that are difficult to produce by the Houdini engine and generalize the asset bank to include arbitrary objects, we propose the complementary NeRF-based Object Mesh Reconstruction
The impressive performance of Neural Radiance Fields (NeRF) in 3D reconstruction from multi-view images offers the potential to build an extensive 3D asset bank.
In our work, we adopt an off-the-shelf method~\cite{tang2023delicate} that combines the advantage of both NeRF and mesh representation
This method reconstructs the object mesh model from multi-view RGB images.
An example of the reconstructed person object mesh is shown in Figure~\ref{fig:data-nerf}, which features rich textures and detailed geometry suitable for following rendering processes.

%% file: sec/X_suppl_place.tex
\section{Object Placement Details}
\label{sec:suppl-place}
In order to accurately position the inserted object within a scene video, the first step involves reconstructing the 3D point cloud representation of the captured environment. 
The object placement point is then determined in 3D space, guided by segmentation mask. 
During the 2D-to-3D projection process, we focus on estimating an appropriate placement plane for the inserted object. 
This plane is conceptualized as the best-fitting plane, represented by the equation:
\begin{equation}
    Ax+By+Cz+D=0
\end{equation}
based on the selected points $(x,y,z)$.

\begin{figure}[t]
    \centering
    \includegraphics[width=0.45\textwidth]{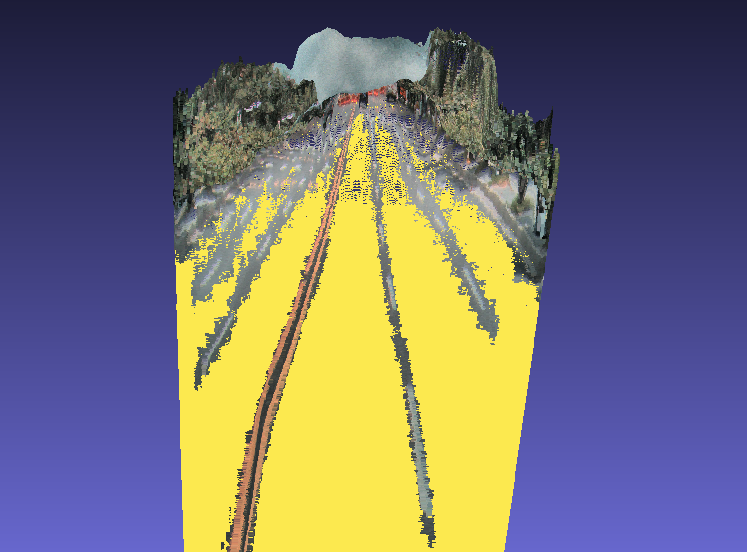}
    \caption{The projected 3D scene for object insertion. The yellow region is the plane that is available to place the inserted object.}
    \label{fig:suppl-place-ground}
\vspace{-0.3cm}
\end{figure}

For a more accurate estimation, we utilize multiple 3D points to determine the optimal fitting plane. 
As illustrated in Figure~\ref{fig:suppl-place-ground}, we select several points within the road region (in yellow in the Figure~\ref{fig:suppl-place-ground}) to estimate the ground plane where the object can be realistically inserted.

\begin{figure}[t]
    \centering
    \includegraphics[width=0.45\textwidth]{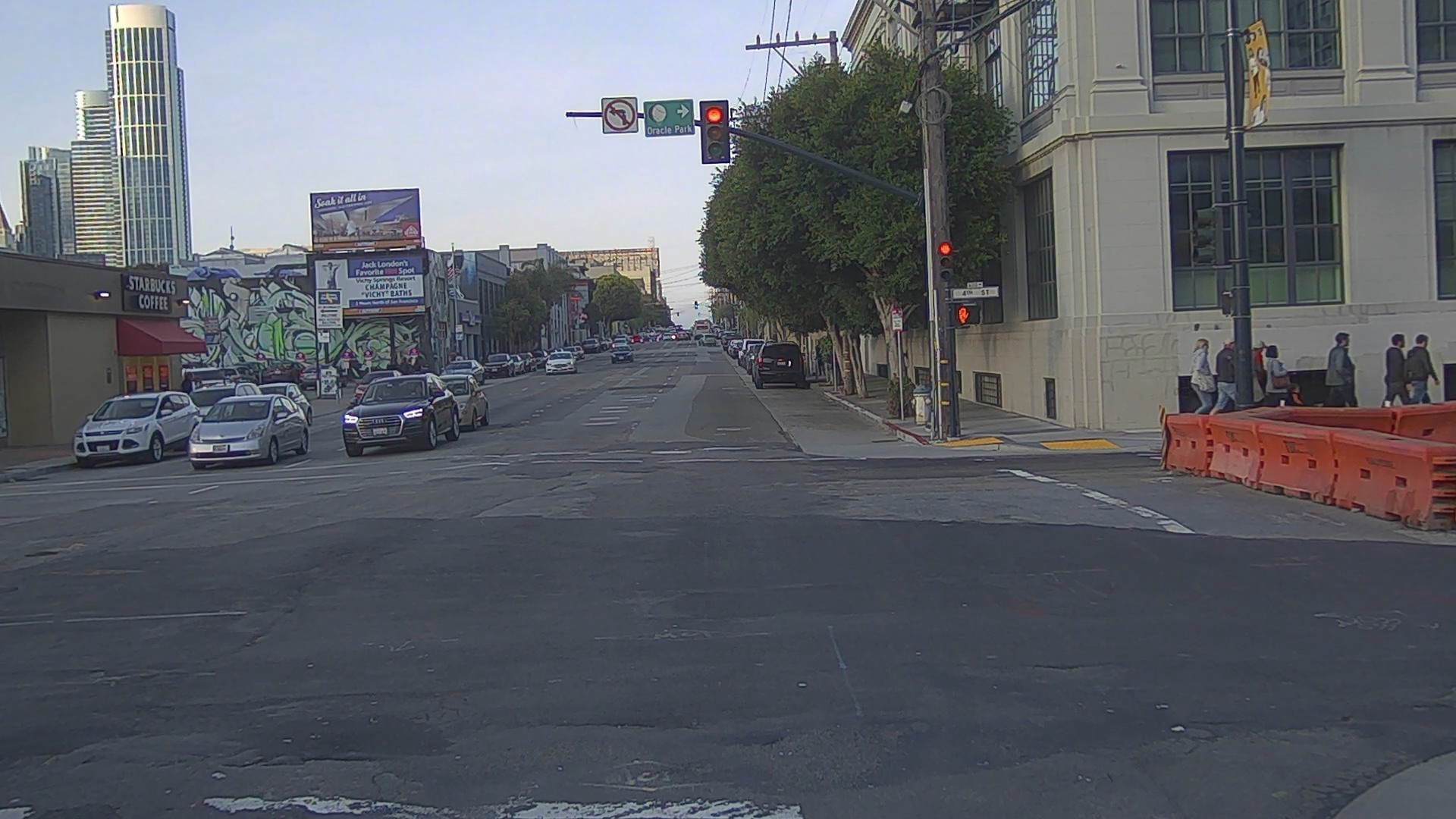}
    \caption{An example of an excluded video clip from the PandaSet. The camera is stable during the entire video clip because the camera is on a car waiting for the traffic light.}
    \label{fig:suppl-place-exclude}
\end{figure}

\textbf{Settings for PandaSet and ScanNet++ Datasets}:
The two datasets we used, PandaSet~\cite{pandaset} for outdoor scene and ScanNet++\cite{scannet} for indoor scene, consist of footage captured by RGB cameras and depth sensors.
These sensors record driving scenes for PandaSet and room scenes for ScanNet++. 
Our selection process for video clips from these datasets involves choosing those captured by a forward-facing camera, particularly focusing on clips where the camera exhibits motion, thus ensuring dynamic and varied frames for composition. 
Regarding the ScanNet++ dataset, we selected 5-second segments from each original video clip, down-sampling them from the original 60Hz to 20Hz.
We ensured that a minimum of 20 frames from each clip were available, providing an adequate number of frames for effective video composition.
We exclude video clips that suffer from low frame rate issues or where the camera remains stationary during the recording, such as the scenario shown in Figure\ref{fig:suppl-place-exclude}, where the camera is affixed to a stationary vehicle at a traffic signal.

Utilizing the depth information and segmentation data available in both dataset, we reconstruct the 3D point cloud for the scenes.
This enables us to precisely select the object placement points from within the designated placeable areas in 3D space.
For instance, in a driving scene from the PandaSet, as illustrated in Figure~\ref{fig:suppl-place-ground}, the road region highlighted in yellow is identified as the appropriate location for inserting a car into the scene.


%% file: sec/X_suppl_light.tex
\section{Lighting Estimation Detail}
\label{sec:suppl-light}
\begin{figure*}[t]
     \centering
     \includegraphics[width=\textwidth]{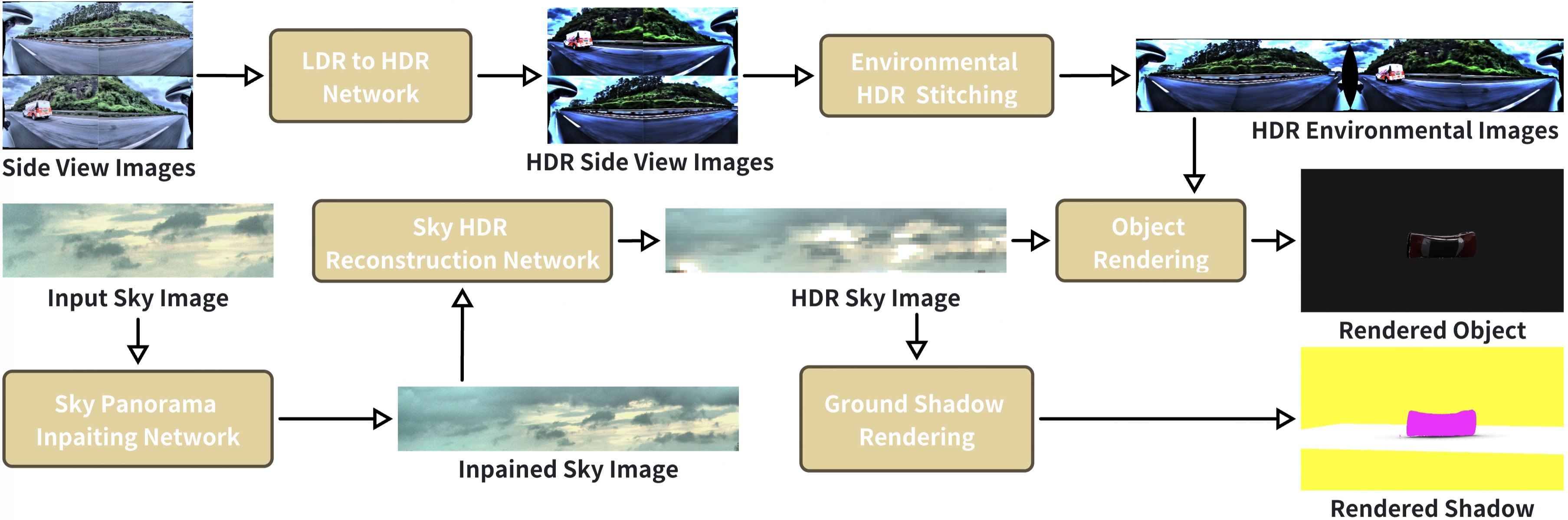}
    \caption{The overview of lighting estimation and shadow generation.}
    \label{fig:light-overview}
\end{figure*}

In Figure~\ref{fig:light-overview}, we provide a comprehensive overview of the lighting estimation and shadow generation process.
To ensure realistic shading effects on objects inserted during rendering, we estimate High Dynamic Range (HDR) images of both the sky and the surrounding environment.

For HDR sky image estimation, an image inpainting network initially infers a panoramic sky image. 
This is followed by employing a sky HDR reconstruction network to transform this panoramic sky image into an HDR one. 
In parallel, the estimation of HDR environmental images involves reconstructing HDR images from Low Dynamic Range (LDR) side-view images of the scene by using an LDR to HDR network. 
These images are then seamlessly stitched together to form an HDR panoramic environmental image.

Both the HDR sky and environmental images are integrated together to achieve realistic lighting effects on the inserted objects in the rendering process. 
Additionally, we leverage the estimated HDR sky image to render shadows for the inserted objects, utilizing the 3D graphics application Vulkan for this purpose.

%% file: sec/X_suppl_style.tex
\section{Photorealistic Style Transfer Detail}
\label{sec:suppl-style}
\begin{figure*}[t]
     \centering
     \includegraphics[width=\textwidth]{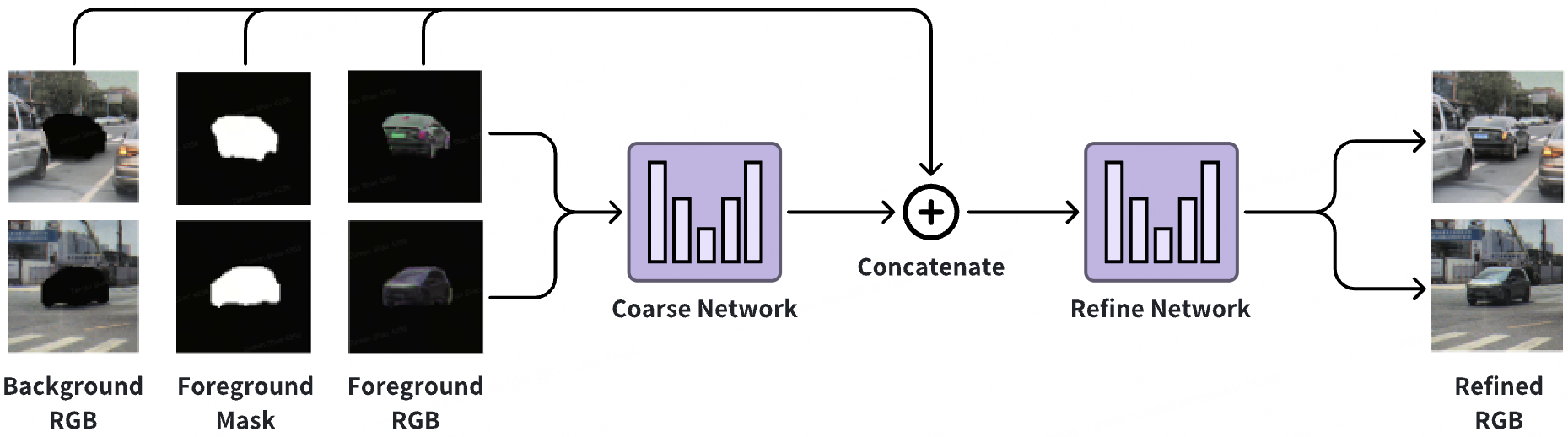}
    \caption{The overview of coarse-to-fine mechanism for photorealistic style transfer. The input is a background RGB image with a black foreground region, the inserted object foreground RGB image, and a foreground segmentation mask. The output is the refined RGB image.}
    \label{fig:refine-net}
\end{figure*}

We utilize the coarse-to-fine mechanism for photorealistic style transfer, and the overview of the network is shown in Figure~\ref{fig:refine-net} where both of the coarse network and refinement network consist of the dilated convolution layers.
We concatenate an image with black pixels filled in the foreground region, a binary mask indicating the foreground region, and an image with black pixels filled in the background region as an input to the coarse network that outputs an initial coarse prediction. 
The refine network takes the composition of the coarse network's input and output, and it generates the final refined completed image.

We followed the same training strategy as described in~\cite{yu2018generative}, the coarse network is trained with the reconstruction loss, and the refinement network is trained with the reconstruction and GAN losses.
We trained and finetuned the networks on the PandaSet dataset for the outdoor scenario.
All input is concatenated together and then resized $256 \times 256$ as input image resolution.

We are also interested in the performance of different style transfer methods on the task of photorealistic style improvement in our proposed framework.
Specifically, we investigated the usage of a CNN-based method DoveNet~\cite{DoveNet2020}, a transformer-based method StyTR2~\cite{deng2022stytr2}, and a diffusion model-based method PHDiffusion~\cite{lu2023painterly} compared to our method introduced in main paper.

\textbf{DoveNet}: a U-Net-based network is used as a generator to translate the visual domain of the inserted foreground region to match the background, and the GAN framework with two different discriminators is leveraged to train the generator for more realistic and harmonious image output.
We follow the same process as described in~\cite{DoveNet2020}, we resize the input images as $256\times 256$ during both the training and testing stages.

\textbf{StyTR2}: a transformer is leveraged as an encoder to capture long-range dependencies of image features for style transfer.
We set the style weight as $10.0$ and content weight as $7.0$ for the StyTR2 model, and we downscale the image to $512\times 512$ and then randomly crop the image to $256\time 256$ as the input image during the training stage.

\textbf{PHDiffusion}: a stable diffusion model is proposed to use two encoders to stylize foreground features. The features from both encoders are combined to finalize the style transfer process.
We loaded the pre-trained Stable Diffusion model weights and all images are resized to $512\times 512$ as input resolution for both training and testing.

%% file: sec/X_suppl_study.tex
\section{Human Study Details}
\label{sec:suppl-study}
\begin{figure*}[t]
    \centering
    \includegraphics[width=\textwidth]{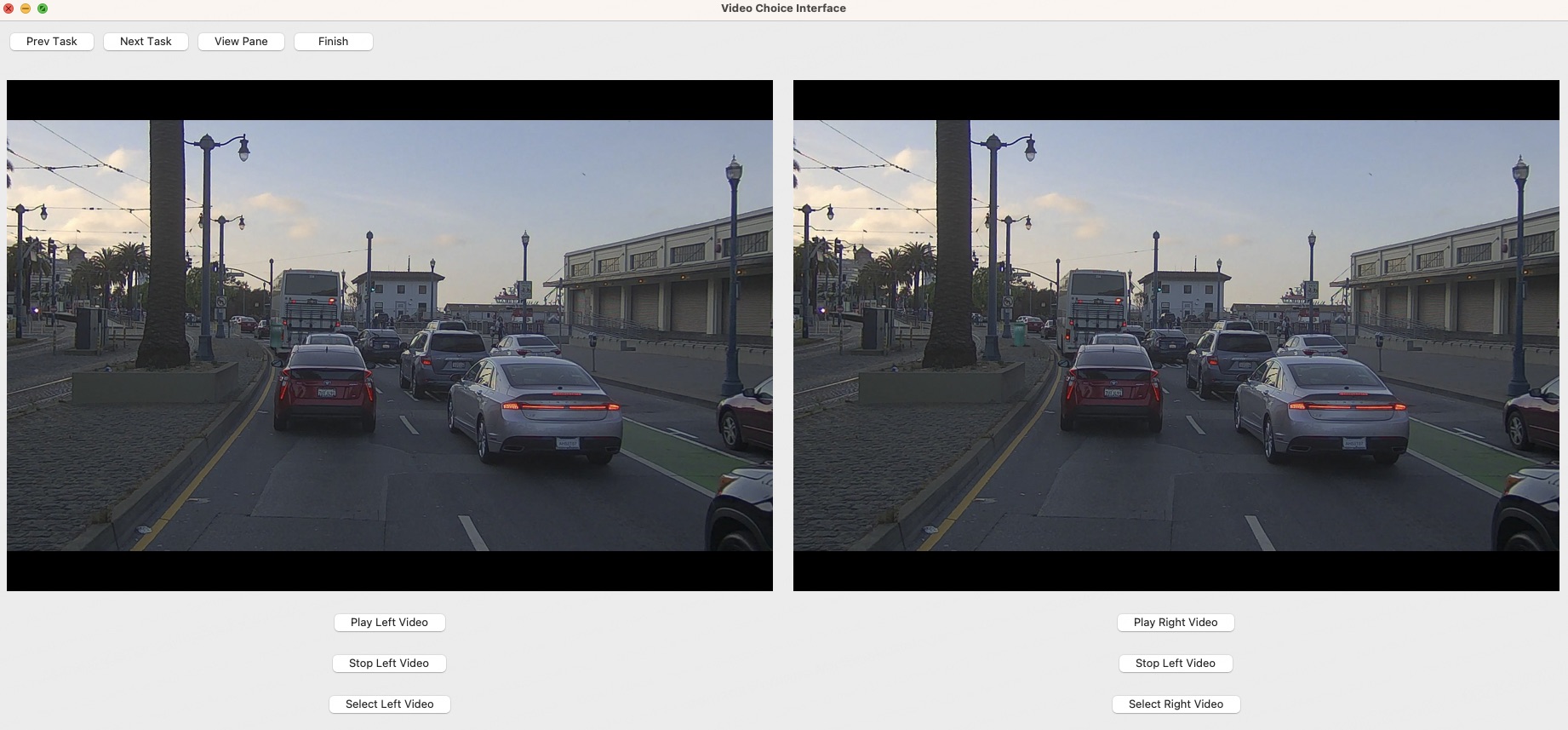}
    \caption{Example of the human study interface for comparing two videos quality. The human judge selects the right video because of its more realistic visual effect.}
    \label{fig:suppl-study}
\end{figure*}

We validated the simulated videos generated by our proposed method through a human A/B test.
We utilized a GUI application~\cite{selector} designed and developed by ourselves, which allows users to compare two videos side by side, and select the preferred one between them. 
We provide instruction as follows to each human judge before they start to conduct the study:

\textit{
You are participating in a study to assess the realism of videos created by computers. 
Each video features a scene with an object inserted. 
Your task is to compare two videos side by side and select the one that appears more realistic to you.\\
Please ensure that you are seated at an appropriate distance in front of the display screen, and familiarize yourself with controls, such as playing the video and going to the previous or next task.\\
For this study, realism is defined by how convincingly the object is integrated into the scene video.
You can consider the following factors in your assessment:
\begin{enumerate}
    \item The consistency of the object with physical rules depicted in the scene.
    \item The naturalness of lighting, shadows, and replications.
    \item The believability of the object's interaction with its environment.
\end{enumerate}
Watch both videos in full at least once before making a decision, and feel free to view as many times as needed, focusing on the inserted object in the scene.
Please select the video that you believe has better realism by pressing the corresponding "select" button.
}\\
The human judges use the application as shown in Figure~\ref{fig:suppl-study}, which provides multiple controls, such as navigation through all video pairs, video playback, video selection, video suspend, and selection view panel.

For the validation of each dataset, we conducted two separate human studies. 
The study for the outdoor dataset involved 24 human judges, while the study for the indoor dataset had 16 participants. 
In validating the PandaSet dataset, we had a pool of 100 videos, from which 38 were randomly selected for the human study. 
In the case of the ScanNet++ dataset, out of the 52 available videos, 30 were randomly chosen for conducting the human study. Note that all videos were used in the calculation of the FID score.
In each study, every judge was tasked with evaluating and labeling a total of 105 pairs of videos.
In the first study, we compare the performance of different style transfer networks plugged into our proposed framework, covering DoveNet, StyTR2, PHDiffusion, and ours.
As for another human study, we analyze the performance of our proposed method if we remove one of the key components.
In both study settings, we set our proposed method without a style transfer process as the baseline.

The human score of method A can be computed as
\begin{equation}
     \frac{\text{times of results by method A selected}}{\text{total times of results by method A and B selected}}
\end{equation}
where method B is the baseline, and method A is the method for comparison.
Suppose method A is also the base, theoretically baseline method has a human score of 50\%.
We provide the quantitative and qualitative results on the outdoor dataset PandaSet in the main paper, and the results on the indoor dataset ScanNet+ are detailed in Section~\ref{sec:suppl-vis}.

%% file: sec/X_suppl_results.tex
\begin{table}[t]
\centering
\begin{tabular}{c  c  c } 
\hline
Method  & \makecell{Human Score(\%)}  &  FID \\ 
\hline
\hline
\textbf{Proposed method} & \textbf{57.92} & \textbf{10.537}  \\ 
\hline
StyTR2 style transfer  & 53.33 & 11.145   \\ 
\hline
PHDiffusion style transfer  & 36.25 & 12.004   \\ 
\hline
DoveNet style transfer  & 44.58 & 10.832   \\ 
\hline
w/o style transfer  & N/A & 11.901   \\ 
\hline
\end{tabular}
\caption{Experimental results of indoor Scene dataset ScanNet+ with different style transfer networks plugged into our Anything in Any Scene framework.}
\label{tab:suppl-style}
\end{table}

\begin{table}[t]
\centering
\begin{tabular}{c  c  c } 
\hline
Method  & \makecell{Human Score(\%)}  &  FID \\ 
\hline
\hline
Proposed method & 57.92 & 10.537  \\ 
\hline
w/o placement   & 9.58 &  9.709  \\ 
\hline
w/o HDR  & 32.92 &  10.824  \\ 
\hline
w/o shadow   & 36.25 &  10.464  \\ 
\hline
w/o style transfer  & N/A & 11.901   \\ 
\hline

\end{tabular}
\caption{Experimental results for ablation analysis of modules in our Anything in Any scene framework. Note that the baseline w/o style transfer method theoretically has a human score of 50\%}
\vspace{-5.0mm}
\label{tab:suppl-ablation} 
\end{table}

\begin{table*}
\centering
\begin{tabular}{llcccccc} 
Method & Data &  \textbf{mAP} & Plastic Bag & Stone & Stroller & Traffic Light & Concrete Block \\
\hline
\hline
\multirow{2}{*}{YOLOX-S}& Original          &  0.321            & 0.302             & 0.020             & 0.218             & 0.193          & \textbf{0.125}\\  
                        & Original + Ours   &  \textbf{0.334}   & \textbf{0.475}    & \textbf{0.093}    & \textbf{0.260}    & \textbf{0.227} & 0.108\\ 
\hline
\multirow{2}{*}{YOLOX-L}& Original          &  \textbf{0.394}   & 0.314             & \textbf{0.105}    & 0.406             & 0.262          & 0.178\\ 
                        & Original + Ours   &  0.391            & \textbf{0.336}    & 0.077             & \textbf{0.474}    & \textbf{0.318} & \textbf{0.309}\\ 
\hline
\multirow{2}{*}{YOLOX-X}& Original          &  0.395            & \textbf{0.319}    & 0.110             & 0.307             & 0.246          & \textbf{0.215}\\ 
                        & Original + Ours   &  \textbf{0.405}   & 0.311             & \textbf{0.133}    & \textbf{0.529}    & \textbf{0.290} & 0.202\\ 
\hline
\end{tabular}
\caption{Performance of the YOLOX models trained on the original images from the CODA dataset compared to their performance when trained on a combination of original and augmented images using our Anything in Any Scene framework. We report the mAP represents the mean for all 25 object categories. We also report individual categories that has a significant improved AP in either one of the three models.}
\vspace{-5.0mm}
\label{tab:suppl-downstream}
\end{table*}

\section{Experimental Results of Indoor Scene}
\label{sec:suppl-indoor}
We followed the same experimental setup detailed in the main paper, and conducted a validation of our proposed method on an indoor scene.
Similarly to the validation on the outdoor scene, we evaluated the performance of various style transfer networks by comparing the following methods: a CNN-based method DoveNet, transformer-based method StyTR2, diffusion model-based method PHDiffusion, and our method.
In the human study, as described in~\ref{fig:suppl-study}, we designated our framework without the style transfer module as the baseline for comparison. 
The comparative results, summarized in Table~\ref{tab:suppl-style}, reveal that our style transfer network achieved the lowest Frechet Inception Distance (FID) score of 10.537 and the highest human score of 57.92\%, surpassing the performance of the other methods.

\textbf{Ablation Studies}:
We also performed ablation studies using indoor dataset ScanNet+ to assess the impact of individual modules on overall performance. 
Similarly. we removed one module from our framework: placement (w/o placement), HDR image reconstruction (w/o HDR), and style transfer (w/o style transfer). 
The results are detailed in Table~\ref{tab:suppl-ablation}. 
Similarly to the result of outdoor dataset PandaSet, the absence of HDR, and style transfer modules resulted in higher FIDs. 
The placement of objects in unrealistic locations significantly reduced their perceived realism among human observers. 
However, this decrease in realism was not accurately reflected in the FID scores. 
One primary reason is the nature of indoor scenes, which often have limited space.
This can result in the inserted object being partially or completely out of the camera’s field of view, impacting the assessment metrics.
Our method consistently received a human score above 50\%, while the others fell below 50\%, emphasizing the contribution of each module to the efficacy of our system.

\section{Downstream Task Details}
\label{sec:suppl-downstream}
We expanded our scope to include 25 object categories for insertion into images from the CODA2022 validation dataset. 
Similarly, we trained three models: YOLOX-S, YOLOX-L, and YOLOX-X~\cite{yolox}. 
Utilizing the "Anything in Any Scene" framework, we augmented the original training images by inserting a variety of objects, thereby generating a new set of training data.
This enhanced dataset was then used to re-train the models, ensuring consistency with the original training strategy.

The performance of the models was evaluated by training on both the original and augmented datasets and then testing on a consistent test dataset. 
The results are presented in Table~\ref{tab:suppl-downstream}, where we detail the mean Average Precision (mAP) across all 25 object categories. 
We also highlight individual categories that showed significant AP improvement in any of the three models.
    

%% file: sec/X_suppl_vis.tex
\section{Qualitative Visualization}
\label{sec:suppl-vis}

The quantitative experimental results show that human judges prefer our proposed method compared to the other which either has one key component missing or another photorealistic style transfer network used. 
We demonstrate more qualitative visualization for both outdoor dataset PandaSet and indoor dataset ScanNet+ as shown in the following.

\begin{figure*}[hbtp]
    \centering
    \begin{tabular}{lrrrr}
    \begin{subfigure}{0.01\linewidth} \caption{}\label{fig:suppl-compare-outdoor-ours} \end{subfigure} 
    \hspace{0.5em}
       \includegraphics[width=.235\hsize, height=.235\hsize]{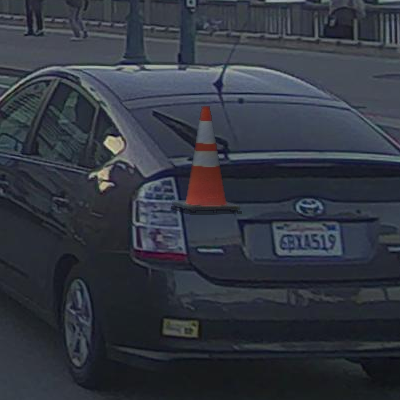} 
       \includegraphics[width=.235\hsize, height=.235\hsize]{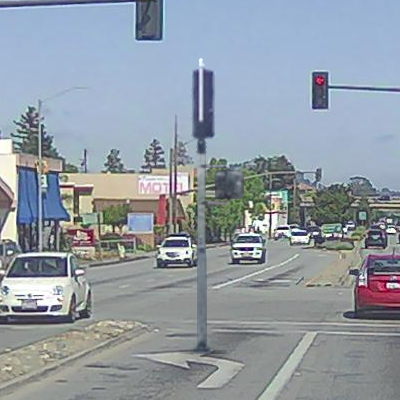}
       \includegraphics[width=.235\hsize, height=.235\hsize]{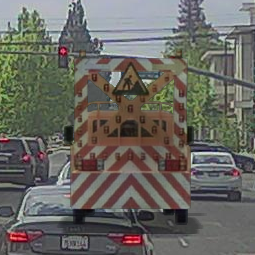} 
       \includegraphics[width=.235\hsize, height=.235\hsize]{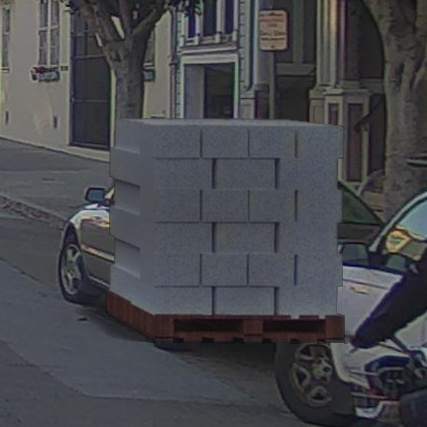}\\
    \begin{subfigure}{0.01\linewidth} \caption{}\label{fig:suppl-compare-outdoor-hdr} \end{subfigure} 
    \hspace{0.5em}
       \includegraphics[width=.235\hsize, height=.235\hsize]{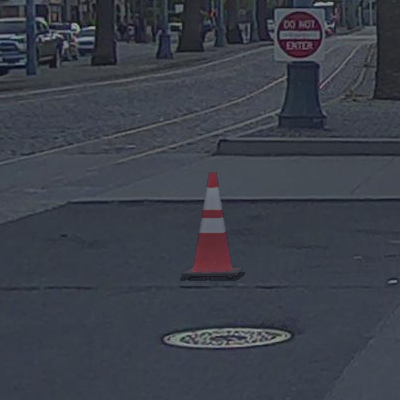}    
       \includegraphics[width=.235\hsize, height=.235\hsize]{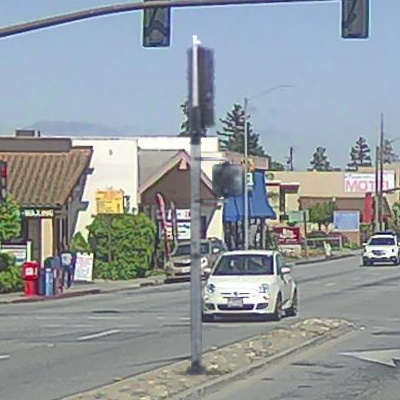}
       \includegraphics[width=.235\hsize, height=.235\hsize]{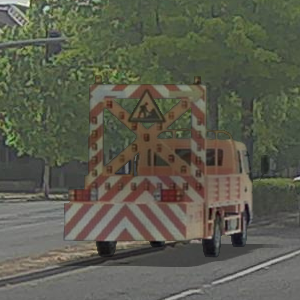}    
       \includegraphics[width=.235\hsize, height=.235\hsize]{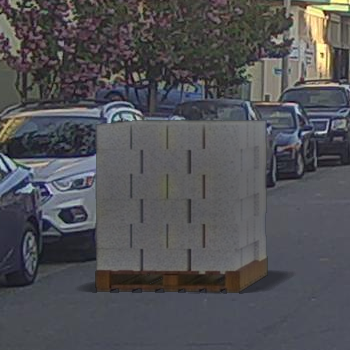}\\ 
    \begin{subfigure}{0.01\linewidth} \caption{}\label{fig:suppl-compare-outdoor-shadow} \end{subfigure} 
    \hspace{0.5em}
       \includegraphics[width=.235\hsize, height=.235\hsize]{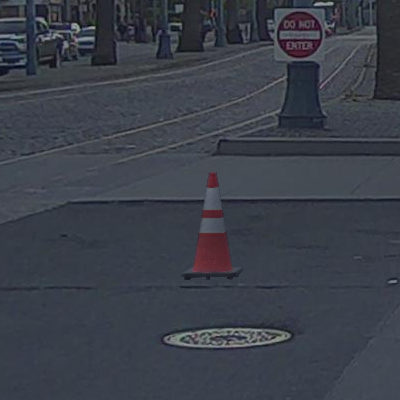} 
       \includegraphics[width=.235\hsize, height=.235\hsize]{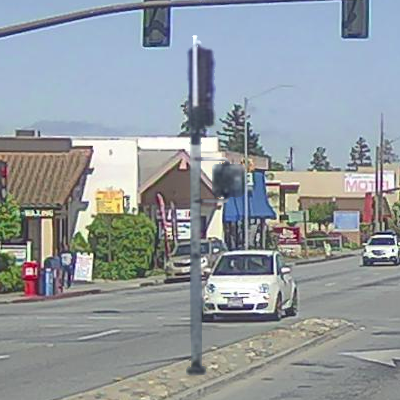}
       \includegraphics[width=.235\hsize, height=.235\hsize]{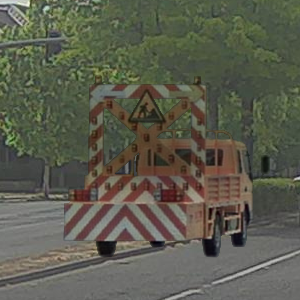} 
       \includegraphics[width=.235\hsize, height=.235\hsize]{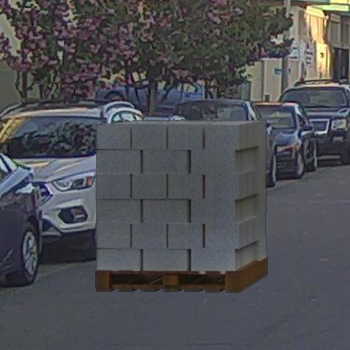}\\ 
    \begin{subfigure}{0.01\linewidth} \caption{}\label{fig:suppl-compare-outdoor-transfer} \end{subfigure} 
    \hspace{0.5em}
       \includegraphics[width=.235\hsize, height=.235\hsize]{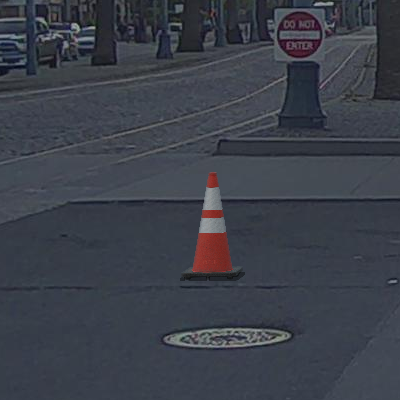} 
       \includegraphics[width=.235\hsize, height=.235\hsize]{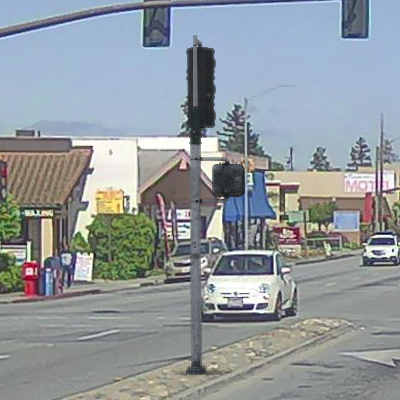}
       \includegraphics[width=.235\hsize, height=.235\hsize]{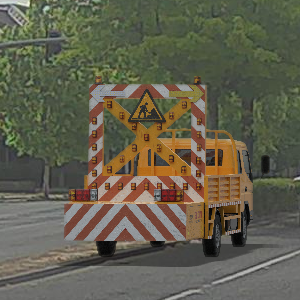} 
       \includegraphics[width=.235\hsize, height=.235\hsize]{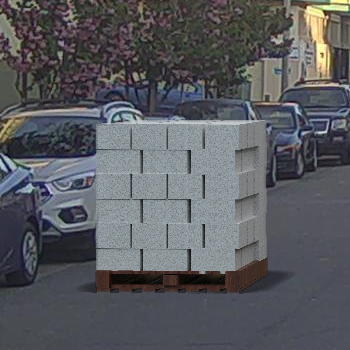}\\
      \begin{subfigure}{0.01\linewidth} \caption{}\label{fig:suppl-compare-outdoor-transfer} \end{subfigure} 
      \hspace{0.5em}
       \includegraphics[width=.235\hsize, height=.235\hsize]{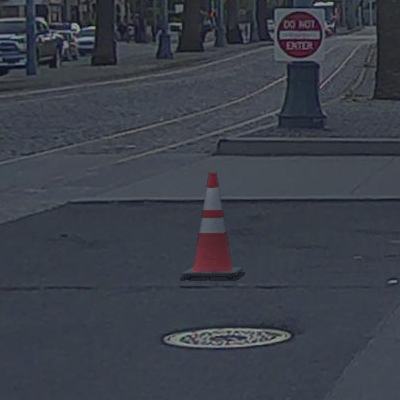} 
       \includegraphics[width=.235\hsize, height=.235\hsize]{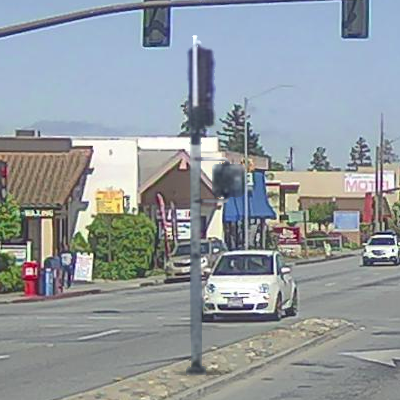}
       \includegraphics[width=.235\hsize, height=.235\hsize]{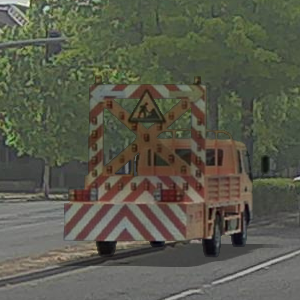} 
       \includegraphics[width=.235\hsize, height=.235\hsize]{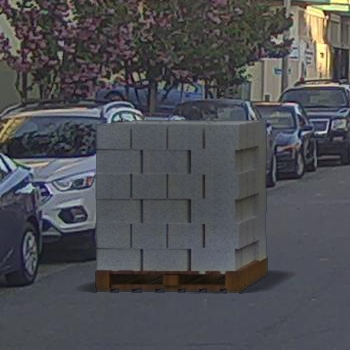}\\
    \end{tabular}

    \caption{Qualitative comparison of a simulated video frame using outdoor scene dataset PandaSet with different rendering options. (a) generated by the method without object placement; (b) generated by the method without HDR image reconstruction; (c) generated by the method without shadow generation (d) generated by the method without style transfer (e) generated by our proposed method including all rendering options.}
    \label{fig:suppl-compare-style-outdoor}
\end{figure*}

\begin{figure*}[hbtp]
    \centering
    \begin{tabular}{lrrrr}
    \begin{subfigure}{0.01\linewidth} \caption{}\label{fig:suppl-compare-outdoor-ours} \end{subfigure} 
    \hspace{0.5em}
       \includegraphics[width=.235\hsize, height=.235\hsize]{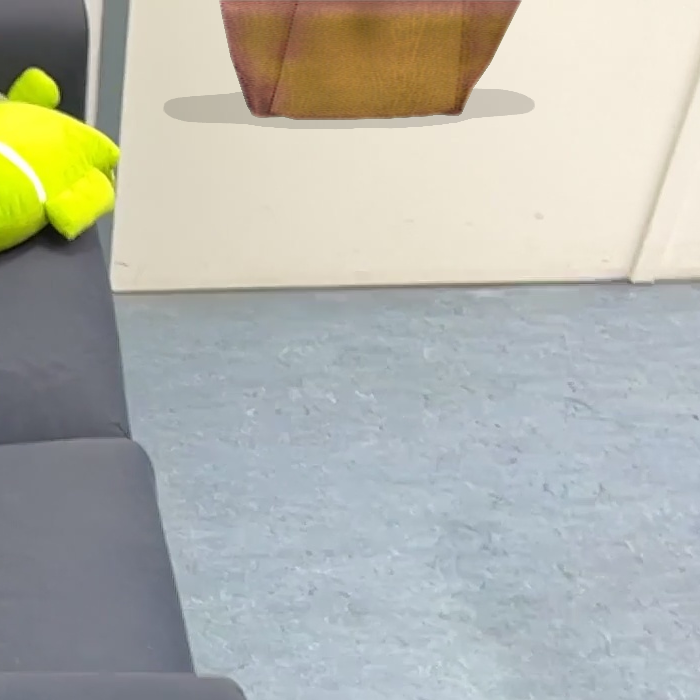} 
       \includegraphics[width=.235\hsize, height=.235\hsize]{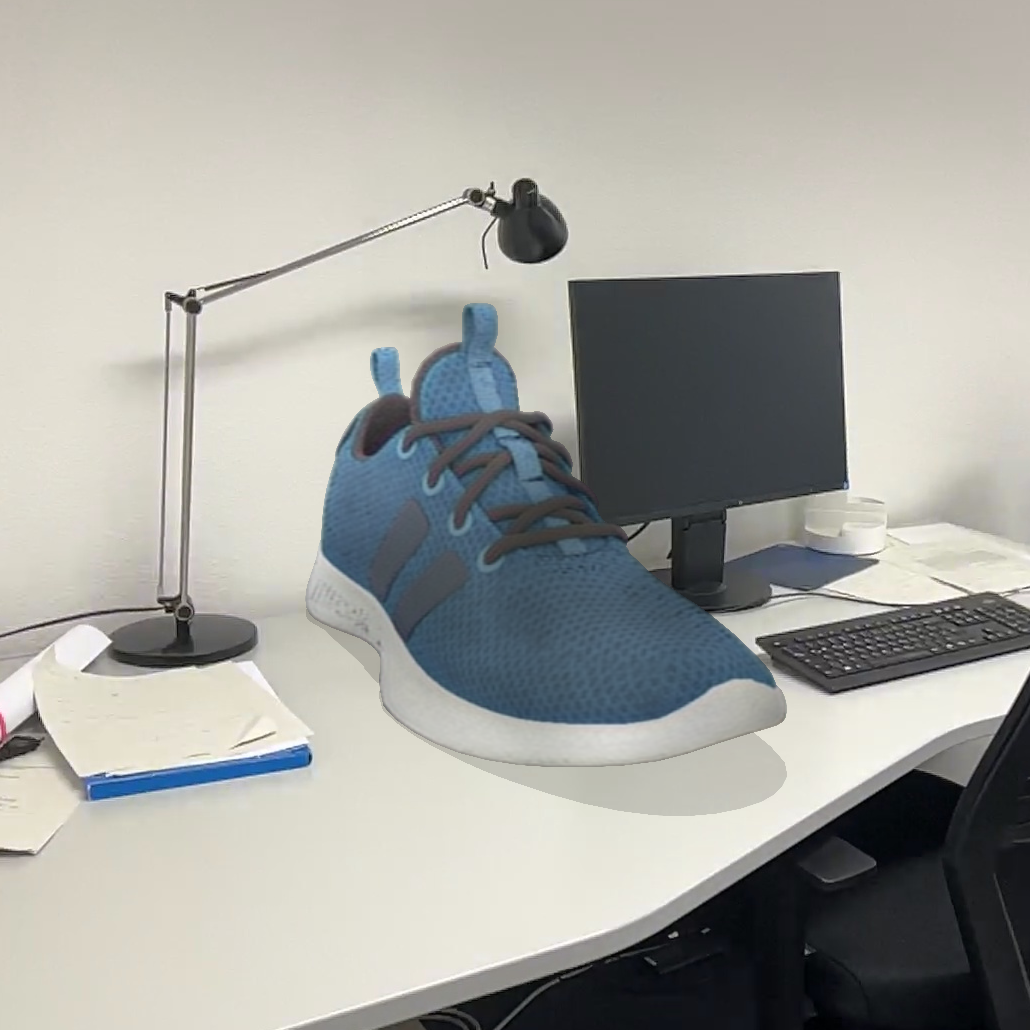}
       \includegraphics[width=.235\hsize, height=.235\hsize]{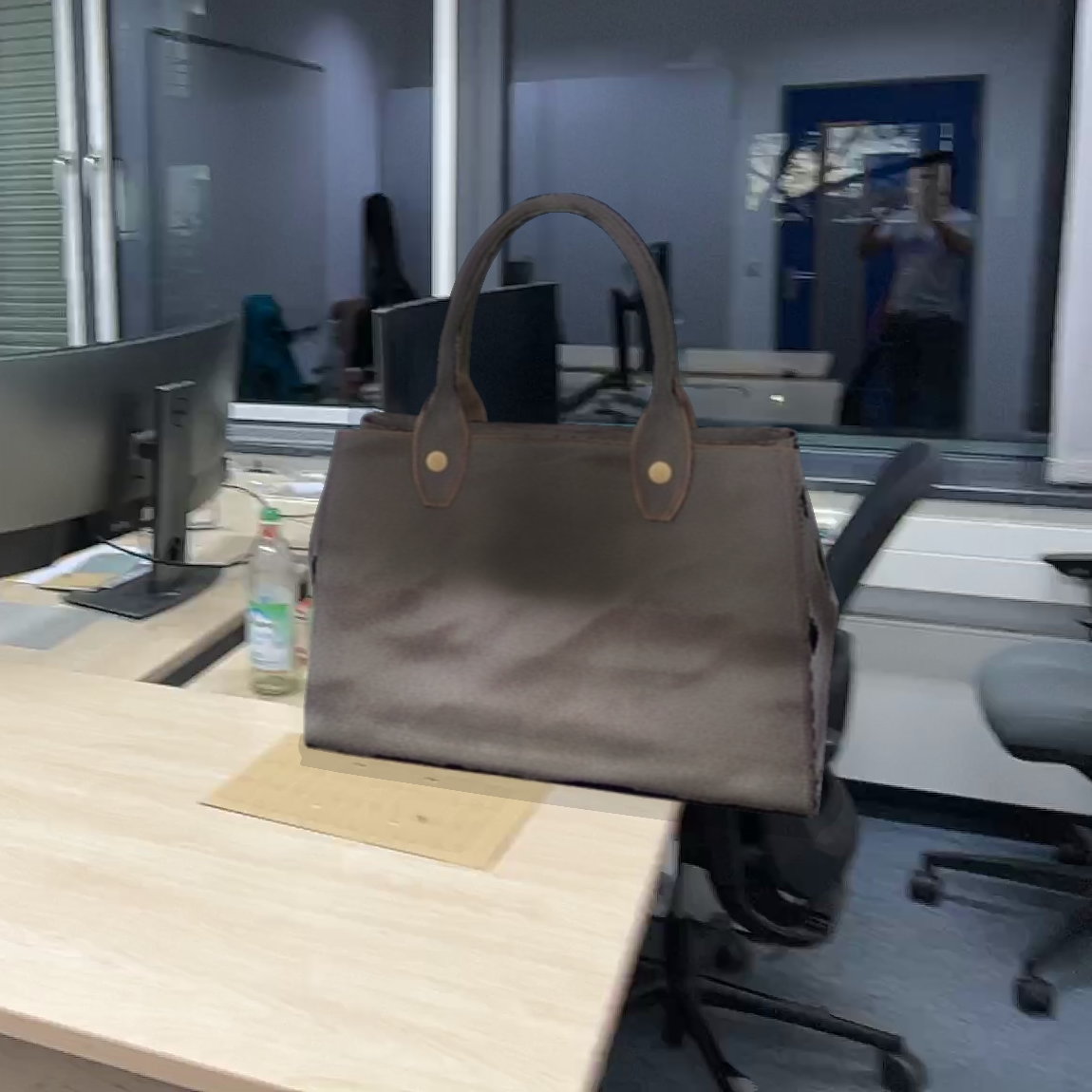} 
       \includegraphics[width=.235\hsize, height=.235\hsize]{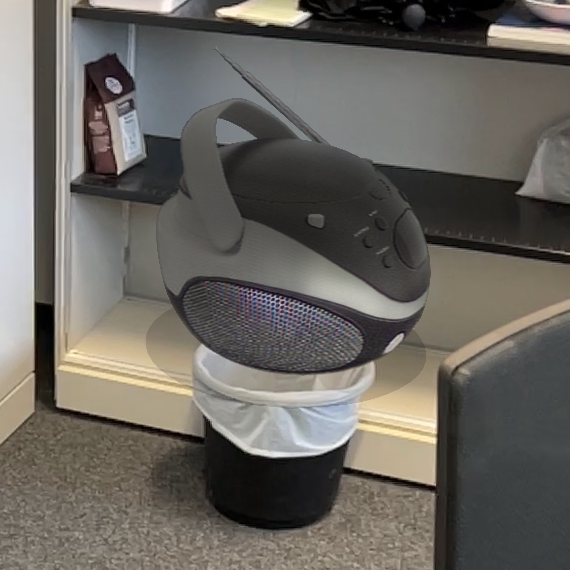}\\
    \begin{subfigure}{0.01\linewidth} \caption{}\label{fig:suppl-compare-outdoor-hdr} \end{subfigure} 
    \hspace{0.5em}
       \includegraphics[width=.235\hsize, height=.235\hsize]{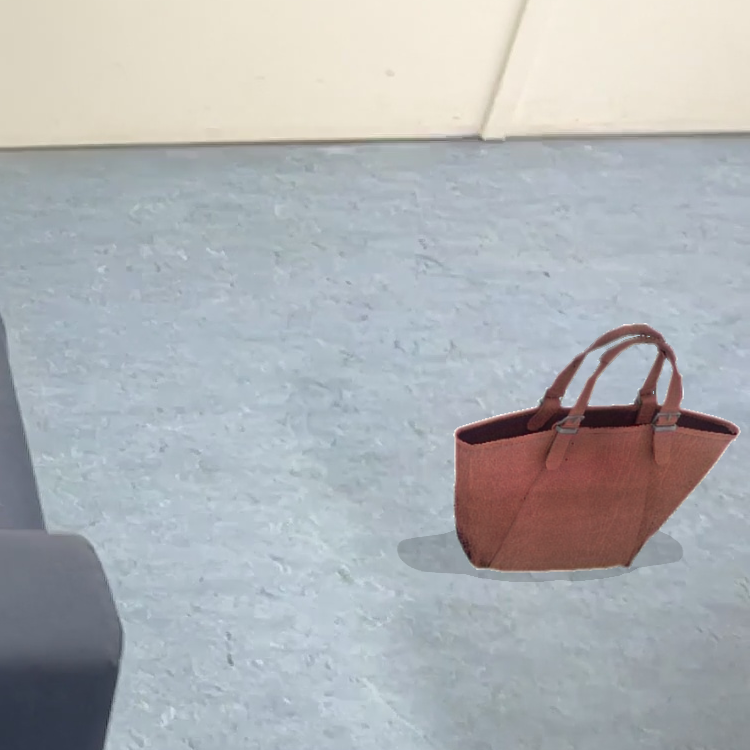}    
       \includegraphics[width=.235\hsize, height=.235\hsize]{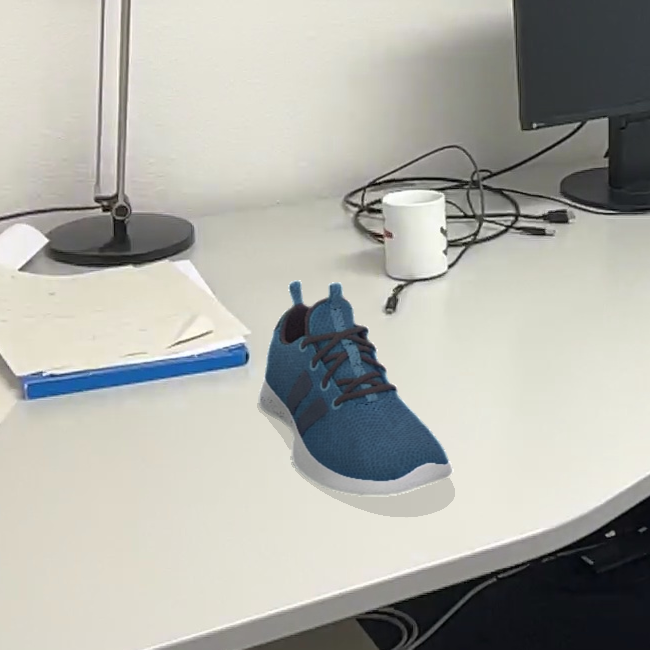}
       \includegraphics[width=.235\hsize, height=.235\hsize]{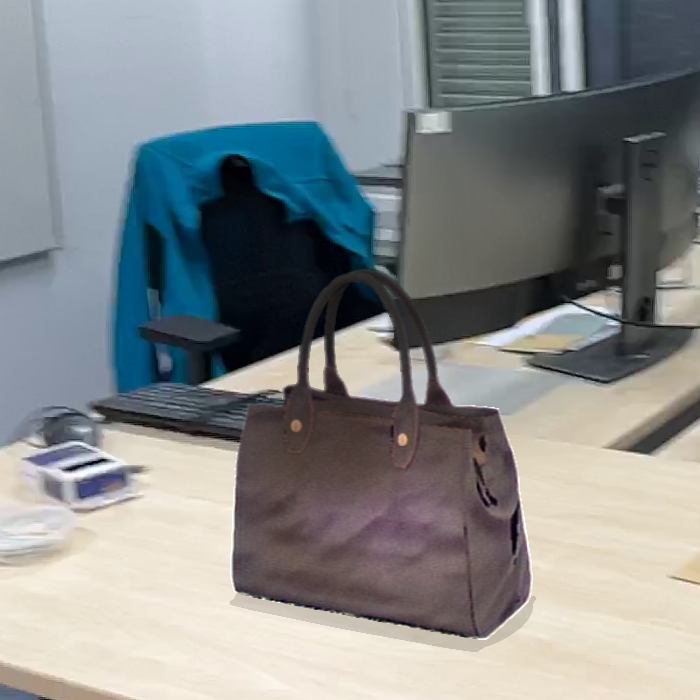}    
       \includegraphics[width=.235\hsize, height=.235\hsize]{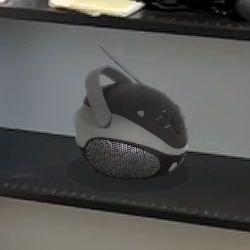}\\ 
    \begin{subfigure}{0.01\linewidth} \caption{}\label{fig:suppl-compare-outdoor-shadow} \end{subfigure} 
    \hspace{0.5em}
       \includegraphics[width=.235\hsize, height=.235\hsize]{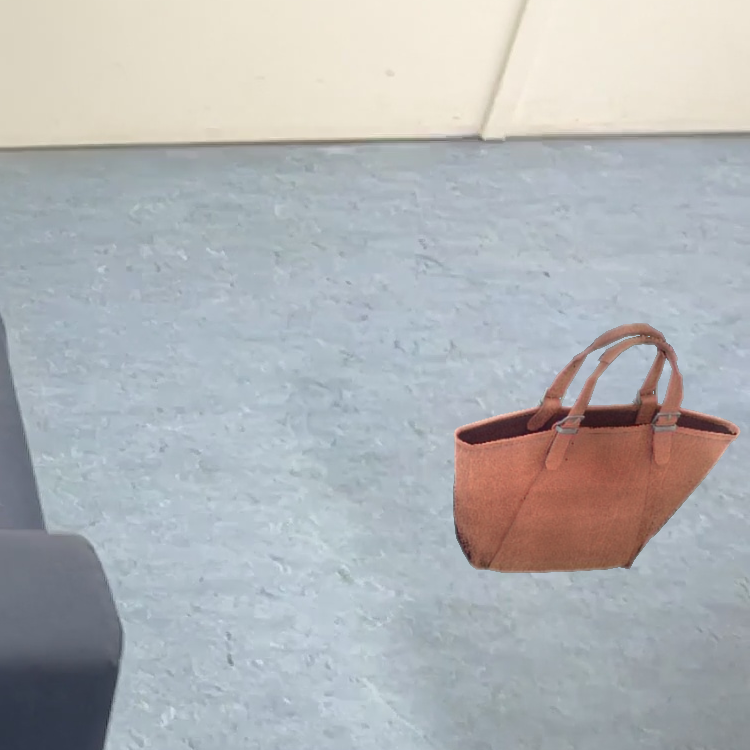} 
       \includegraphics[width=.235\hsize, height=.235\hsize]{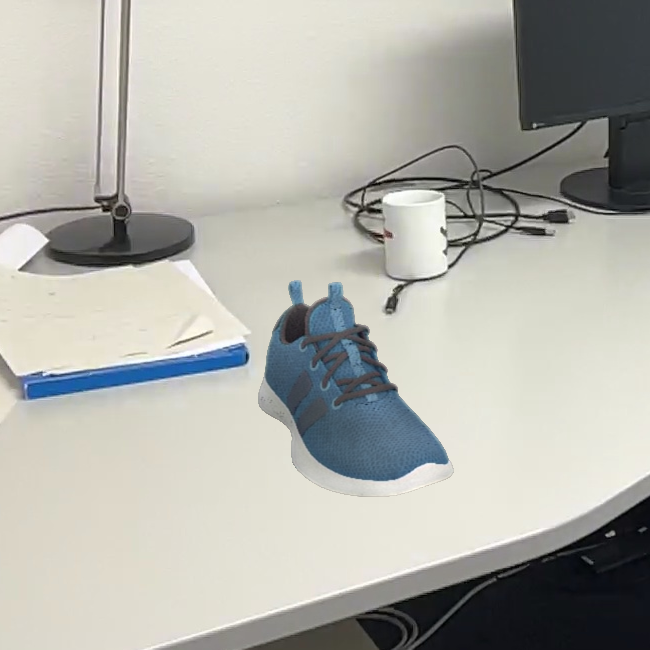}
       \includegraphics[width=.235\hsize, height=.235\hsize]{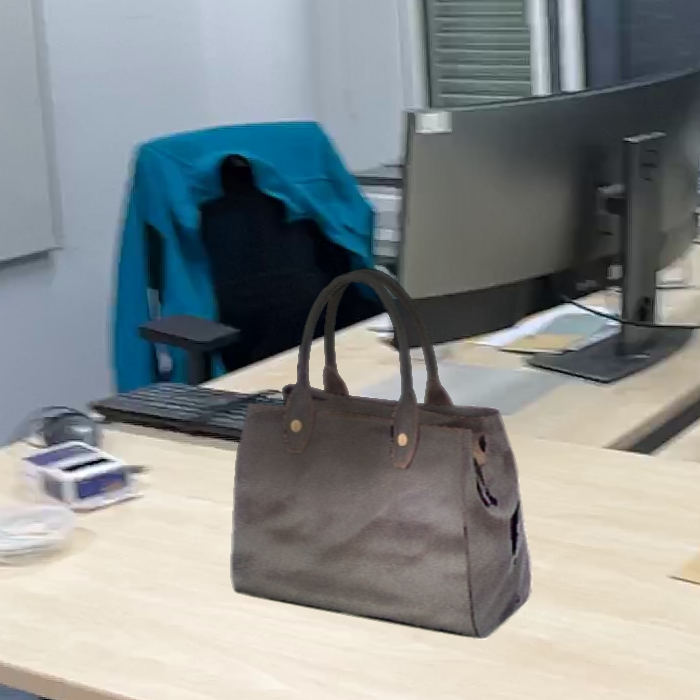} 
       \includegraphics[width=.235\hsize, height=.235\hsize]{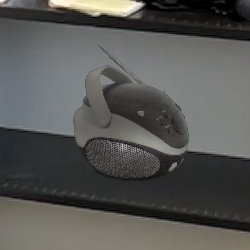}\\ 
    \begin{subfigure}{0.01\linewidth} \caption{}\label{fig:suppl-compare-outdoor-transfer} \end{subfigure} 
    \hspace{0.5em}
       \includegraphics[width=.235\hsize, height=.235\hsize]{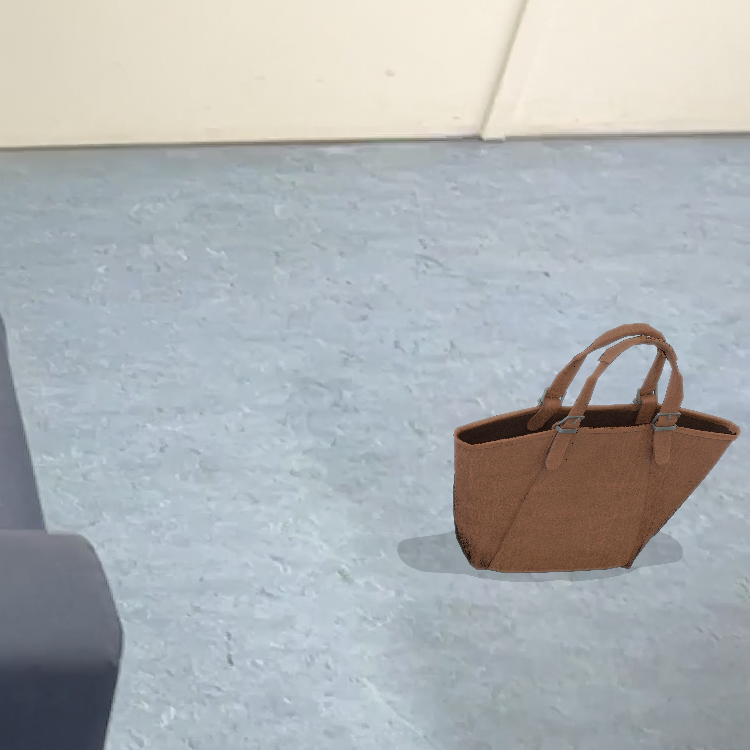} 
       \includegraphics[width=.235\hsize, height=.235\hsize]{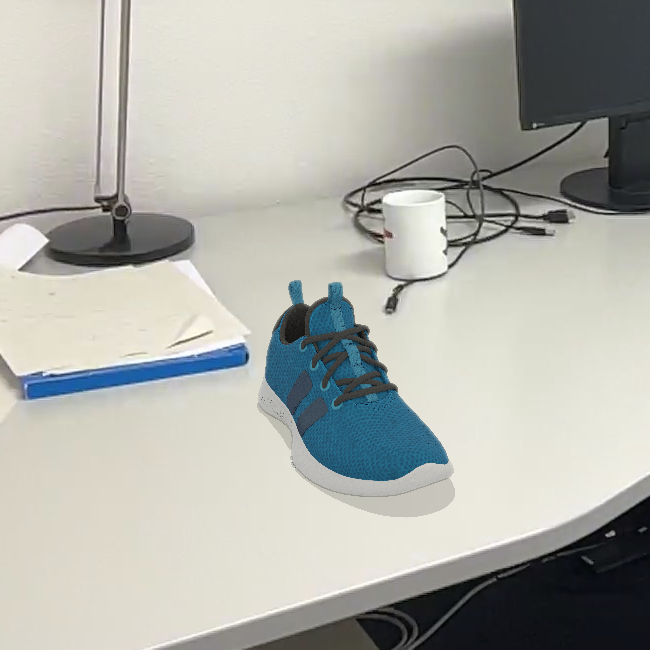}
       \includegraphics[width=.235\hsize, height=.235\hsize]{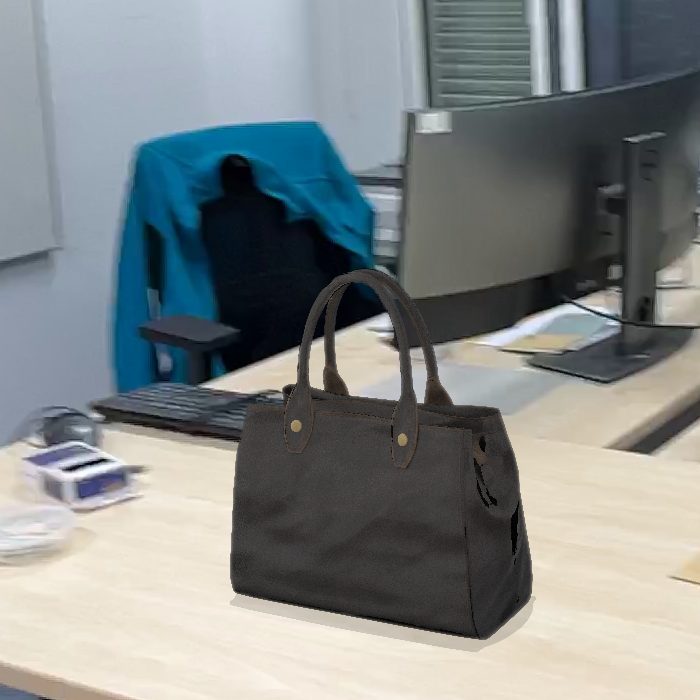} 
       \includegraphics[width=.235\hsize, height=.235\hsize]{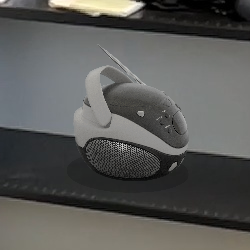}\\
      \begin{subfigure}{0.01\linewidth} \caption{}\label{fig:suppl-compare-outdoor-transfer} \end{subfigure} 
      \hspace{0.5em}
       \includegraphics[width=.235\hsize, height=.235\hsize]{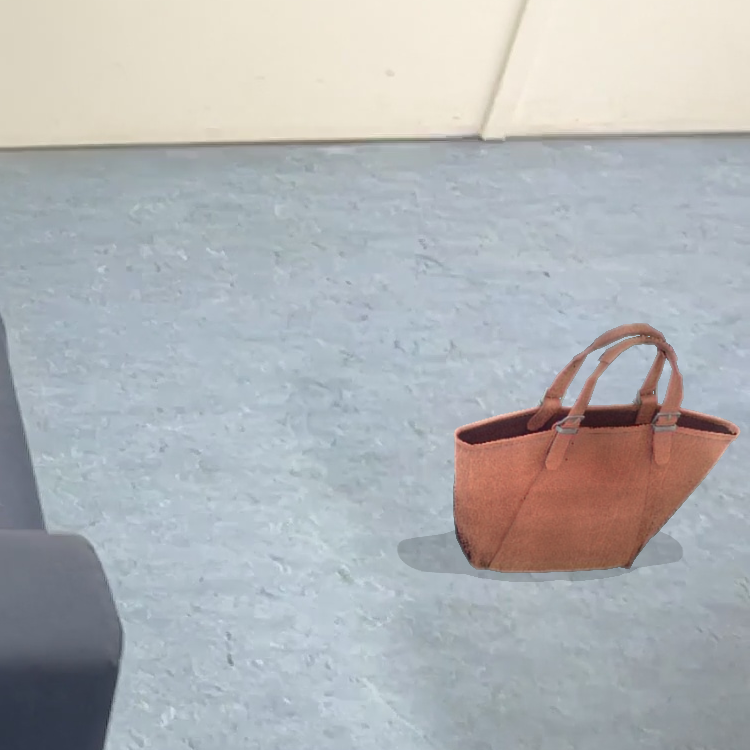} 
       \includegraphics[width=.235\hsize, height=.235\hsize]{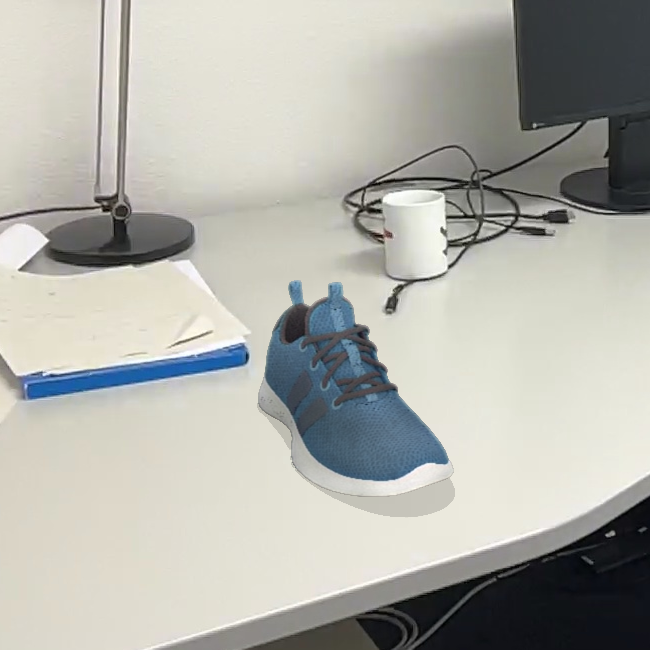}
       \includegraphics[width=.235\hsize, height=.235\hsize]{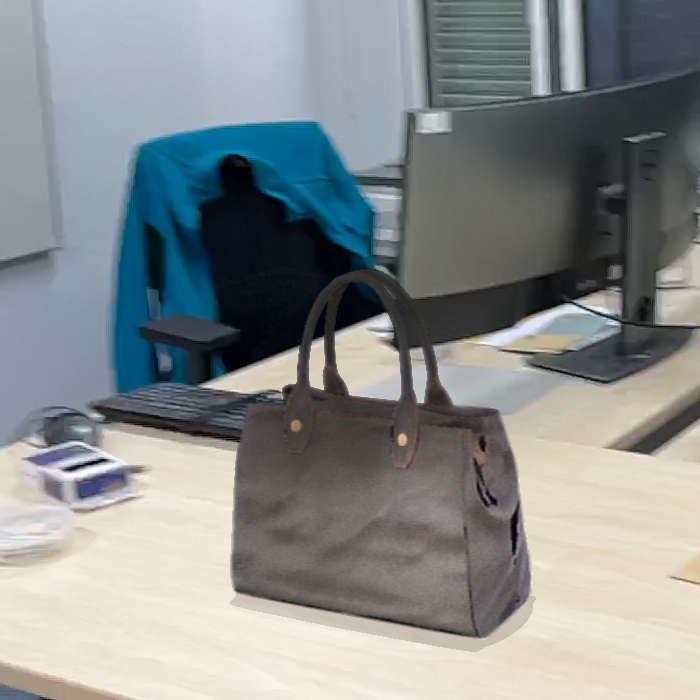} 
       \includegraphics[width=.235\hsize, height=.235\hsize]{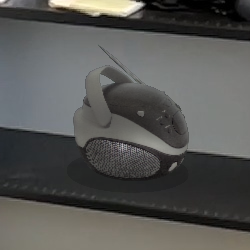}\\
    \end{tabular}

    \caption{Qualitative comparison of a simulated video frame using indoor scene dataset ScanNet++ with different rendering options. (a) generated by the method without object placement; (b) generated by the method without HDR image reconstruction; (c) generated by the method without shadow generation (d) generated by the method without style transfer (e) generated by our proposed method including all rendering options.}
    \label{fig:suppl-compare-style-indoor}
\end{figure*}

\textbf{Style Transfer Network:}
In Figure~\ref{fig:suppl-compare-style-outdoor} and~\ref{fig:suppl-compare-style-indoor}, we demonstrate the qualitative comparison of sample video frames generated by different style transfer networks using both the outdoor scene dataset PandaSet and the indoor scene dataset ScanNet+.
The foreground region of images refined by DoveNet has significant grid pattern artifacts and is much blurry compared to the background regions.
As for the refined images generated by StyTR2 and PHDiffusion, the color tone of the inserted object is not consistent with the weather and sunlight environment in the scene.
The refined image generated by our proposed method has the best visual quality compared to the other three, and our proposed method achieved the best result in both FID and human study scores as reported in the quantitative result.
    


\begin{figure*}[hbtp]
    \centering
    \begin{tabular}{cccc}
    
    \begin{subfigure}{0.01\linewidth} \caption{}\label{fig:suppl-compare-outdoor-dove} \end{subfigure} 
    \hspace{0.5em}
       \includegraphics[width=.3\hsize,height=.3\hsize]{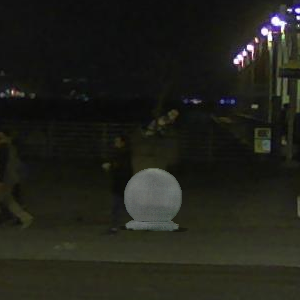}    
       \includegraphics[width=.3\hsize,height=.3\hsize]{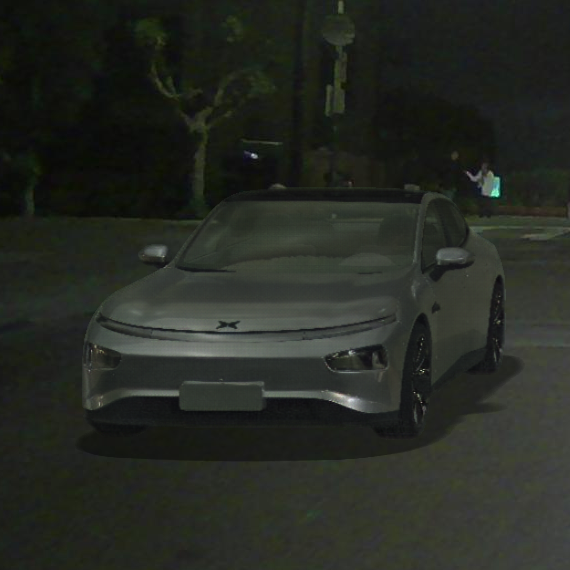}
       \includegraphics[width=.3\hsize,height=.3\hsize]{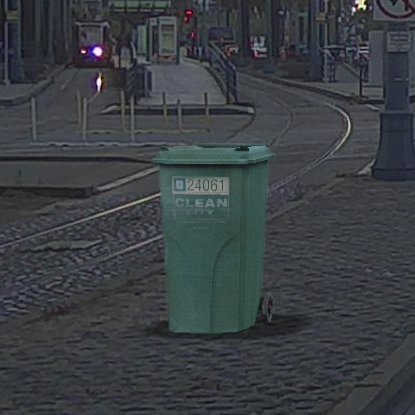}\\ 
    \begin{subfigure}{0.01\linewidth} \caption{}\label{fig:suppl-compare-outdoor-stytr} \end{subfigure} 
    \hspace{0.5em}
       \includegraphics[width=.3\hsize,height=.3\hsize]{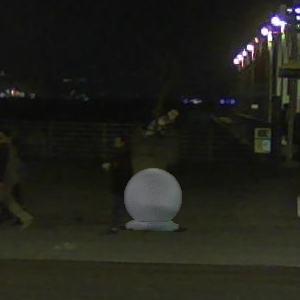} 
       \includegraphics[width=.3\hsize,height=.3\hsize]{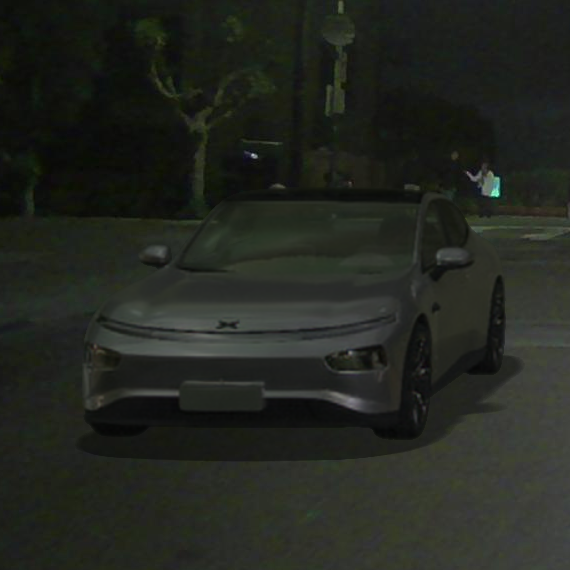}
       \includegraphics[width=.3\hsize,height=.3\hsize]{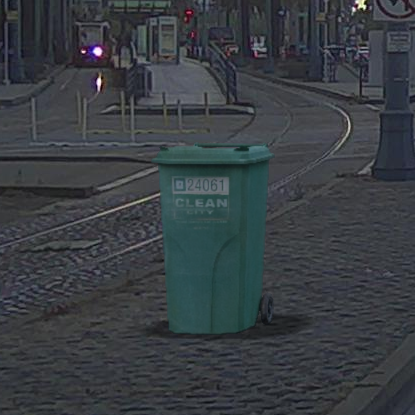}\\ 
    \begin{subfigure}{0.01\linewidth} \caption{}\label{fig:suppl-compare-outdoor-diffusion} \end{subfigure} 
    \hspace{0.5em}
       \includegraphics[width=.3\hsize,height=.3\hsize]{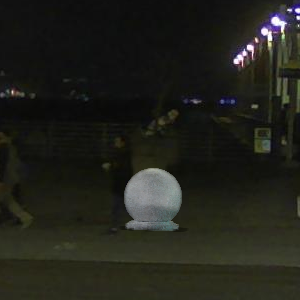} 
       \includegraphics[width=.3\hsize,height=.3\hsize]{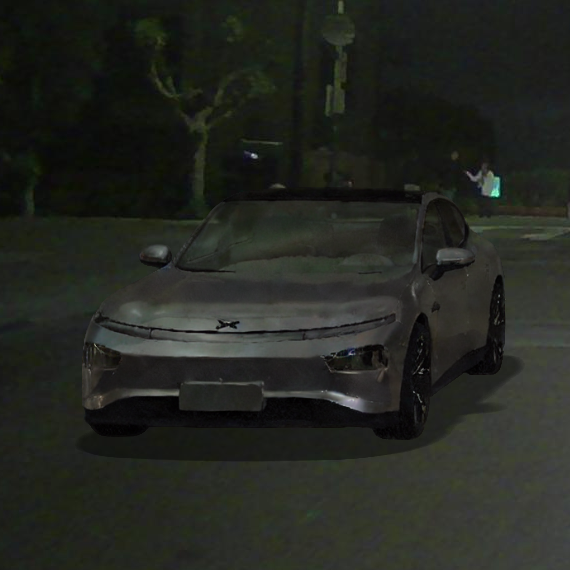}
       \includegraphics[width=.3\hsize,height=.3\hsize]{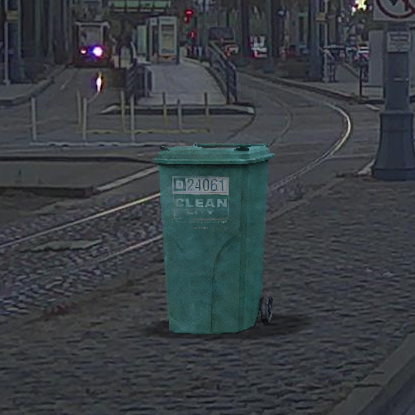}\\
    \begin{subfigure}{0.01\linewidth} \caption{}\label{fig:suppl-compare-outdoor-ours} \end{subfigure} 
    \hspace{0.5em}
       \includegraphics[width=.3\hsize,height=.3\hsize]{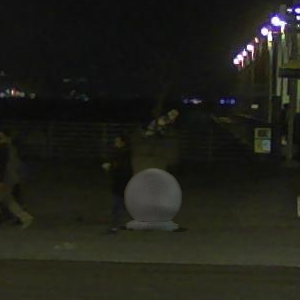} 
       \includegraphics[width=.3\hsize,height=.3\hsize]{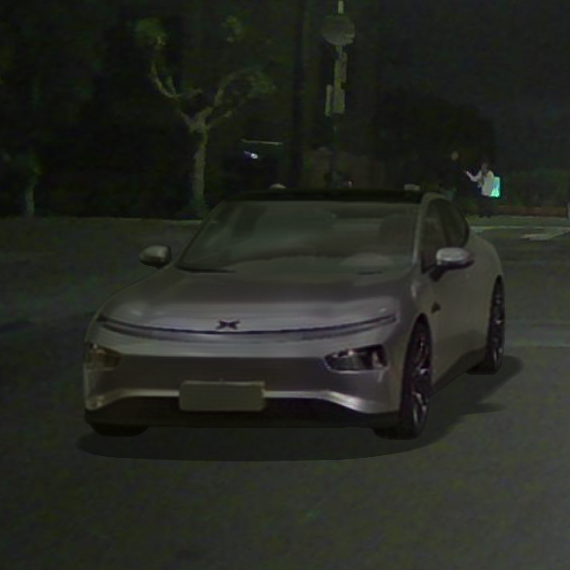}
       \includegraphics[width=.3\hsize,height=.3\hsize]{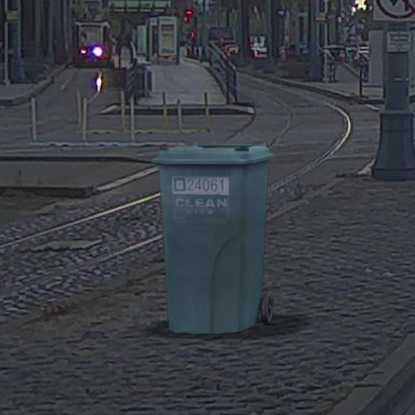}\\ 
    \end{tabular}

    \caption{Qualitative comparison of a simulated video frame using outdoor scene dataset PandaSet with different style transfer networks. (a) generated by DoveNet; (b) generated by StyTR2; (c) generated by PHDiffusion (d) generated by our proposed style transfer network}
    \label{fig:suppl-compare-ablation-outdoor}
\end{figure*}

\begin{figure*}[hbtp]
    \centering
    \begin{tabular}{cccc}
    
    \begin{subfigure}{0.01\linewidth} \caption{}\label{fig:suppl-compare-outdoor-dove} \end{subfigure} 
    \hspace{0.5em}
       \includegraphics[width=.3\hsize,height=.3\hsize]{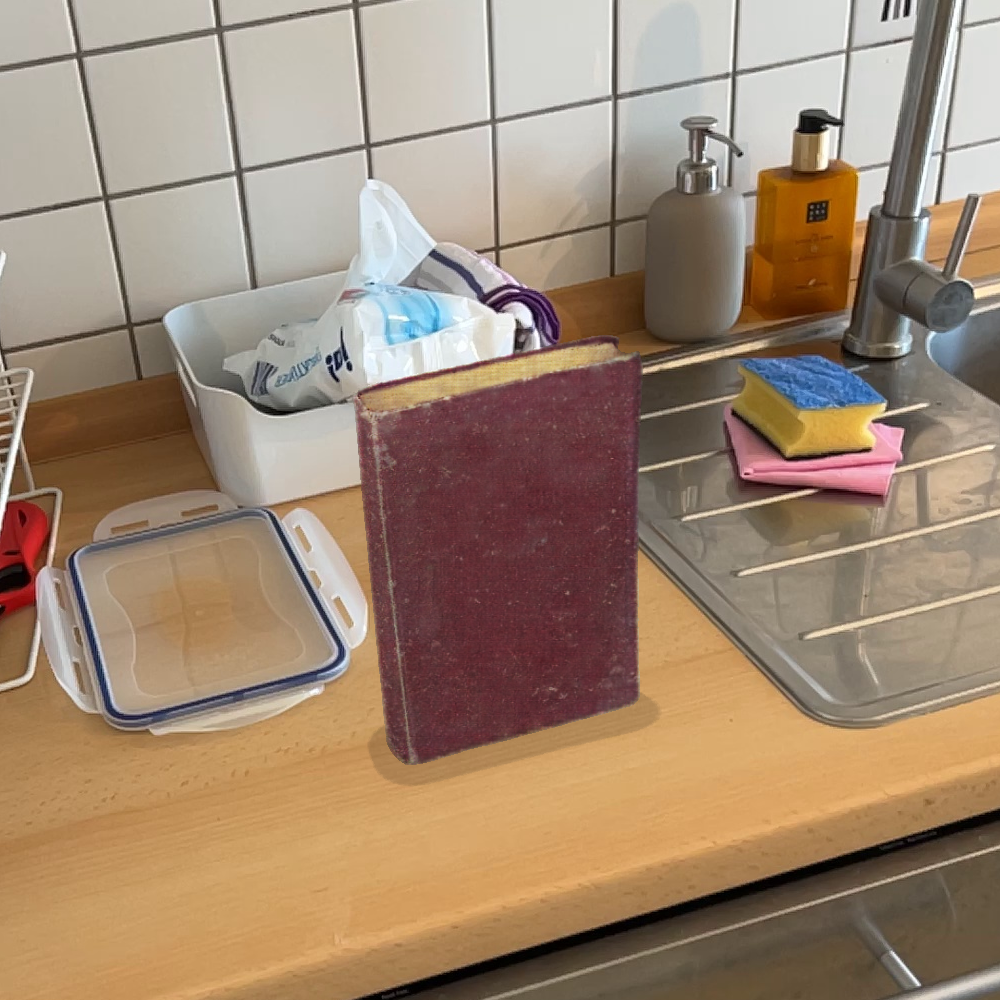}    
       \includegraphics[width=.3\hsize,height=.3\hsize]{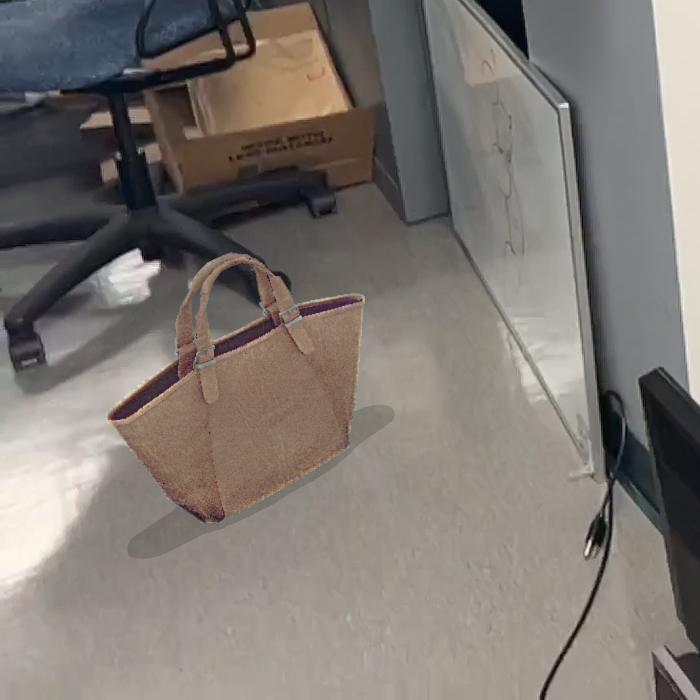}
       \includegraphics[width=.3\hsize,height=.3\hsize]{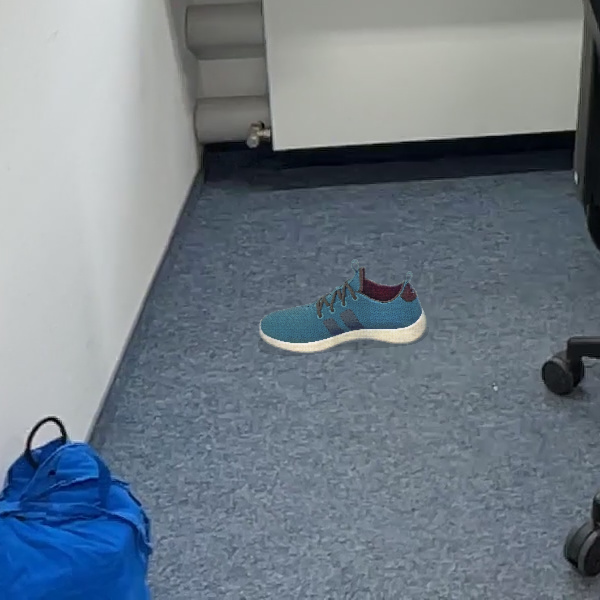}\\ 
    \begin{subfigure}{0.01\linewidth} \caption{}\label{fig:suppl-compare-outdoor-stytr} \end{subfigure} 
    \hspace{0.5em}
       \includegraphics[width=.3\hsize,height=.3\hsize]{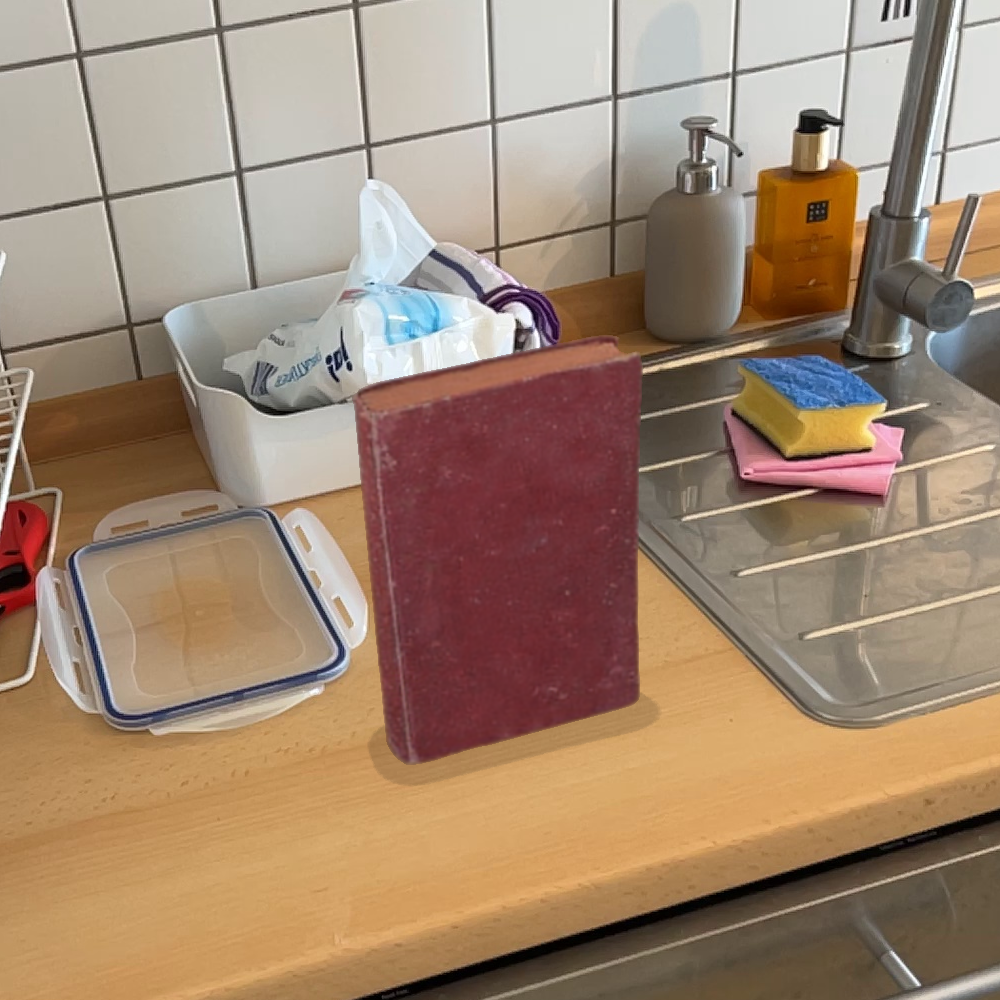} 
       \includegraphics[width=.3\hsize,height=.3\hsize]{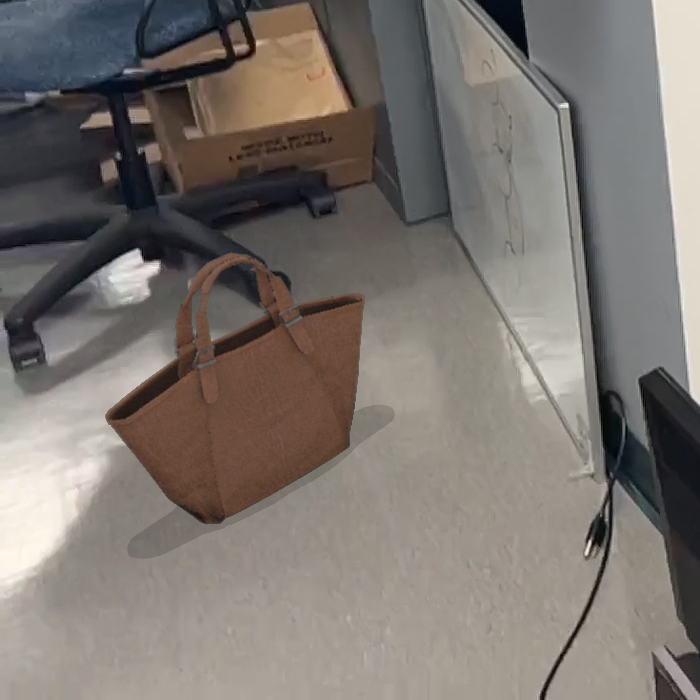}
       \includegraphics[width=.3\hsize,height=.3\hsize]{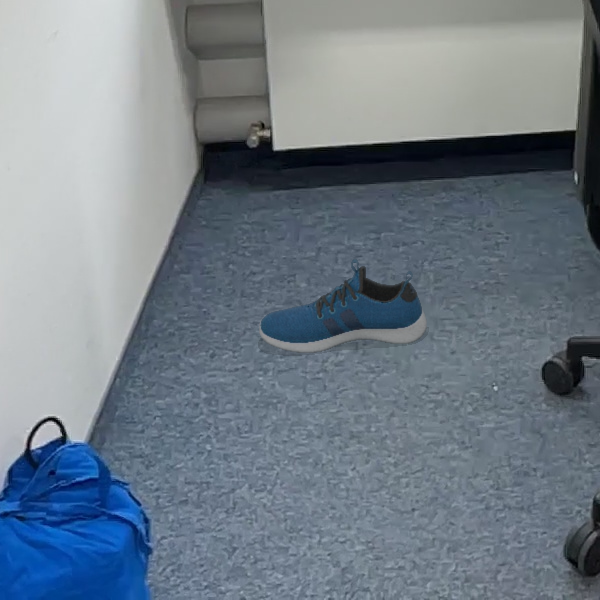}\\ 
    \begin{subfigure}{0.01\linewidth} \caption{}\label{fig:suppl-compare-outdoor-diffusion} \end{subfigure} 
    \hspace{0.5em}
       \includegraphics[width=.3\hsize,height=.3\hsize]{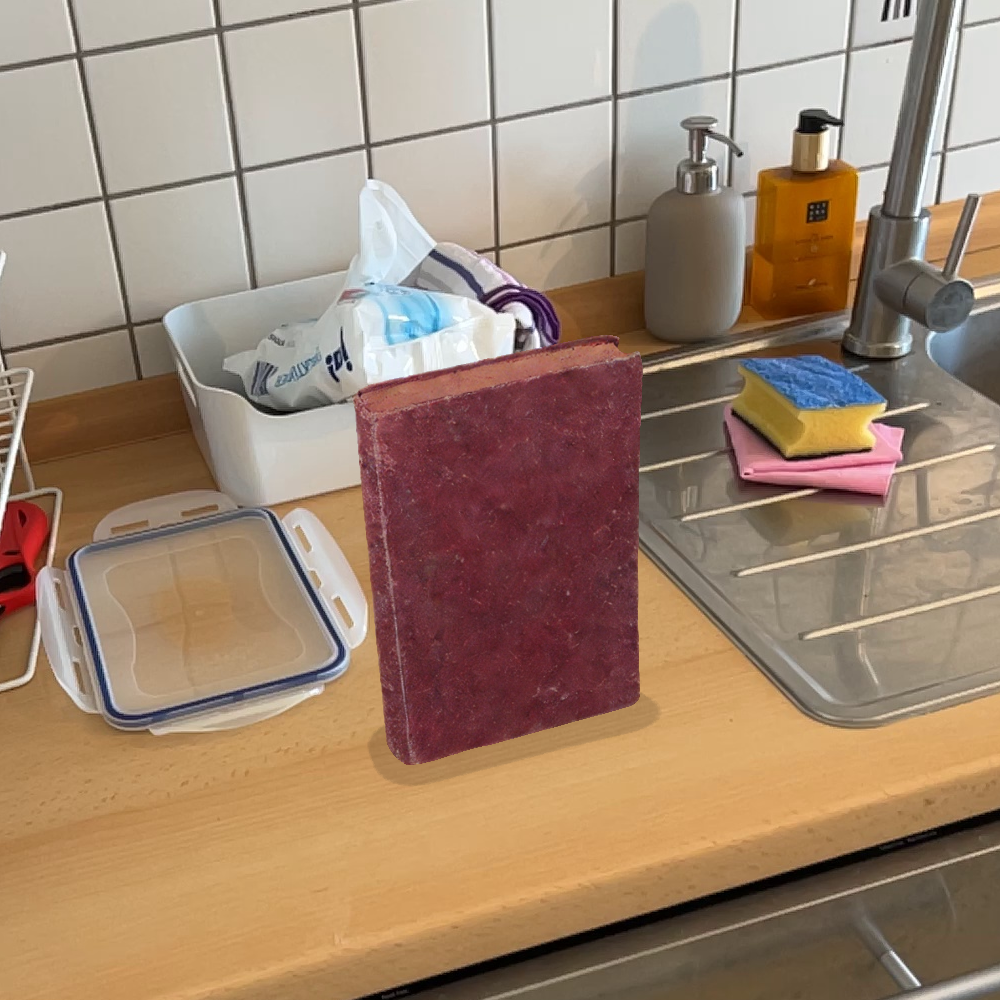} 
       \includegraphics[width=.3\hsize,height=.3\hsize]{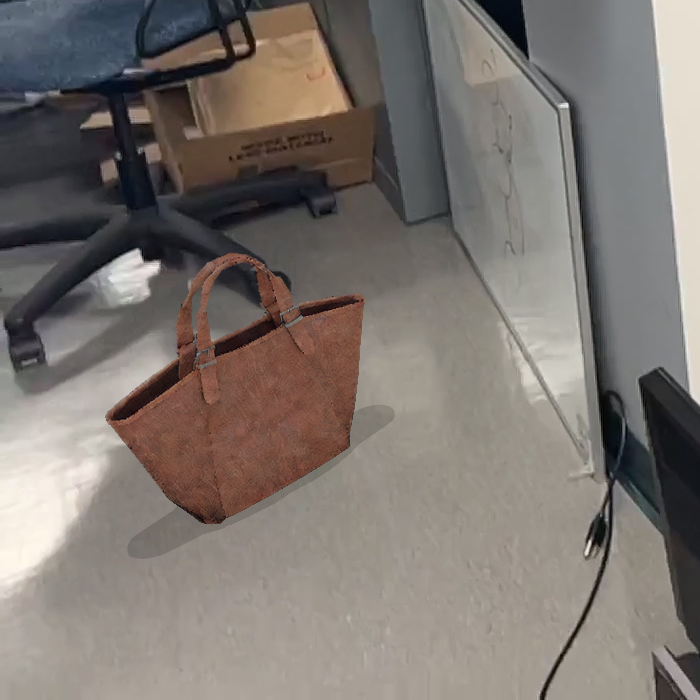}
       \includegraphics[width=.3\hsize,height=.3\hsize]{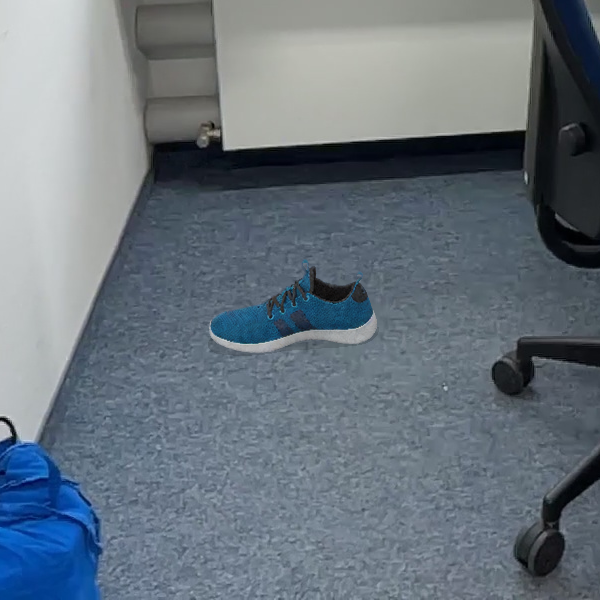}\\
    \begin{subfigure}{0.01\linewidth} \caption{}\label{fig:suppl-compare-outdoor-ours} \end{subfigure} 
    \hspace{0.5em}
       \includegraphics[width=.3\hsize,height=.3\hsize]{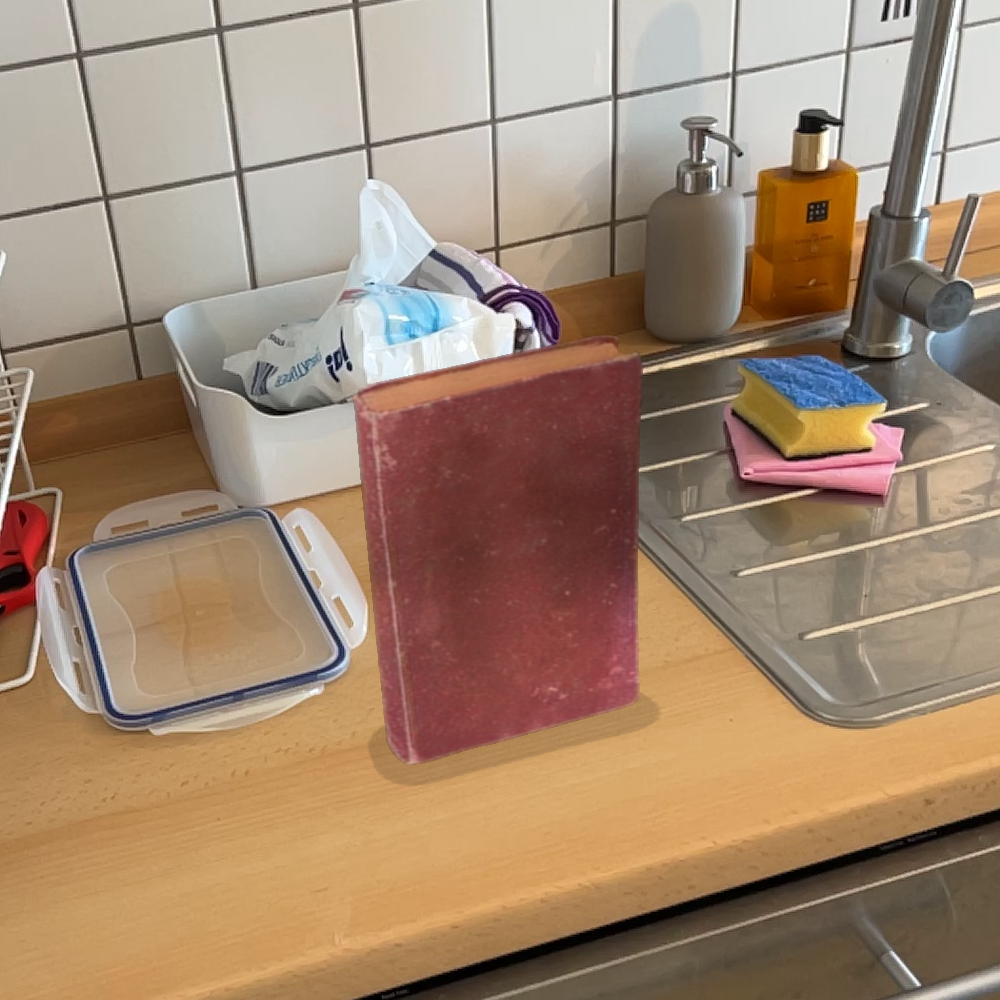} 
       \includegraphics[width=.3\hsize,height=.3\hsize]{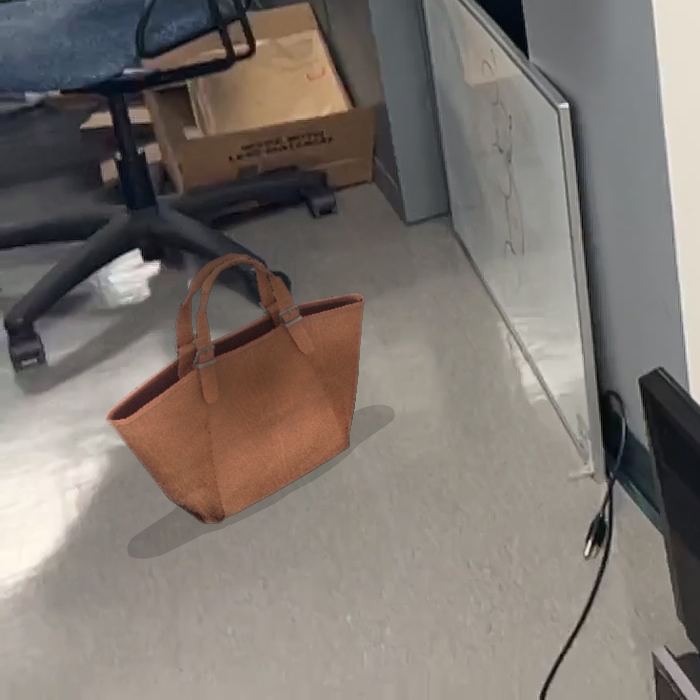}
       \includegraphics[width=.3\hsize,height=.3\hsize]{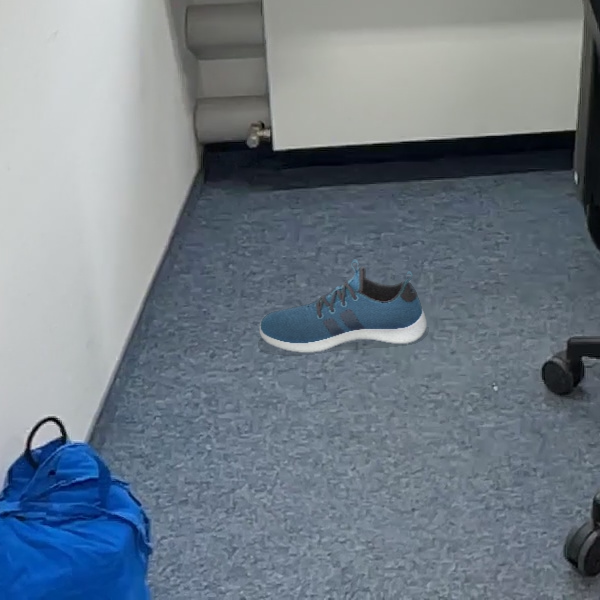}\\ 
    \end{tabular}

    \caption{Qualitative comparison of a simulated video frame using indoor scene dataset ScanNet++ with different style transfer networks. (a) generated by DoveNet; (b) generated by StyTR2; (c) generated by PHDiffusion (d) generated by our proposed style transfer network}
    \label{fig:suppl-compare-ablation-indoor}
\end{figure*}

\textbf{Ablation Analysis:}
We conducted an ablation analysis of each key rendering component including HDR image reconstruction, shadow generation, and style transfer. In Figure~\ref{fig:suppl-compare-ablation-outdoor} and Figure~\ref{fig:suppl-compare-ablation-indoor}, we show more qualitative comparisons of the simulated video frame with different rendering options using both the outdoor scene dataset PandaSet and the indoor scene dataset ScanNet+.

The inserted objects show inconsistent illumination a color balancing if there is no style transfer module or HDR reconstruction module involved in the video simulation process.
The inserted objects with no rendered shadow seem to be off the ground.



